\documentclass{article}
\usepackage{iclr2027_conference}
\usepackage{times}
\begingroup\fontfamily{ptm}\selectfont\endgroup
\DeclareFontShape{OT1}{ptm}{m}{scit}{<->ssub*ptm/m/sc}{}
\usepackage[nohints]{minitoc}
\usepackage{url}
\usepackage{graphicx}
\usepackage{xcolor}
\usepackage{wrapfig}
\usepackage{minted}
\usepackage[most]{tcolorbox}
\usepackage{booktabs}
\usepackage{multirow}
\usepackage{adjustbox}
\usepackage{enumitem}

\usepackage{microtype}
\usepackage{xspace}
\usepackage{amsmath}
\usepackage{amsfonts}
\usepackage{amssymb}
\usepackage{mathtools}
\usepackage{amsthm}
\usepackage{bm}
\tcbuselibrary{minted,skins,breakable}

\definecolor{specblue}{RGB}{38,84,150}
\definecolor{specbg}{RGB}{238,244,252}
\definecolor{axiomgreen}{RGB}{27,108,63}
\definecolor{axiombg}{RGB}{236,247,239}
\definecolor{patchgray}{RGB}{90,90,90}
\definecolor{patchbg}{RGB}{244,244,244}
\definecolor{failred}{RGB}{176,0,32}
\definecolor{provisional}{RGB}{200,30,30}

\definecolor{swb}{HTML}{00B0F0}
\definecolor{axo}{HTML}{E97132}
\definecolor{veri}{HTML}{1E8449}
\definecolor{ink}{HTML}{2C3E50}

\DeclareUnicodeCharacter{00D7}{\ensuremath{\times}}  
\DeclareUnicodeCharacter{2260}{\ensuremath{\ne}}     
\DeclareUnicodeCharacter{2227}{\ensuremath{\land}}   
\DeclareUnicodeCharacter{2200}{\ensuremath{\forall}} 
\DeclareUnicodeCharacter{2208}{\ensuremath{\in}}     
\DeclareUnicodeCharacter{2264}{\ensuremath{\le}}     
\DeclareUnicodeCharacter{2203}{\ensuremath{\exists}} 

\definecolor{mydarkblue}{rgb}{0,0.08,0.45}
\usepackage[colorlinks=true,
    linkcolor=mydarkblue,
    citecolor=mydarkblue,
    filecolor=mydarkblue,
    urlcolor=mydarkblue]{hyperref}
\usepackage{subcaption}
\usepackage{cleveref}
\usepackage{comment}

\makeatletter
\AtBeginDocument{\@savsf=\@m}
\makeatother

\newtheorem{claim}{Claim}

\crefname{claim}{Claim}{Claims}
\Crefname{claim}{Claim}{Claims}

\crefname{property}{Property}{Properties}
\crefname{theorem}{Theorem}{Theorems}
\crefname{lemma}{Lemma}{Lemmas}
\crefname{figure}{Figure}{Figures}
\crefname{equation}{Equation}{Equations}
\crefname{section}{Section}{Sections}
\crefname{table}{Table}{Tables}
\crefname{definition}{Definition}{Definitions}
\crefname{task}{Task}{Tasks}

\Crefname{property}{Property}{Properties}
\Crefname{theorem}{Theorem}{Theorems}
\Crefname{lemma}{Lemma}{Lemmas}
\Crefname{figure}{Figure}{Figures}
\Crefname{equation}{Equation}{Equations}
\Crefname{section}{Section}{Sections}
\Crefname{table}{Table}{Tables}
\Crefname{definition}{Definition}{Definitions}
\Crefname{task}{Task}{Tasks}

\newcommand{\rot}[1]{\rotatebox{60}{\small #1}}
\newcommand{\cmark}{\ensuremath{\bullet}}

\newcommand{\modefill}{\leavevmode\textcolor{gray!45}{\leaders\hrule height 3.8pt depth -1.6pt \hfill}\kern0pt}

\newcommand{\benchproofer}{\textsc{BenchProofer}\xspace}
\newcommand{\benchmark}{\textsc{SWE-Proof}\xspace}
\newcommand{\sbv}{\textsc{SWE-bench Verified}\xspace}
\newcommand{\swepro}{\textsc{SWE-bench Pro}\xspace}
\newcommand{\swebench}{\textsc{SWE-bench}\xspace}

\newcommand{\Description}[2][]{}

\setminted{fontsize=\footnotesize,breaklines=true,
  tabsize=4,autogobble,escapeinside=||}

\usepackage{seqsplit}
\DeclareRobustCommand{\id}[1]{\texttt{\seqsplit{#1}}}

\newtcolorbox[auto counter,number within=section,
  list inside=boxes]{exbox}[2][]{breakable,enhanced,
  colback=specbg,colframe=specblue,fonttitle=\bfseries\scriptsize,
  title={Box~\thetcbcounter: #2},
  title after break={Box~\thetcbcounter: #2 (continued)},lines before break=4,#1,
  boxrule=0.6pt,arc=2pt,left=3pt,right=3pt,top=2pt,bottom=2pt,
  before skip=5pt,after skip=5pt}
\newtcolorbox[use counter from=exbox]{specbox}[2][]{breakable,enhanced,
  colback=specbg,colframe=specblue,fonttitle=\bfseries\scriptsize,
  title={Box~\thetcbcounter: #2},
  title after break={Box~\thetcbcounter: #2 (continued)},lines before break=4,#1,
  boxrule=0.6pt,arc=2pt,left=3pt,right=3pt,top=2pt,bottom=2pt,
  before skip=5pt,after skip=5pt}
\newtcolorbox[use counter from=exbox]{axiombox}[2][]{breakable,enhanced,
  colback=axiombg,colframe=axiomgreen,fonttitle=\bfseries\scriptsize,
  title={Box~\thetcbcounter: #2},
  title after break={Box~\thetcbcounter: #2 (continued)},lines before break=4,#1,
  boxrule=0.6pt,arc=2pt,left=3pt,right=3pt,top=2pt,bottom=2pt,
  before skip=5pt,after skip=5pt}
\newtcolorbox[use counter from=exbox]{patchbox}[2][]{breakable,enhanced,
  colback=patchbg,colframe=patchgray,fonttitle=\bfseries\scriptsize,
  title={Box~\thetcbcounter: #2},
  title after break={Box~\thetcbcounter: #2 (continued)},lines before break=4,#1,
  boxrule=0.6pt,arc=2pt,left=3pt,right=3pt,top=2pt,bottom=2pt,
  before skip=5pt,after skip=5pt}

\usepackage{fvextra}

\newcommand{\nagini}{\textsc{Nagini}\xspace}
\newcommand{\velvet}{\textsc{Velvet}\xspace}
\newcommand{\lean}{\textsc{Lean}\xspace}
\newcommand{\ears}{\textsc{Ears}\xspace}

\usepackage{array}
\newcolumntype{L}[1]{>{\raggedright\arraybackslash}p{#1}}
\newcommand{\na}{\textendash}
\newcommand{\hd}[1]{\textbf{#1}}

\definecolor{agentbg}{HTML}{F5F8FC}
\definecolor{agentframe}{HTML}{4C72B0}
\definecolor{obsbg}{HTML}{F8F6F1}
\definecolor{obsframe}{HTML}{B08535}
\definecolor{promptbg}{HTML}{F4F4F6}
\definecolor{promptframe}{HTML}{8A8F98}
\definecolor{metabg}{HTML}{F1F6F2}
\definecolor{metaframe}{HTML}{55A868}
\definecolor{verdictbg}{HTML}{FBF3F4}
\definecolor{verdictframe}{HTML}{B0324A}

\newtcolorbox{trajmeta}[1][]{enhanced,breakable,colbacktitle=metabg,colback=metabg,colframe=metaframe,
  boxrule=0.4pt,left=4pt,right=4pt,top=3pt,bottom=3pt,
  fonttitle=\bfseries\footnotesize,coltitle=black,
  attach boxed title to top left={xshift=4pt,yshift=-2.4pt},
  title after break={\tcbtitletext\ (continued)},lines before break=4,
  boxed title style={colback=metabg,colframe=metaframe,boxrule=0.4pt,
                     left=3pt,right=3pt,top=1pt,bottom=1pt},#1}

\newtcolorbox{trajprompt}[1][]{enhanced,breakable,colbacktitle=promptbg,colback=promptbg,colframe=promptframe,
  boxrule=0.4pt,left=4pt,right=4pt,top=3pt,bottom=3pt,
  fonttitle=\bfseries\footnotesize,coltitle=black,
  attach boxed title to top left={xshift=4pt,yshift=-2.4pt},
  title after break={\tcbtitletext\ (continued)},lines before break=4,
  boxed title style={colback=promptbg,colframe=promptframe,boxrule=0.4pt,
                     left=3pt,right=3pt,top=1pt,bottom=1pt},#1}

\newtcolorbox{trajagent}[1][]{enhanced,breakable,colbacktitle=agentbg,colback=agentbg,colframe=agentframe,
  boxrule=0.4pt,left=4pt,right=4pt,top=3pt,bottom=3pt,
  fonttitle=\bfseries\footnotesize,coltitle=black,
  attach boxed title to top left={xshift=4pt,yshift=-2.4pt},
  title after break={\tcbtitletext\ (continued)},lines before break=4,
  boxed title style={colback=agentbg,colframe=agentframe,boxrule=0.4pt,
                     left=3pt,right=3pt,top=1pt,bottom=1pt},#1}

\newtcolorbox{trajobs}[1][]{enhanced,breakable,colbacktitle=obsbg,colback=obsbg,colframe=obsframe,
  boxrule=0.4pt,left=4pt,right=4pt,top=3pt,bottom=3pt,
  fonttitle=\bfseries\footnotesize,coltitle=black,
  attach boxed title to top left={xshift=4pt,yshift=-2.4pt},
  title after break={\tcbtitletext\ (continued)},lines before break=4,
  boxed title style={colback=obsbg,colframe=obsframe,boxrule=0.4pt,
                     left=3pt,right=3pt,top=1pt,bottom=1pt},#1}

\newtcolorbox{trajverdict}[1][]{enhanced,breakable,colbacktitle=verdictbg,colback=verdictbg,colframe=verdictframe,
  boxrule=0.4pt,left=4pt,right=4pt,top=3pt,bottom=3pt,
  fonttitle=\bfseries\footnotesize,coltitle=black,
  attach boxed title to top left={xshift=4pt,yshift=-2.4pt},
  title after break={\tcbtitletext\ (continued)},lines before break=4,
  boxed title style={colback=verdictbg,colframe=verdictframe,boxrule=0.4pt,
                     left=3pt,right=3pt,top=1pt,bottom=1pt},#1}

\DefineVerbatimEnvironment{trajtext}{Verbatim}
  {fontsize=\scriptsize,baselinestretch=0.95,breaklines=true,
   breakindent=0pt,xleftmargin=0pt}

\DeclareUnicodeCharacter{00B7}{\ensuremath{\cdot}}
\DeclareUnicodeCharacter{00AC}{\ensuremath{\neg}}
\DeclareUnicodeCharacter{00B1}{\ensuremath{\pm}}
\DeclareUnicodeCharacter{2018}{\textquoteleft}
\DeclareUnicodeCharacter{2019}{\textquoteright}
\DeclareUnicodeCharacter{201C}{\textquotedblleft}
\DeclareUnicodeCharacter{201D}{\textquotedblright}
\DeclareUnicodeCharacter{2013}{\textendash}
\DeclareUnicodeCharacter{2014}{\textemdash}
\DeclareUnicodeCharacter{2026}{\ldots}
\DeclareUnicodeCharacter{2190}{\ensuremath{\leftarrow}}
\DeclareUnicodeCharacter{2191}{\ensuremath{\uparrow}}
\DeclareUnicodeCharacter{2192}{\ensuremath{\rightarrow}}
\DeclareUnicodeCharacter{2193}{\ensuremath{\downarrow}}
\DeclareUnicodeCharacter{21A6}{\ensuremath{\mapsto}}
\DeclareUnicodeCharacter{21D2}{\ensuremath{\Rightarrow}}
\DeclareUnicodeCharacter{21D4}{\ensuremath{\Leftrightarrow}}
\DeclareUnicodeCharacter{2205}{\ensuremath{\emptyset}}
\DeclareUnicodeCharacter{2209}{\ensuremath{\notin}}
\DeclareUnicodeCharacter{2211}{\ensuremath{\sum}}
\DeclareUnicodeCharacter{2212}{\ensuremath{-}}
\DeclareUnicodeCharacter{2218}{\ensuremath{\circ}}
\DeclareUnicodeCharacter{221A}{\ensuremath{\surd}}
\DeclareUnicodeCharacter{2228}{\ensuremath{\vee}}
\DeclareUnicodeCharacter{2229}{\ensuremath{\cap}}
\DeclareUnicodeCharacter{222A}{\ensuremath{\cup}}
\DeclareUnicodeCharacter{2237}{\ensuremath{::}}
\DeclareUnicodeCharacter{2248}{\ensuremath{\approx}}
\DeclareUnicodeCharacter{2261}{\ensuremath{\equiv}}
\DeclareUnicodeCharacter{2265}{\ensuremath{\ge}}
\DeclareUnicodeCharacter{2282}{\ensuremath{\subset}}
\DeclareUnicodeCharacter{2286}{\ensuremath{\subseteq}}
\DeclareUnicodeCharacter{22A2}{\ensuremath{\vdash}}
\DeclareUnicodeCharacter{22A5}{\ensuremath{\bot}}
\DeclareUnicodeCharacter{2308}{\ensuremath{\lceil}}
\DeclareUnicodeCharacter{2309}{\ensuremath{\rceil}}
\DeclareUnicodeCharacter{230A}{\ensuremath{\lfloor}}
\DeclareUnicodeCharacter{230B}{\ensuremath{\rfloor}}
\DeclareUnicodeCharacter{2500}{\textendash}
\DeclareUnicodeCharacter{2502}{\textbar}
\DeclareUnicodeCharacter{2514}{\textendash}
\DeclareUnicodeCharacter{251C}{\textbar}
\DeclareUnicodeCharacter{2550}{\textendash}
\DeclareUnicodeCharacter{25B6}{\ensuremath{\blacktriangleright}}
\DeclareUnicodeCharacter{2713}{\ensuremath{\checkmark}}
\DeclareUnicodeCharacter{2717}{\ensuremath{\times}}
\DeclareUnicodeCharacter{27E8}{\ensuremath{\langle}}
\DeclareUnicodeCharacter{27E9}{\ensuremath{\rangle}}
\DeclareUnicodeCharacter{2115}{\ensuremath{\mathbb{N}}}
\DeclareUnicodeCharacter{2124}{\ensuremath{\mathbb{Z}}}
\DeclareUnicodeCharacter{211A}{\ensuremath{\mathbb{Q}}}
\DeclareUnicodeCharacter{211D}{\ensuremath{\mathbb{R}}}
\DeclareUnicodeCharacter{03B1}{\ensuremath{\alpha}}
\DeclareUnicodeCharacter{03B2}{\ensuremath{\beta}}
\DeclareUnicodeCharacter{03B3}{\ensuremath{\gamma}}
\DeclareUnicodeCharacter{03B4}{\ensuremath{\delta}}
\DeclareUnicodeCharacter{03B5}{\ensuremath{\varepsilon}}
\DeclareUnicodeCharacter{03BB}{\ensuremath{\lambda}}
\DeclareUnicodeCharacter{03BC}{\ensuremath{\mu}}
\DeclareUnicodeCharacter{03C0}{\ensuremath{\pi}}
\DeclareUnicodeCharacter{03C3}{\ensuremath{\sigma}}
\DeclareUnicodeCharacter{03C6}{\ensuremath{\varphi}}
\DeclareUnicodeCharacter{03C8}{\ensuremath{\psi}}
\DeclareUnicodeCharacter{2080}{\ensuremath{_0}}
\DeclareUnicodeCharacter{2081}{\ensuremath{_1}}
\DeclareUnicodeCharacter{2082}{\ensuremath{_2}}
\DeclareUnicodeCharacter{2099}{\ensuremath{_n}}
\DeclareUnicodeCharacter{2C7C}{\ensuremath{_j}}

\title{SWE-Proof: Can Language Models Resolve Real-World Issues with Machine-Checked Proofs?}

\author{%
\begin{minipage}{\dimexpr\textwidth-2\tabcolsep\relax}
\centering\normalfont
George Ma\textsuperscript{1,}\thanks{Equal contribution.}\quad
Benjamin Mikek\textsuperscript{2,}\footnotemark[1]\quad
Haoyu Li\textsuperscript{3}\quad
Ferhat Erata\textsuperscript{4}\\[0.5ex]
Yuhao Zhang\textsuperscript{4}\quad
Zeren Shui\textsuperscript{4}\quad
Behrooz Omidvar Tehrani\textsuperscript{4}\quad
Jun Huan\textsuperscript{4}\\[0.5ex]
Murali Krishna Ramanathan\textsuperscript{4}\quad
Somayeh Sojoudi\textsuperscript{1}\quad
Hao Zhou\textsuperscript{4}\quad
Anoop Deoras\textsuperscript{4}\\[1.2ex]
{\small
\textsuperscript{1}UC Berkeley\quad
\textsuperscript{2}Georgia Tech\quad
\textsuperscript{3}UIUC\quad
\textsuperscript{4}AWS AI Labs}
\end{minipage}%
}

\iclrfinalcopy

\begin{document}

\maketitle
\lhead{Preprint. Under review.}

\doparttoc
\faketableofcontents

\begin{abstract}
Ensuring the correctness of LLM-generated code is a core challenge for modern software engineering.
Benchmarks for agentic code generation check correctness with held-out test suites, which are inherently incomplete and increasingly susceptible to memorization.
Formal verification avoids both problems, but existing work covers only standalone tasks whose specifications are given as input, not real issues, which touch large repositories and state intent in vague natural language.
We present \benchproofer, a pipeline that turns a coding task with a known correct patch into a formally verified one: it writes a specification for the new code, summarizes the existing functions that code calls with axioms, and admits an instance only after mechanical and adversarial gates agree.
Applying it to \sbv yields \benchmark, $500$ real issues whose correctness is formally verified rather than tested, and it extends to \swepro.
Evaluating Claude Opus~4.8, we find that verification catches what tests miss: nearly a third of test-passing patches admit counterexamples, which a structured natural-language specification does not fix, while a correct formal one lifts resolution from $85\%$ to $95\%$.
Writing that specification is the hard part: an agent that must write its own gains nothing over an unaided baseline, and only $56\%$ of those specifications pass our audit.
The usual failure is faithfulness, where a specification constrains part of the required behavior and leaves the rest free. Specification quality also tracks the outcome, failing on 92\% of unresolved instances against 51\% of resolved ones, making faithful specification synthesis a concrete open problem.
Code is available at \url{https://github.com/amazon-science/BenchProofer}.
\end{abstract}

\section{Introduction}
The dominant paradigm for evaluating LLM-based code generation agents is execution against a held-out test suite: an agent is given a programming task and its output is checked against hidden test cases. Testing is cheap, easy to implement, and reuses the tests already present in existing codebases. It is the standard correctness metric for coding agent benchmarks~\citep{jimenez2024swebench,openai2024swebenchverified}, and has driven rapid progress over the last two years~\citep{yang2024sweagent, xia2024agentless,wang2025openhands,zhang2024autocoderover}.

Tests are nevertheless a weak correctness criterion. First, testing is inherently incomplete: a finite set of inputs cannot check behavior on the ones it leaves out. Audits of \swebench have found that many patches credited as correct pass only because the associated tests are too weak to separate a real fix from a superficial one~\citep{aleithan2024swebenchplus}. Second, agents can reward hack, exploiting that incompleteness to produce code that passes the given tests but is not general. \citet{zhong2025impossiblebench} find test-passing but incorrect solutions on a large proportion of \swebench and other benchmark problems.

Formal verification addresses both problems~\citep{ye2025verina,thakur2025clever,loughridge2025dafnybench}. Existing approaches hand an agent a specification in a formal language and ask it for an implementation annotated with invariants and pre- and post-conditions, checked by a verifier such as Dafny~\citep{leino2010dafny} or Verus~\citep{lattuada2023verus,yang2025autoverus,chen2025safe}. Verification certifies the program on its entire specified domain rather than on sampled inputs.

A gap remains between these approaches and realistic coding tasks. Verified generation has so far targeted small curated exercises~\citep{loughridge2025dafnybench}, standalone HumanEval-style tasks~\citep{shefer2025mainstream}, and competition mathematics~\citep{wu2022autoformalization}, and most techniques take the formal specification as input~\citep{ye2025verina,thakur2025clever}. Real programming tasks instead require new code to fit into large codebases with many files and dependencies, and state intent in vague natural language. Neither existing approaches to verified code generation \textit{nor the benchmarks used to evaluate them} apply a sound verification oracle to such tasks.

We close this gap with \benchmark, a benchmark that pairs realistic programming tasks with formal verification. \benchmark provides formal specifications, implementations, and proofs of correctness for each of the $500$ natural language coding tasks in \sbv~\citep{openai2024swebenchverified}, under three verification backends: the \textsc{Nagini} verifier for statically-typed Python~\citep{eilers2018nagini,muller2016viper}, \textsc{Velvet}, a DSL for imperative-style program verification in \textsc{Lean}~\citep{gladshtein2026velvet}, and pure \textsc{Lean}~\citep{moura2015lean}, in which proof obligations are discharged by a proof-writing agent rather than by an SMT solver.

We build \benchmark with \benchproofer, a construction pipeline that turns a coding task with a known ground truth solution into a formally verified one. It rests on two ideas. First, it keeps verification tractable by modeling only the new or modified code and summarizing its calls into unchanged functions with \emph{axioms}, overapproximative summaries of the callee properties the fix relies on. Second, it checks the soundness and completeness of generated formal specifications with a suite of correctness gates, combining mechanical verification with adversarial LLM auditors. While \benchproofer's gates cannot resolve the ambiguity inherent in natural language, our methodology uses all available resources to produce high-fidelity formal representations of natural language specifications and checks them against an input benchmark's ground truth code patch. We release \benchproofer with \benchmark, and show it generalizes by constructing a verified version of \swepro.

We then evaluate a frontier model, Claude Opus~4.8~\citep{anthropic2025opus}, and report five findings. Nearly a third of patches that pass every hidden test are still flawed, and a structured natural language ``specification'' does not close that gap. Asking an agent to write a specification and verify against it does not improve resolution over an unaided baseline, but supplying a correct one raises it from $85.0\%$ to $95.1\%$. Specification synthesis is the bottleneck: only $56\%$ of synthesized specifications survive our adversarial audit, and the dominant failure is faithfulness, a specification that constrains part of the required behavior and leaves the rest free. That quality tracks resolution: $92\%$ of specifications from unresolved instances fail the audit, against $51\%$ from resolved ones.

In summary, we contribute \benchproofer, a construction pipeline that turns an agentic coding task with a known ground truth solution into ground truth specifications, implementations, and proofs, using environment axiomatization and a sequence of mechanical and adversarial gates (\cref{sec:method}); \benchmark, to our knowledge the first benchmark to attach a verification oracle to real-world software engineering tasks, covering all $500$ instances of \sbv and extended to the Python fragment of \swepro; and an evaluation that identifies specification synthesis as the key unmet challenge for verified code generation (\cref{sec:experiments}).

\section{Overview}
\label{sec:overview}

\begin{figure*}[t]
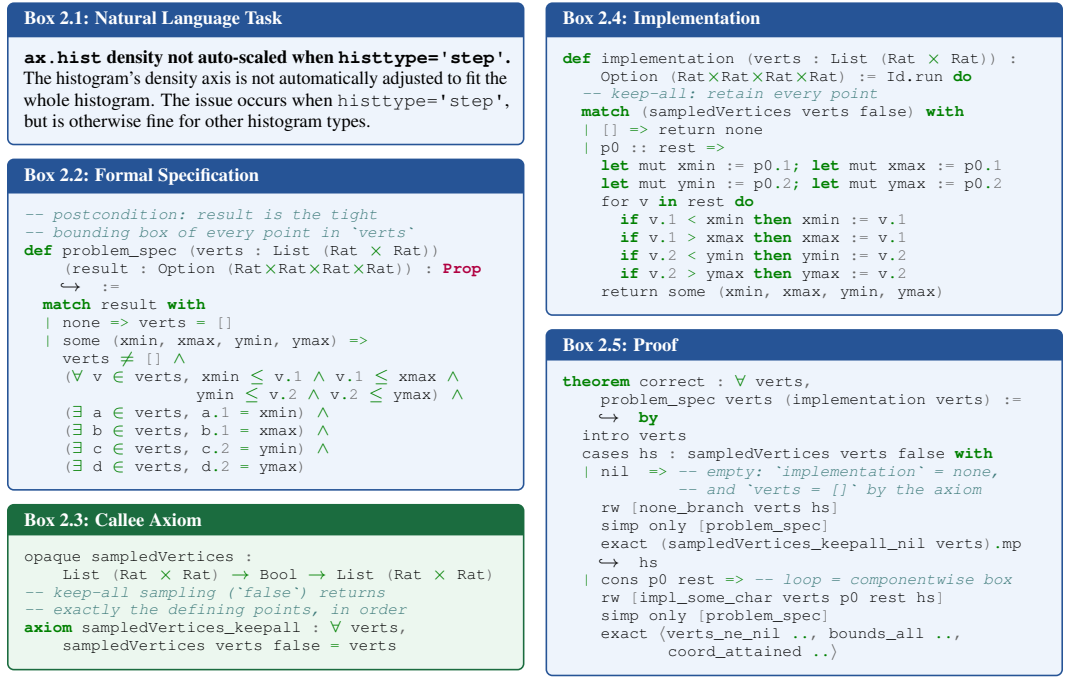

\begin{minipage}[t]{0.49\textwidth}
\vspace{0pt}
\begin{exbox}[unbreakable,label=box:mot-nl]{Natural Language Task}
\scriptsize
\textbf{\texttt{ax.hist} density not auto-scaled when \texttt{histtype=\textquotesingle step\textquotesingle}.}

The histogram's density axis is not automatically adjusted to fit the whole histogram. The issue occurs when \texttt{histtype=\textquotesingle step\textquotesingle}, but is otherwise fine for other histogram types.
\end{exbox}

\begin{specbox}[unbreakable,label=box:mot-spec]{Formal Specification}
\begin{minted}[fontsize=\tiny,escapeinside=]{lean}
-- postcondition: result is the tight
-- bounding box of every point in `verts`
def problem_spec (verts : List (Rat × Rat))
    (result : Option (Rat×Rat×Rat×Rat)) : Prop :=
  match result with
  | none => verts = []
  | some (xmin, xmax, ymin, ymax) =>
    verts ≠ [] ∧
    (∀ v ∈ verts, xmin ≤ v.1 ∧ v.1 ≤ xmax ∧
                  ymin ≤ v.2 ∧ v.2 ≤ ymax) ∧
    (∃ a ∈ verts, a.1 = xmin) ∧
    (∃ b ∈ verts, b.1 = xmax) ∧
    (∃ c ∈ verts, c.2 = ymin) ∧
    (∃ d ∈ verts, d.2 = ymax)
\end{minted}
\end{specbox}

\begin{axiombox}[unbreakable,label=box:mot-axiom]{Callee Axiom}
\begin{minted}[fontsize=\tiny,escapeinside=]{lean}
opaque sampledVertices :
    List (Rat × Rat) → Bool → List (Rat × Rat)
-- keep-all sampling (`false`) returns
-- exactly the defining points, in order
axiom sampledVertices_keepall : ∀ verts,
    sampledVertices verts false = verts
\end{minted}
\end{axiombox}
\end{minipage}\hfill
\begin{minipage}[t]{0.49\textwidth}
\vspace{0pt}
\begin{specbox}[unbreakable,label=box:mot-impl]{Implementation}
\begin{minted}[fontsize=\tiny,escapeinside=]{lean}
def implementation (verts : List (Rat × Rat)) :
    Option (Rat×Rat×Rat×Rat) := Id.run do
  -- keep-all: retain every point
  match (sampledVertices verts false) with
  | [] => return none
  | p0 :: rest =>
    let mut xmin := p0.1; let mut xmax := p0.1
    let mut ymin := p0.2; let mut ymax := p0.2
    for v in rest do
      if v.1 < xmin then xmin := v.1
      if v.1 > xmax then xmax := v.1
      if v.2 < ymin then ymin := v.2
      if v.2 > ymax then ymax := v.2
    return some (xmin, xmax, ymin, ymax)
\end{minted}
\end{specbox}

\begin{specbox}[unbreakable,label=box:mot-proof]{Proof}
\begin{minted}[fontsize=\tiny,escapeinside=]{lean}
theorem correct : ∀ verts,
    problem_spec verts (implementation verts) := by
  intro verts
  cases hs : sampledVertices verts false with
  | nil  => -- empty: `implementation` = none,
            -- and `verts = []` by the axiom
    rw [none_branch verts hs]
    simp only [problem_spec]
    exact (sampledVertices_keepall_nil verts).mp hs
  | cons p0 rest => -- loop = componentwise box
    rw [impl_some_char verts p0 rest hs]
    simp only [problem_spec]
    exact ⟨verts_ne_nil .., bounds_all ..,
           coord_attained ..⟩
\end{minted}
\end{specbox}
\end{minipage}
\caption{The \benchmark artifacts for \id{matplotlib-24177} with \textsc{Lean} as the backend. Box~\ref{box:mot-nl} is taken from \sbv; all other artifacts are newly-constructed for \benchmark.}
\Description{Five boxes laying out the artifacts of one benchmark instance. The left column shows the natural-language bug report taken from SWE-bench Verified, a formal specification in \textsc{Lean} stating that the result is the tight bounding box of every input vertex, and a callee axiom declaring that keep-all sampling returns exactly the defining vertices in order. The right column shows the \textsc{Lean} implementation, which loops over the vertices computing component-wise minima and maxima, and an excerpt of the \textsc{Lean} proof that the implementation satisfies the specification using the axiom as a lemma.}
\label{fig:motivating}
\end{figure*}

We illustrate \benchmark with one instance, \id{matplotlib-24177} from \sbv~\citep{jimenez2024swebench}, whose artifacts are shown in \cref{fig:motivating}.

\paragraph{Task and specification}
The input is a natural language bug report (Box~\ref{box:mot-nl}): the bounds of a generated plot are computed from a simplified vertex set, so parts of the plot are cut off. \benchmark adds a formal specification of the intended behavior (Box~\ref{box:mot-spec}), here stating that each side of the bounding box is tightly aligned with the furthest outlying point in \id{verts}.

\paragraph{Axioms}
Generating a specification and proof for every function the fix calls, and their transitive closure, is infeasible. We therefore axiomatize each \emph{unchanged callee}, a function the new code calls but does not modify. Box~\ref{box:mot-axiom} summarizes \id{sampledVertices}, asserting only that its keep-all mode returns exactly the defining points (\id{verts}), in order.

\paragraph{Implementations and patch}
A verifier can only check code in its backend's language, so each instance carries the fix in that language (Box~\ref{box:mot-impl}), here a \textsc{Lean} loop that widens the bounds as it walks the vertices. It carries two implementations: a \emph{reference implementation}, which must verify, and a \emph{pre-fix implementation}, which must fail to verify and so witnesses that the specification rules the bug out. Since the hidden tests run against the repository, each instance also carries an \emph{equivalent patch}, the Python diff making the same change; \cref{claim:verify-resolve} guarantees that the equivalent patch of a verified implementation resolves the issue.

\paragraph{Proof}
Box~\ref{box:mot-proof} excerpts the \textsc{Lean} proof that the implementation satisfies the specification, using the axiom of Box~\ref{box:mot-axiom} as a lemma. Once the compiler accepts it, the proof certifies the implementation subject to the axioms. \cref{sec:construction} describes how we synthesize and validate all of these artifacts.

\paragraph{Backends and challenges}
\benchmark supports three verification backends. Under \textsc{Lean} the proof is agent-written and compiler-checked, while \textsc{Nagini} and \textsc{Velvet} discharge the obligations with an SMT solver, which is faster but offers no recourse when the solver times out; we compare the trade-offs in Appendix~\ref{app:backends}. The key challenge is ensuring these artifacts are correct and free of reward hacking and hallucination, which is what the gates of \cref{sec:method} check.

\section{Methodology}
\label{sec:method}

\begin{figure*}[t]
\centering
\includegraphics[width=\textwidth]{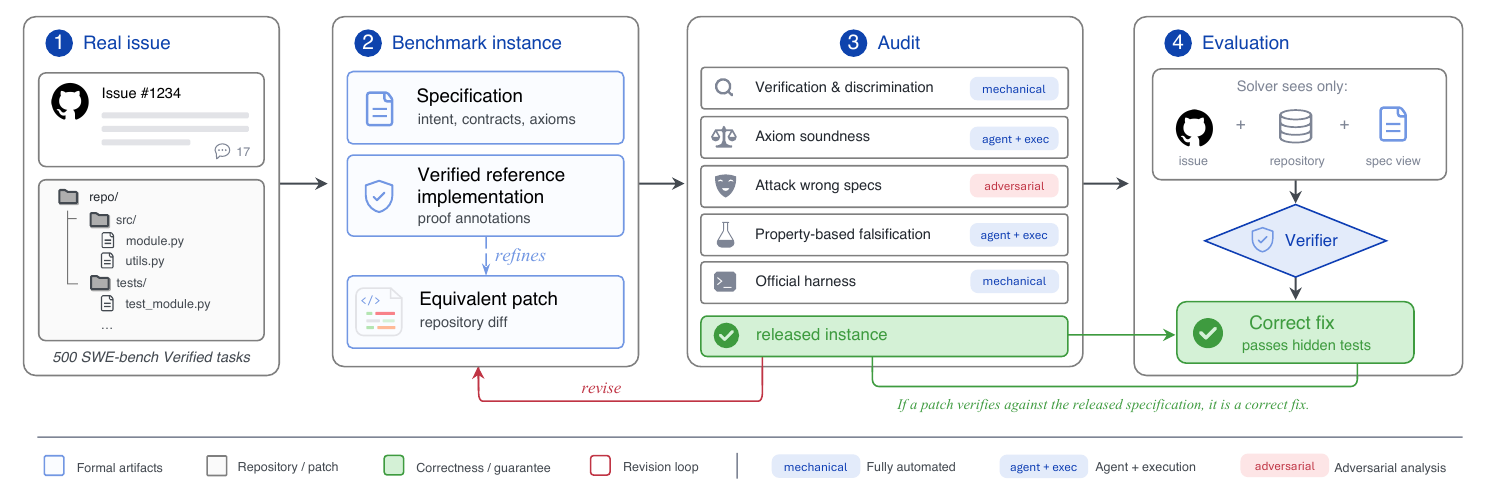}
\caption{The \benchproofer workflow. From a natural language issue description, \benchproofer builds a specification, an implementation, and a verification certificate, and admits each artifact only after it passes a suite of gates that combine mechanical checks with adversarial audits.}
\Description{A left-to-right workflow diagram. A real GitHub issue and repository are turned by agentic construction into three linked artifacts: a specification with axioms, a verified reference implementation, and an equivalent patch. Every instance must pass a stack of audit gates: mechanical verification and discrimination, axiom soundness and boundary checking, adversarial soundness and completeness attacks, property-based falsification with a released test suite, and resolution in the official harness. Passing gates yield the guarantee verify implies resolve. At evaluation, the solver sees only the specification view and a verifying implementation is a correct fix.}
\label{fig:workflow}
\vspace{-3mm}
\end{figure*}

\cref{fig:workflow} gives an overview of \benchproofer, the construction pipeline we use to generate the ground truth artifacts for \benchmark. 

\subsection{The Guarantee}
\label{sec:claim}

The goal of benchmark construction is to produce ground truth artifacts against which candidate implementations can be formally verified. To make these artifacts useful, an implementation that verifies against the specification should resolve the underlying task; we formally state this property as Claim~\ref{claim:verify-resolve}.

\begin{claim}[Verify $\Rightarrow$ Resolve]
\label{claim:verify-resolve}
Consider an instance of \benchmark with specification $S$ and axioms $A$. Let $I$ be an implementation written in the language of the instance's verification backend (\textsc{Nagini}, \textsc{Velvet}, or \textsc{Lean}), and let $P$ be a Python patch that is equivalent to $I$ in the repository (\cref{sec:boundaries}). If $I$ verifies against $S$ under $A$, then $P$ resolves the instance by passing the hidden tests in \swebench.
\end{claim}

Verification buys more than the tests can: a verified $I$ is correct on \emph{every input} $S$ admits, including the ones no test exercises. The equivalence of $I$ and $P$ is a hypothesis of the claim rather than a consequence of it. An agent evaluated on \benchmark supplies its own equivalent patch, and we report how we check it in \cref{sec:experiments}; for the reference implementation we release, the equivalence gate of \cref{sec:correctness} audits it against the ground-truth patch.

\subsection{Benchmark Construction}
\label{sec:construction}

\paragraph{Specification synthesis.}
Realistic tasks state intent in natural language that is often vague, or open to several formal readings. An agent turns that description into a formal specification and refines it in a feedback loop. During construction, though never during evaluation (\cref{sec:experiments}), it also sees the ground-truth patch, the buggy code it replaces, the test patch, and the instance's \texttt{FAIL\_TO\_PASS} (F2P) and \texttt{PASS\_TO\_PASS} lists, and calibrates the specification's strength by asking whether an implementation could satisfy it and still fail an F2P test. Rather than solve the issue from scratch, the agent renders the patched and the buggy code into the backend's language, which yields the reference and pre-fix implementations of \cref{sec:overview}. \cref{tab:components} lists the components of an instance and shows which construction step and which gate touches each.

\paragraph{Environment axiomatization.}
Verifying every function the new code calls, together with their transitive closure, quickly exceeds the limits of existing automated verifiers. Moreover, some callees are library routines whose source code is unavailable. We therefore summarize each unchanged callee with an axiom stating the properties the fix relies on. Axioms are the boundary between what our approach verifies and what it assumes.

\benchproofer produces and validates them with an agentic procedure. It first collects the unchanged callees $f_1, f_2, \ldots$ that the new implementation invokes, from the repository and from libraries alike, and for each $f_i$ builds a fuzzer that mixes random inputs with cases an agent writes. An agent then drafts a candidate axiom $A_i$, which two opposing loops refine. The first checks \emph{correctness}: the fuzzer exercises $f_i$, and any input that violates $A_i$ sends the axiom back for revision. The second checks \emph{usefulness}: the verifier tries to close the proof relating the reference implementation to the specification using $A_i$ as a lemma, and failure sends it back as well. The loops repeat until the axiom is both consistent with observed behavior and strong enough to finish the proof. Unlike the specification, an axiom need not be precise: it may over-approximate $f_i$, constraining less than the callee really guarantees, and still yield a correct instance, as long as the proof closes.

\begin{table}[t]
\centering
\caption{The components that each construction step and each correctness gate relates. Columns are the components of one instance: a) the original \sbv artifacts, whose issue description and human-written tests together express the intent, and b) the newly-constructed \benchmark artifacts. A \cmark{} marks each component the step or gate touches; a gate spanning multiple components audits their correspondence jointly.}
\label{tab:components}
\small
\setlength{\tabcolsep}{4pt}
\resizebox{\columnwidth}{!}{%
\begin{tabular}{@{}llcccccccc@{}}
\toprule
& & \multicolumn{3}{c}{\textbf{\sbv}} & \multicolumn{5}{c}{\textbf{\benchmark}} \\
\cmidrule(lr){3-5}\cmidrule(lr){6-10}
& \textbf{Stage}
& \rot{Issue $+$ tests} & \rot{Unchanged code} & \rot{Ground-truth patch}
& \rot{Specification} & \rot{Axioms}
& \rot{Reference impl.} & \rot{Pre-fix impl.} & \rot{Proof} \\
\midrule
\multicolumn{10}{@{}l}{\textit{Construction} (\cref{sec:construction})}\\
\quad Specification synthesis   & constr.\ & \cmark & & \cmark & \cmark & \cmark & \cmark & \cmark & \cmark \\
\quad Environment axiomatization & constr.\ & & \cmark & & & \cmark & & & \\
\midrule
\multicolumn{10}{@{}l}{\textit{Correctness gates} (\cref{sec:correctness})}\\
\quad Verification            & mech.\ & & & & \cmark & \cmark & \cmark &        & \cmark \\
\quad Discrimination          & mech.\ & & & & \cmark &        &        & \cmark &        \\
\quad Mutation                & mech.\ & & & & \cmark &        & \cmark &        &        \\
\quad Hygiene                 & mech.\ & & & & \cmark & \cmark & \cmark &        & \cmark \\
\quad Axiom soundness         & adv.\ & & \cmark & &        & \cmark &        &        &        \\
\quad Conformance             & adv.\ & & \cmark & \cmark &        &        & \cmark &        &        \\
\quad Soundness \& completeness & adv.\ & & & \cmark & \cmark & & \cmark & \cmark &        \\
\quad Property-based falsification & adv.\ & \cmark & & \cmark & \cmark & & \cmark & \cmark &        \\
\quad Equivalence             & adv.\ & & & \cmark &        &        & \cmark &        &        \\
\bottomrule
\end{tabular}%
}
\end{table}

\subsection{Correctness Gates}
\label{sec:correctness}

No single check establishes \cref{claim:verify-resolve} end to end, so \benchproofer secures it with a suite of \emph{gates}. Once specification synthesis and environment axiomatization produce a candidate instance, it must pass every gate before we accept it, and any counterexample sends it back for revision. The gates are of two kinds: \emph{mechanical} checks, such as running an SMT solver or the \textsc{Lean} compiler, and \emph{adversarial} gates, in which LLM auditors try to refute the instance with a concrete counterexample. \cref{tab:components} lists the components each gate relates.

\paragraph{Mechanical gates.}
The \emph{verification check} confirms that the reference implementation verifies under the instance's backend; under pure \textsc{Lean} this means the agent-written proof is accepted by the kernel. The \emph{discrimination check} pairs the same specification with the pre-fix implementation and confirms that it fails to verify, which establishes that the specification is strong enough to catch the defect. The \emph{mutation check} extends this to first-order mutations of the reference implementation, one operator at a time, and requires that the variants no longer verify; a surviving mutant means the specification misses a local perturbation, so we require a high kill rate. The \emph{hygiene check} confirms that the instance uses only what the verifier can check soundly: imports come from an allowlist, the specification is well formed, and the proof uses no escape hatch such as \textsc{Lean}'s \texttt{sorry}. The \emph{resolution check} applies the ground-truth patch in the official Docker harness and confirms that the instance resolves. It tests the original benchmark rather than anything we build, which is what makes the pipeline safe to point at a new benchmark.

\paragraph{Axiom-soundness gate.}
This gate re-runs the fuzzing check of \cref{sec:construction} with independent adversarial auditors, adjudicating their inputs by executing the real callee inside the instance's Docker image, so a single observed violation rejects the axiom. It also confirms that each axiomatized function is genuinely unchanged rather than one the fix modifies. The resulting axiom tests ship as re-runnable evidence.

\paragraph{Conformance gate.}
This gate checks that the reference implementation agrees with the repository code it stands for. It runs an executable \emph{shadow} of the implementation against the real patched repository function inside the instance's container, on at least $10^5$ generated inputs, and reports any disagreement. Under \textsc{Nagini} the shadow is the implementation with its annotations stripped, so nothing is translated; under \textsc{Velvet} and \textsc{Lean}, whose languages cannot be executed, it is an agent-written Python translation, which we treat as the weakest link in the guarantee.

\paragraph{Soundness and completeness gates.}
A panel of independent adversarial auditors attacks the specification from two directions. The \emph{soundness} attack tries to show that the specification is too strong, by hunting for an input on which the specification and the repository code disagree; such an input means the specification has drifted from the behavior the repository actually has, so correct code need not verify. The \emph{completeness} attack tries to show that it is too weak, by building an incorrect implementation that verifies anyway, for instance one that satisfies a postcondition without doing the real work. Together the two require that verification be achievable by correct code and unachievable by incorrect code.

\paragraph{Property-based falsification.}
This gate makes those two attacks reproducible. For each modeled function, adversarial auditors write a \emph{property-based test suite}: implementations labeled with the verdict they should receive against the frozen specification, correct ones to verify and buggy ones to be rejected. Each case is adjudicated by re-running the verifier, so a disagreement with a label puts the specification at fault. The suites ship with the benchmark as reusable evidence.

\paragraph{Equivalence gate.}
This gate audits the correspondence between the reference implementation and the ground-truth patch. The auditor walks the documented mapping and confirms that each step is one of three behavior-preserving operations: a \emph{type refinement}, which replaces a repository type with one the verifier can express, such as a matrix row as a sequence of integers; an \emph{axiomatized callee}, which replaces a call to an unchanged function with the axiom summarizing it; or an \emph{identical operation}, which reproduces the patch verbatim. Any other step is a divergence and sends the instance back, for example an axiom standing in for a callee the patch modifies.

\paragraph{Leakage gate.}
A final gate guards a secondary property: the \emph{specification view}, the specification with its axioms, is what an agent sees at evaluation time, and it must reveal nothing about the fix beyond what the issue already discloses. A mechanical screen looks for views exposing patched identifiers, file paths, or narration of the fix, and an adversarial auditor then tries to reconstruct the ground-truth patch from the view alone. If either succeeds the instance is rejected, since a leaked view would hand the answer to an agent being evaluated on \benchmark.

\subsection{Boundaries of the Guarantee}
\label{sec:boundaries}

Three boundaries limit where a true correspondence between artifacts can be established. The first is the intent-specification boundary: no guarantee can be established that a specification matches informal natural language intent. The intent is supplied to construction as two inputs, the issue description and the human-written \sbv tests, together with the ground-truth patch that implements them (\cref{tab:components}). The property-based falsification gate of \cref{sec:correctness} is what guards the correspondence between the specification and that intent.

The remaining two are the harder ones, and they share a cause: each backend verifies code in its own language while the repository is Python, so a formal artifact and the code it stands for can never be the same text. The repository side has two parts, the unchanged code and the patch, and each is paired with a formal artifact. No formal guarantee can establish the correspondence within either pair, so we close both with the adversarial gates of \cref{sec:correctness}. The second boundary pairs an axiom with the unchanged callee it summarizes, and rests on the axiom-soundness gate together with the fuzzing procedure of \cref{sec:construction}. The third pairs the reference implementation with the patch; several gates bear on it, but the conformance and equivalence gates address it directly.

Each instance ships additional artifacts that can be used to reproduce the checking of the instance: the axiom tests and the property suite the gates produced, the operation-by-operation correspondence between the reference implementation and the patch, a record of which repository entity each modeled symbol abstracts, and the outcome of every gate. Appendix~\ref{app:structure} documents them and Appendix~\ref{app:examples} walks through one instance, while \cref{sec:discussion} covers the remaining trust assumptions.

\subsection{Extensibility}
\label{sec:extensibility}

To see how far \benchproofer generalizes, we applied it to \swepro, a second set of natural language GitHub issues whose tasks are considerably larger than \sbv's: they modify or add an average of nine functions and $146$ lines, against two functions and $14$ lines. \benchproofer built ground truth artifacts that pass every correctness gate under every backend for $242$ of the $266$ Python tasks.

The $22$ tasks that fail under all backends fail for one reason: the patch does not alter behavior that pre- and post-condition specifications can observe. Most only rename or relocate code, and the rest change signatures, internal data structures, or side effects outside the value domain. The remaining two tasks are blocked by backend-specific limits and build successfully elsewhere. Appendix~\ref{app:pro-nongreen} diagnoses each case, and Appendix~\ref{app:pro-results} reports the \swepro evaluation across the full grid of settings. \benchproofer therefore applies to any task with a known ground truth patch whose change an existing verifier can model.

\section{Experiments}
\label{sec:experiments}

Our experiments ask what verification is worth. \cref{sec:res-tests} asks whether the hidden tests are complete and, where they are not, whether a specification has to be formal to close the gap; \cref{sec:res-construct} whether an agent gains from writing and verifying its own specification, and \cref{sec:res-help} whether being handed a correct one helps. \cref{sec:res-synth} and \cref{sec:res-impact} then ask what goes wrong when models write specifications themselves.

\subsection{Evaluation protocol}
\label{sec:protocol}

\paragraph{Settings.}
All settings share one agent scaffold, a shell and a submission action following~\citet{yang2024sweagent}, and differ only in what the agent is given and in what a pass requires; \cref{tab:main} lists both. The agent can run the in-repository tests but never sees the hidden ones, and a sanitization step strips the version-control history that would leak the fix.

\paragraph{Metrics.}
Rows scored on tests alone report the \emph{resolution rate}, the fraction of instances whose patch passes the hidden tests under the official harness. Rows~3, 5, and~7 are scored jointly: an instance passes only when the patch resolves, the implementation verifies against the specification, and a majority of three adversarial auditors agree that the implementation corresponds to the patch.

\paragraph{Models and verifiers.}
We evaluate Claude Opus~4.8 \citep{anthropic2025opus} under \textsc{Nagini}, \textsc{Velvet}, and \textsc{Lean} under identical scaffolds, prompts, and budgets. Every number is over the full $500$ instances.

\begin{table}[t]
\centering
\caption{Evaluation settings and results for Claude Opus~4.8 over all $500$ \sbv instances (\%).
\textbf{Provided} is what the agent is given: the \emph{verifier} as a callable
tool, \emph{edit localization} (which functions and files the fix touches), and
the ground-truth \emph{specification} view. \textbf{Evaluated} is what a pass
requires. \emph{Tests} means the submitted patch resolves the issue under the
official \swebench harness. \emph{Verify} means the submitted
implementation verifies against a specification, the agent's own in rows~3
and~5 or the ground-truth one in row~7. Rows~3, 5, and~7 further require a judge to confirm
that the verified implementation and the submitted patch agree. Rows~0, 1, and~4
supply neither a specification nor a verifier, so a single run serves every
backend (gray rules).
Two marks carry extra meaning. In row~1, $\checkmark^*$ means the row~0 patches
were re-scored by an adversarial audit that looks for inputs on which the agent-generated patch differ from the ground-truth one, and an instance passes when no such input
is found. In rows~6 and~7, $\checkmark^\dagger$ means the provided specification
already reveals the localization.
Finally, EARS is natural language and has no verifier, so in the EARS column
\emph{Verify} is that same adversarial audit, and in rows~3, 5, and~7 the EARS agent works
without a verifier tool.}
\label{tab:main}
\setlength{\tabcolsep}{2.5pt}
\begin{adjustbox}{max width=\textwidth}

\begin{tabular}{@{}cp{4.6cm}ccccccccc@{}}
\toprule
& & \multicolumn{3}{c}{Provided} & \multicolumn{2}{c}{Evaluated} & \multicolumn{4}{c}{Pass rate (\%)} \\
\cmidrule(lr){3-5}\cmidrule(lr){6-7}\cmidrule(lr){8-11}
\# & Setting & Verifier & Local. & Spec. & Tests & Verify & \textsc{Nagini} & \textsc{Velvet} & \textsc{Lean} & EARS \\
\midrule
0 & Unaided baseline        & ---        & ---        & ---        & \checkmark & ---        & \multicolumn{4}{c}{\modefill\enspace 85.0 \enspace\modefill} \\
1 & Baseline + Audit        & ---        & ---        & ---        & \checkmark & $\checkmark^*$        & \multicolumn{4}{c}{\modefill\enspace 58.2 \enspace\modefill} \\ \midrule
2 & End-to-End                     & \checkmark & ---        & ---        & \checkmark & ---        & 85.6 & 85.0 & 83.8 & 84.2 \\
3 & Verified End-to-End           & \checkmark & ---        & ---        & \checkmark & \checkmark & 84.0 & 85.4 & 84.6 & 57.2 \\ \midrule
4 & Localization Provided        & ---        & \checkmark & ---        & \checkmark & ---        & \multicolumn{4}{c}{\modefill\enspace 88.2 \enspace\modefill} \\
5 & Verified from Localization               & \checkmark & \checkmark & ---        & \checkmark & \checkmark & 92.0 & 91.8 & 89.6 & 64.6 \\ \midrule
6 & Specification Provided       & \checkmark & $\checkmark^\dagger$ & \checkmark & \checkmark & ---        & 96.2 & 94.0 & 95.2 & 97.8 \\
7 & Verified from Specification & \checkmark & $\checkmark^\dagger$ & \checkmark & \checkmark & \checkmark & 95.0 & 94.0 & 93.6 & 80.0 \\
\bottomrule
\end{tabular}

\end{adjustbox}
\end{table}

\subsection{Finding 1: Tests are incomplete; only formal specifications close the gap}
\label{sec:res-tests}

Passing the hidden tests is not the same as being equivalent to the ground-truth patch, because the suite is finite: a patch can satisfy every hidden test and still diverge from the ground-truth patch on inputs no test exercises. Rows~0 and~1 measure how often that happens.
For every patch Row~0 accepted, an adversarial auditor writes new tests to distinguish it from the ground-truth patch, and a test counts only if it fails under the agent's patch and passes under the ground-truth one. Such a test is evidence that the patch is wrong even though the suite accepted it.
Under this audit, the overall pass rate drops from 85.0\% to 58.2\%, a decrease of 26.8 percentage points across all 500 instances (Row 0 minus Row 1). Equivalently, 134 of the 425 test-passing patches (31.5\%) are overturned, indicating that the benchmark's test suites are substantially incomplete: \textbf{passing all tests does not establish correctness}.

Rows~6 and~7 ask whether a specification closes that gap, and whether the specification has to be formal. Both rows hand the agent a specification, with the structured EARS format~\citep{ears} as the non-formal control (\cref{app:backend-ears}). Row~7 then checks the result against ground truth: the formal backends verify the implementation against the ground-truth specification, while EARS uses the adversarial auditor above. Both check the same ground-truth specification, and verification is the stronger of the two: it proves that no violating input exists, where the audit only searches for one. First, the formal backends close the gap: averaged over the three, a patch that resolves almost always survives verification as well, losing only $0.9$ points from Row~6 to Row~7. Second, EARS does not close it: $17.8$ points of its pass rate come from patches that remain distinguishable from the ground-truth patch despite passing every hidden test. Rows~2 and~3 show the same split when the agent writes the specification itself, where the formal average loses $0.1$ points against EARS's $27.0$. \textbf{A structured non-formal specification confers none of the improvement that a formal one does.}

\subsection{Finding 2: Self-constructed specifications do not help}
\label{sec:res-construct}

In the end-to-end setting (Rows~2 and~3) the agent is prompted to write a specification and verify an implementation against it before submitting its patch. It gets no more information than the unaided baseline of Row~0. \textbf{Resolution does not improve over the baseline under any backend} (Row~2 against Row~0). The best backend gains $0.6$ points ($85.6\%$ against $85.0\%$, under \textsc{Nagini}); the other two are negative, the worst by $1.2$. Row~3, which also requires the implementation to verify against the self-constructed specification and to match the patch, stays within $1.6$ points of Row~2 under every formal backend, so the bottleneck is not writing the implementation or the proof but the self-constructed specification (\cref{sec:res-synth}).

\subsection{Finding 3: A correct specification helps substantially}
\label{sec:res-help}

\textbf{Handing the agent \benchmark's ground-truth specification raises resolution sharply} (Row~6): from $85.0\%$ to $96.2\%$ under \textsc{Nagini}, $94.0\%$ under \textsc{Velvet}, and $95.2\%$ under \textsc{Lean}. Some of the gain is localization, since the specification reveals which functions are in scope, but most is not: against Row~4, which supplies localization alone, it is still worth $+6.9$ points averaged over the three backends. The gain also survives the stricter scoring of Row~7 (\cref{sec:res-tests}), where the equivalence judges accept the submitted patch as making the same change as the verified implementation on $99.3\%$ of the judged episodes (\cref{tab:equiv}), so these scores are not inflated by patches that diverge from the implementation they were derived from.

The telling comparison is Row~7 against Row~3, where the agent had to write the specification itself (\cref{sec:res-construct}). Being handed a correct one is worth $+11.0$ points under \textsc{Nagini}, $+8.6$ under \textsc{Velvet}, and $+9.0$ under \textsc{Lean}. 
The two settings differ only in whether the ground-truth specification is given or the agent must construct it, so a gap this large shows that
\textbf{producing a correct specification is the bottleneck for verified code generation.}

\subsection{Finding 4: Specification synthesis is the bottleneck, and faithfulness is why}
\label{sec:res-synth}

We score a specification against five properties. Three are criteria proposed by~\cite{feng2026certified}, and constrain the specification of each function. A specification is \emph{admissible} if its precondition excludes no input the corrected code legitimately handles, since an over-strong precondition lets an implementation verify on a shrunken domain. It is \emph{sound} if the intended behavior of each modeled function satisfies its contract, so a correct implementation can verify, and \emph{complete} if no buggy implementation can verify against it.

Two novel properties are specific to repairing code inside a repository. A specification has \emph{sound axioms} if every axiom is true of the real callee and describes only code the fix leaves alone, never behavior the fix introduces. It is \emph{faithful} if the functions it models cover the whole behavioral surface the issue requires. The first three ask whether each modeled function is pinned down correctly; faithfulness asks whether enough of the task was modeled at all. It matters because a real fix usually spans several functions and files, and whatever the specification leaves out stays unconstrained: a specification can be admissible, sound, and complete on the functions it models, yet an implementation that verifies against it still fails the issue by being wrong on the rest.

\begin{wraptable}{r}{0.36\textwidth}
\centering
\vspace{-12pt}
\caption{Specification synthesis by backend (\%). The first row is the share of specifications passing the full five-property audit; the rest are per-property failure rates.}
\label{tab:synth}
\footnotesize
\setlength{\tabcolsep}{4pt}
\begin{adjustbox}{max width=\linewidth}
\begin{tabular}{@{}lccc@{}}
\toprule
& \textbf{\textsc{Nagini}} & \textbf{\textsc{Velvet}} & \textbf{\textsc{Lean}} \\
\midrule
Passes audit & 46.0 & 60.0 & 61.4 \\
\midrule
\multicolumn{4}{@{}l}{\textit{Per-property failure rate}}\\
\quad Axiom soundness & 5.0 & 0.0 & 0.8 \\
\quad Admissibility & 5.0 & 3.6 & 0.8 \\
\quad Soundness & 11.2 & 12.4 & 15.6 \\
\quad Completeness & 5.6 & 2.8 & 3.4 \\
\quad \textbf{Faithfulness} & \textbf{52.2} & \textbf{38.0} & \textbf{37.2} \\
\bottomrule
\end{tabular}
\end{adjustbox}
\vspace{-10pt}
\end{wraptable}

To locate the bottleneck of \cref{sec:res-help}, we isolate specification synthesis: the agent gets only the issue and the repository and must produce a specification and a witness implementation that verifies under it, with no patch. Three adversarial auditors score it against the five properties, running candidate implementations through the verifier and probing the axioms against the real callees; it passes only if a majority accept all five. Pass rates fall far below what the same model reaches when a specification is supplied: $46.0\%$ under \textsc{Nagini}, $60.0\%$ under \textsc{Velvet}, and $61.4\%$ under \textsc{Lean}. The failures are lopsided (\cref{tab:synth}). Sound axioms, admissibility, and completeness almost never fail, each on at most $3.9\%$ of instances, so the specifications do reject the pre-fix implementation and do not shrink the input domain. Soundness fails more often, on $13.1\%$. \emph{Faithfulness} dominates, failing on $42.5\%$ and remaining the most violated property under every backend. \textbf{Models write specifications that are correct on the functions they model but cover too few of them, so faithful specification synthesis is the open problem.}

\subsection{Finding 5: Specification correctness tracks patch correctness}
\label{sec:res-impact}

\cref{sec:res-synth} scores synthesized specifications on their own; here we ask how that correctness relates to resolving the issue. In the end-to-end setting of Rows~2 and~3 the model writes both a specification and a patch, so we re-run the five-property audit on the specification it wrote for itself and split the results by whether the patch resolved. The auditors never see the patch or its test outcome, so their verdict is independent of the result. The two groups separate sharply under every backend (\cref{tab:spec-impact}): averaged over the three, specifications from unresolved instances fail the audit $92.3\%$ of the time, against $50.5\%$ on resolved ones. The gap holds under every backend for four of the five properties, all but axiom soundness, and is widest on the two that govern how much behavior a specification constrains: faithfulness ($42.9$ points on average) and soundness ($26.9$). \textbf{Specification correctness tracks resolution, and faithfulness accounts for most of the gap.}

\begin{table}[htbp]
\centering
\caption{Specification correctness against patch correctness for Opus~4.8, by verification backend (\% of specifications failing each property). The five-property audit of \cref{sec:res-synth} is re-run on the specifications the end-to-end setting constructs for itself, blind to the agent's patch and its outcome, and split by whether the episode resolved the instance (Res.) or not (Unres.).}
\label{tab:spec-impact}
\footnotesize
\setlength{\tabcolsep}{3pt}
\begin{tabular}{@{}lcccccc@{}}
\toprule
& \multicolumn{2}{c}{\textbf{\textsc{Nagini}}} & \multicolumn{2}{c}{\textbf{\textsc{Velvet}}} & \multicolumn{2}{c}{\textbf{\textsc{Lean}}} \\
\cmidrule(lr){2-3}\cmidrule(lr){4-5}\cmidrule(lr){6-7}
& Res. & Unres. & Res. & Unres. & Res. & Unres. \\
\midrule
\multicolumn{7}{@{}l}{\textit{Per-property failure rate}}\\
\quad Axiom soundness & 7.6 & 7.1 & 0.9 & 1.8 & 3.1 & 4.7 \\
\quad Admissibility & 3.8 & 5.8 & 3.1 & 3.6 & 1.8 & 3.1 \\
\quad Soundness & 9.2 & 36.8 & 12.6 & 37.5 & 12.5 & 40.6 \\
\quad Completeness & 8.8 & 11.8 & 5.5 & 12.5 & 2.8 & 4.7 \\
\quad \textbf{Faithfulness} & \textbf{56.2} & \textbf{94.2} & \textbf{44.6} & \textbf{92.9} & \textbf{46.7} & \textbf{89.1} \\
\midrule
Fails the full audit & 58.1 & 94.9 & 44.9 & 92.9 & 48.5 & 89.1 \\
\bottomrule
\end{tabular}
\end{table}

\section{Discussion and Limitations}
\label{sec:discussion}

The verify-implies-resolve claim is not a proof. It rests on two audited trust points, axiom soundness and the fidelity of the refinement from the verified implementation to the executed patch (\cref{sec:correctness}), which we establish by execution, fuzzing, and adversarial audit. Trust points of this kind are normal for verified systems: a verified kernel or compiler also holds only under assumptions its authors make explicit. We release the evidence behind ours for readers to attack. Three narrower limits apply. First, each backend inherits the expressiveness of its language, and a few instances needed manual modeling where a faithful specification fell outside it, such as an import-timing side effect a partial-correctness logic cannot express. Second, a few instances relax the secondary no-leakage property (Appendix~\ref{app:leakage}): their specification view reveals a little more about the fix than the issue does, which weakens the evaluation for them but leaves \cref{claim:verify-resolve} intact. Third, the specification audit is a calibrated panel of LLM auditors, so its verdicts are evidence rather than proof.

\section{Conclusion}
We presented \benchproofer, a pipeline that turns a coding task with a known correct patch into a formally verified one, and \benchmark, the benchmark it builds for all $500$ instances of \sbv and the Python fragment of \swepro, each with a specification, a verifying implementation, and the ground-truth patch.
Nearly a third of test-passing patches admit counterexamples, and a correct specification lifts resolution well above the unaided baseline, but agents gain nothing writing one themselves: what they write misses part of the behavior the issue requires. Faithful specification synthesis from informal intent is the open problem.

\subsection*{Reproducibility statement}

We release both \benchproofer and \benchmark. \benchproofer includes the gate suite, agent scaffolds, and prompts from \cref{sec:method}, so the corpora can be rebuilt from scratch or the pipeline applied to a new benchmark. For each instance, \benchmark includes the specification module, the verified reference implementation, the pre-fix implementation that fails verification, the equivalent patch, the operation correspondence, the axiom provenance, and the gate outcomes, so each artifact and each step of the verify-implies-resolve argument can be checked independently (\cref{sec:correctness} and Appendix~\ref{app:structure}). Appendix~\ref{app:settings} gives the construction and evaluation settings, including scaffolds, budgets, and sanitization. Appendix~\ref{app:audit} describes the specification-synthesis audit and its calibration, and Appendix~\ref{app:licenses} covers licensing and intended use.

\bibliography{references}
\bibliographystyle{iclr2027_conference}

\clearpage
\appendix
\newpage
\renewcommand\thepart{}
\renewcommand\partname{}
\setcounter{secnumdepth}{-1}
\part{Appendix}
\setcounter{secnumdepth}{3}
\vspace{-3ex}
\setcounter{tocdepth}{2}
\parttoc
\newpage

\makeatletter
\setlength{\@fptop}{0pt}
\setlength{\@fpsep}{14pt}
\setlength{\@fpbot}{0pt plus 1fil}
\makeatother

\section{The benchmark corpora}
\label{app:corpora}

\benchmark is released as two corpora built by the same pipeline from two
different sources of issue-resolution tasks. The first covers all $500$
instances of \sbv; the second covers all $266$ Python instances of \swepro and
exists to test whether the construction methodology transfers to a task
distribution it was not designed around. Every instance is built independently
under each of the four specification backends of \cref{app:backends}, so the
unit of release is an \emph{(instance, backend)} pair, which we call a
\emph{bundle}. \cref{tab:release} summarizes the release.

\begin{table}[htbp]
\centering
\caption{The \benchmark release. A bundle is one instance under one
specification backend. ``Green'' counts bundles that pass the complete
admission suite of \cref{app:gates}; the \sbv corpus is green everywhere, and
the \swepro shortfall is analyzed in \cref{app:pro-nongreen}. Churn is added
plus removed lines in the gold patch.}
\label{tab:release}
\small
\begin{tabular}{@{}lcc@{}}
\toprule
& \textbf{\sbv} & \textbf{\swepro} \\
\midrule
Instances                         & 500   & 266 \\
Repositories                      & 12    & 3 \\
Bundles (instance $\times$ backend) & 2000  & 1064 \\
Green bundles                     & 2000 \,($100\%$) & 996 \,($93.6\%$) \\
Instances green under all four backends & 500 & 242 \\
\midrule
Gold patch churn, mean / median    & 14.3 / 7 & 145.5 / 73.5 \\
Files touched, mean / median       & 1.25 / 1  & 3.42 / 3 \\
Hunks, mean / median               & 2.44 / 1  & 9.84 / 7 \\
Problem statement, median characters & 1185    & 1226 \\
\midrule
Modeled functions (all backends)   & 1867  & 910 \\
Axioms for unchanged callees       & 347   & 69 \\
Property test-suite cases             & 16{,}118 & 7{,}364 \\
Differential inputs against real code & 129.8\,M & 57.9\,M \\
\bottomrule
\end{tabular}
\end{table}

Two properties hold by construction across both corpora. First, the underlying
task is passed through untouched: the problem statement, base commit, gold
patch, and hidden tests are byte-identical to the source benchmark, and an
integrity check re-verifies this for every bundle at release time. Any result
on \benchmark is therefore comparable to the corresponding result upstream.
Second, the gold patch of every released instance resolves in the official
per-instance container, which anchors the external correctness oracle to the one
the source benchmark already defines.

\subsection{The \sbv corpus}
\label{app:composition}

The $500$ \sbv instances span $12$ open-source Python projects, with Django
supplying just under half; \cref{app:licenses} gives each project's license.
\cref{tab:composition} gives the per-project
breakdown together with three indicators of how much formal modeling each
project demanded: the number of repository functions modeled, the number of
instances that required at least one axiom for an unchanged callee, and the
number that required modeling more than one function. The three prover backends
are reported separately because they make different modeling choices on the same
task, a point we return to in \cref{app:where-it-binds}.

\begin{table}[htbp]
\centering
\caption{Per-project composition of the \sbv corpus. ``Fns'' is the number of
repository functions modeled under that backend, summed over the project;
``Ax.'' is the number of axioms for unchanged callees. ``$\geq$1 axiom'' and
``Multi-site'' count instances that require, respectively, at least one axiom
and more than one modeled function under at least one backend.}
\label{tab:composition}
\small
\setlength{\tabcolsep}{4pt}
\begin{tabular}{@{}lccccccccc@{}}
\toprule
& & \multicolumn{2}{c}{\textbf{\nagini}} & \multicolumn{2}{c}{\textbf{\velvet}}
& \multicolumn{2}{c}{\textbf{\lean}} & & \\
\cmidrule(lr){3-4}\cmidrule(lr){5-6}\cmidrule(lr){7-8}
\textbf{Project} & \textbf{Inst.} & Fns & Ax. & Fns & Ax. & Fns & Ax.
& \textbf{$\geq$1 axiom} & \textbf{Multi-site} \\
\midrule
Django       & 231 & 305 & 97 & 271 & 19 & 255 & 0 & 55 & 59 \\
SymPy        & 75  & 115 & 63 & 107 & 27 & 84  & 1 & 31 & 28 \\
Sphinx       & 44  & 74  & 16 & 49  & 5  & 48  & 0 & 12 & 22 \\
Matplotlib   & 34  & 49  & 8  & 43  & 7  & 41  & 0 & 7  & 13 \\
scikit-learn & 32  & 43  & 25 & 37  & 7  & 34  & 1 & 17 & 9  \\
Astropy      & 22  & 35  & 22 & 30  & 8  & 25  & 0 & 7  & 8  \\
xarray       & 22  & 29  & 11 & 26  & 0  & 23  & 0 & 6  & 7  \\
pytest       & 19  & 28  & 6  & 26  & 0  & 20  & 1 & 5  & 8  \\
Pylint       & 10  & 11  & 8  & 10  & 4  & 10  & 0 & 4  & 1  \\
Requests     & 8   & 10  & 4  & 8   & 1  & 8   & 0 & 3  & 1  \\
seaborn      & 2   & 4   & 3  & 3   & 2  & 3   & 0 & 2  & 2  \\
Flask        & 1   & 1   & 1  & 1   & 0  & 1   & 0 & 1  & 0  \\
\midrule
\textbf{Total} & \textbf{500} & \textbf{704} & \textbf{264} & \textbf{611}
& \textbf{80} & \textbf{552} & \textbf{3} & \textbf{150} & \textbf{158} \\
\bottomrule
\end{tabular}
\end{table}

\paragraph{What the tasks look like.}
The \sbv fixes are small and local. The median gold patch adds four lines and
removes two, touches one file, and consists of a single hunk; $429$ of $500$
patches are confined to one file, and the largest touches $21$. The upstream
difficulty labels place $194$ instances at under fifteen minutes of developer
time, $261$ between fifteen minutes and an hour, $42$ between one and four
hours, and $3$ above four hours. The graded test sets are asymmetric: a median
of one fail-to-pass test decides the instance, against a median of $50$
pass-to-pass tests that guard against regression. The behavioral change is
concentrated in a small number of functions, so a specification can pin it
without modeling the whole repository. The surrounding code still has to be
respected; the axioms of \cref{app:axioms} do that.

\paragraph{Modeling load.}
Under \nagini, the corpus models $704$ repository functions, a mean of $1.41$
and a maximum of $7$ per instance, and assumes $264$ axioms spread over $137$
instances. \velvet models $611$ functions with $80$ axioms over $44$ instances,
and pure \lean models $552$ functions with only $3$ axioms in total. The
downward trend in axiom count reflects how deeply each backend can model a
callee. \lean over Mathlib can usually
define an unchanged helper outright, leaving no assumed contract to state,
whereas \nagini must summarize anything it cannot express in its verification
fragment. Taking the union over backends, $150$ instances ($30.0\%$) need at
least one axiom somewhere and $158$ ($31.6\%$) are multi-site somewhere; only a
single instance requires an axiom under all three prover backends, and $350$
need none anywhere.

\paragraph{Representation choices.}
Verification forces a choice of carrier for values that the prover cannot
represent natively. Each bundle records its choices. Under \nagini, $11$
instances require reasoning about real-valued arithmetic and are verified under
an interpreted real encoding; the remaining $489$ need no floating-point
reasoning at all. Strings are the other recurring decision: $227$ \nagini
bundles model string content explicitly, most often as a sequence of integer
code points or as an interned identity token, while $186$ declare that the
instance's behavior does not depend on string content and treat strings
opaquely. \velvet, which sits inside \lean, more often carries strings as a list
of characters. \Cref{app:where-it-binds} works through the cases where this
choice changes what a specification can say.

\subsection{The \swepro corpus}
\label{app:pro-composition}

\swepro draws its tasks from three repositories that \sbv does not cover, and
its patches are an order of magnitude larger: a mean of $145.5$ changed lines
across $3.4$ files and $9.8$ hunks, against $14.3$ lines in $1.25$ files for
\sbv. \cref{tab:pro-composition} gives the breakdown. The three projects also
differ in kind from the \sbv set. Ansible is a configuration-management engine
whose behavior is dominated by process orchestration and templating; OpenLibrary
is a web application with a large persistence surface; and qutebrowser is a
desktop application built around an event loop and a Qt widget hierarchy. None
of the three is the kind of numerically-oriented library that verification
tooling is usually demonstrated on. The issue text, however, is about as long:
the median problem statement is $1{,}226$ characters against $1{,}185$ on \sbv.
A \swepro task therefore asks for ten times the code change on the same amount
of stated intent.

\begin{table}[htbp]
\centering
\caption{Per-project composition of the \swepro corpus. Columns are as in
\cref{tab:composition}; ``Churn'' is the mean added plus removed lines in the
gold patch.}
\label{tab:pro-composition}
\small
\setlength{\tabcolsep}{4pt}
\begin{tabular}{@{}lcccccccc@{}}
\toprule
& & & \multicolumn{2}{c}{\textbf{\nagini}} & \multicolumn{2}{c}{\textbf{\velvet}}
& \multicolumn{2}{c}{\textbf{\lean}} \\
\cmidrule(lr){4-5}\cmidrule(lr){6-7}\cmidrule(lr){8-9}
\textbf{Project} & \textbf{Inst.} & \textbf{Churn} & Fns & Ax. & Fns & Ax.
& Fns & Ax. \\
\midrule
ansible/ansible             & 96 & 171.0 & 117 & 25 & 92  & 0 & 106 & 1 \\
internetarchive/openlibrary & 91 & 175.0 & 107 & 15 & 102 & 8 & 107 & 0 \\
qutebrowser/qutebrowser     & 79 & 79.4  & 100 & 17 & 88  & 1 & 91  & 2 \\
\midrule
\textbf{Total} & \textbf{266} & \textbf{145.5} & \textbf{324} & \textbf{57}
& \textbf{282} & \textbf{9} & \textbf{304} & \textbf{3} \\
\bottomrule
\end{tabular}
\end{table}

\begin{figure}[htbp]
\centering
\includegraphics{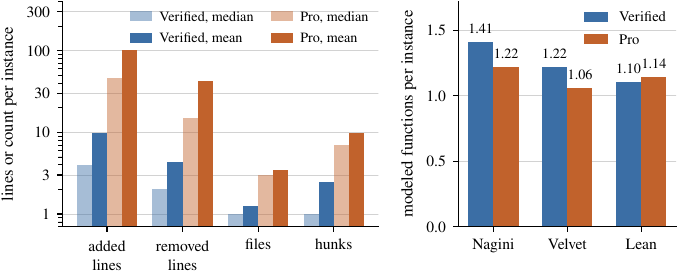}
\caption{Fix size and specification size in the two corpora.
\textbf{Left:} the size of the reference fix, median and mean, on a logarithmic
scale. \textbf{Right:} the number of functions a specification models per
instance. A \swepro fix is an order of magnitude larger by added lines and
touches three times as many files, yet its specifications model no more
functions than an \sbv specification does.}
\label{fig:corpus-shape}
\end{figure}

The larger patches do not translate into proportionally larger specifications.
The number of modeled functions per instance is slightly \emph{lower} than on
\sbv ($1.22$ against $1.41$ under \nagini), because a \swepro patch usually
spreads a single behavioral change over many call sites, boilerplate updates,
and test-support edits, only a few of which carry the behavior the graded tests
observe. Identifying that core is what gets harder. Mean verification time on the green \swepro bundles is
comparable to \sbv under \nagini ($29.4$\,s against $25.8$\,s) and lower under
both \lean-based backends ($3.6$\,s against $6.0$\,s under \velvet, $4.6$\,s
against $10.5$\,s under \lean), because the modeled functions themselves are no
more complex. \cref{fig:corpus-shape} puts the two quantities side by side.

\begin{figure}[htbp]
\centering
\includegraphics{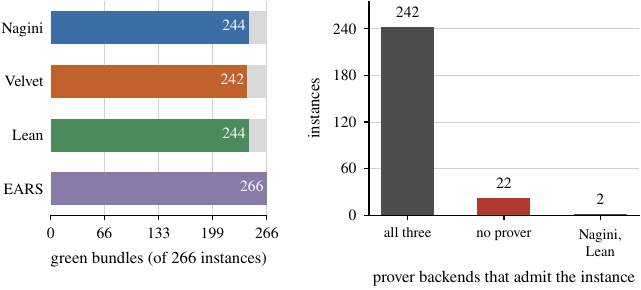}
\caption{Admission on the \swepro corpus. \textbf{Left:} green bundles per
backend. \textbf{Right:} the joint pattern over the three prover backends. The
$22$ instances that no prover backend admits fail for a shared, diagnosable
reason (\cref{app:pro-nongreen}) rather than through independent per-backend
attrition.}
\label{fig:pro-greenness}
\end{figure}

The admission rate is where the difference shows. $996$ of $1064$ \swepro
bundles are green, against $2000$ of $2000$ on \sbv, and $242$ of $266$
instances are green under all four backends. The shortfall is highly structured:
$22$ instances fail under every prover backend for the same reason, which we
analyze next, and only $2$ more are backend-specific.
\cref{fig:pro-greenness} shows the pattern.

\subsection{Tasks that admit no verified twin}
\label{app:pro-nongreen}

Of the $266$ \swepro tasks, $22$ are not constructed on any prover backend:
\nagini, \velvet, and \lean independently agree that no \emph{verify
$\Rightarrow$ correct} twin can be built. A specification in our setting
constrains the \emph{values} a function computes, and each of these $22$ tasks
changes something else: a name, a call shape, an internal representation, or an
effect outside the value domain. In every case the fail-to-pass test
discriminates on that non-value property. A specification that pinned the
computed values would be satisfied equally by the pre-fix and post-fix code, and
the discrimination check of \cref{app:mechanical} would correctly refuse the
instance.

We list all $22$ below, grouped by root cause, with the diagnosis recorded for
each task during construction.

\paragraph{Renaming and relocation ($16$ tasks).}
These tasks move or rename code without altering what it computes. The
fail-to-pass test therefore discriminates on a name, an import source, or a
class identity, none of which a specification written over values can see.
\begin{itemize}[leftmargin=1.2em,itemsep=2pt,topsep=2pt]
  \item \id{ansible\_\_ansible-379058e10f3dbc0fdcaf80394bd09b18927e7d33-v1055803c3a812189a1133297f7f5468579283f86}\\*
        Swaps the import source for the collections ABCs; the only other changes are a changelog entry and a lint literal.
  \item \id{ansible\_\_ansible-502270c804c33d3bc963930dc85e0f4ca359674d-v7eee2454f617569fd6889f2211f75bc02a35f9f8}\\*
        Method bodies are moved verbatim into \texttt{CommandStrategy} and \texttt{FileStrategy}.
  \item \id{internetarchive\_\_openlibrary-25858f9f0c165df25742acf8309ce909773f0cdd-v13642507b4fc1f8d234172bf8129942da2c2ca26}\\*
        A pure relocation whose fail-to-pass test checks only that an \texttt{ImportError} no longer occurs.
  \item \id{internetarchive\_\_openlibrary-308a35d6999427c02b1dbf5211c033ad3b352556-ve8c8d62a2b60610a3c4631f5f23ed866bada9818}\\*
        \texttt{List} and \texttt{ListChangeset} are moved byte-for-byte; the test asserts \texttt{isinstance} and registry membership.
  \item \id{internetarchive\_\_openlibrary-3aeec6afed9198d734b7ee1293f03ca94ff970e1-v13642507b4fc1f8d234172bf8129942da2c2ca26}\\*
        Wikidata methods are renamed to private names, with two of them merged by concatenation.
  \item \id{internetarchive\_\_openlibrary-757fcf46c70530739c150c57b37d6375f155dc97-ve8c8d62a2b60610a3c4631f5f23ed866bada9818}\\*
        Relocation only; before the patch the test fails solely because the name is missing.
  \item \id{internetarchive\_\_openlibrary-798a582540019363d14b2090755cc7b89a350788-v430f20c722405e462d9ef44dee7d34c41e76fe7a}\\*
        Relocation only; the test asserts which module the class is imported from.
  \item \id{internetarchive\_\_openlibrary-8a5a63af6e0be406aa6c8c9b6d5f28b2f1b6af5a-v0f5aece3601a5b4419f7ccec1dbda2071be28ee4}\\*
        \texttt{bash\_run} and \texttt{limit\_server} are byte-identical after the move.
  \item \id{qutebrowser\_\_qutebrowser-3d01c201b8aa54dd71d4f801b1dd12feb4c0a08a-v5fc38aaf22415ab0b70567368332beee7955b367}\\*
        Renames in the resource-path module together with a dead-import cleanup.
  \item \id{qutebrowser\_\_qutebrowser-3fd8e12949b8feda401930574facf09dd4180bba}\\*
        Six commands are renamed to a \texttt{cmd-} prefix, keeping deprecated-name aliases.
  \item \id{qutebrowser\_\_qutebrowser-5fdc83e5da6222fe61163395baaad7ae57fa2cb4-v363c8a7e5ccdf6968fc7ab84a2053ac78036691d}\\*
        \texttt{parse\_font\_families} is wrapped in a \texttt{FontFamilies} type while the loop body stays identical, confirmed by fuzzing $200{,}000$ inputs.
  \item \id{qutebrowser\_\_qutebrowser-bedc9f7fadf93f83d8dee95feeecb9922b6f063f-v2ef375ac784985212b1805e1d0431dc8f1b3c171}\\*
        Relocation only; before the patch the test fails with an \texttt{AttributeError} on the new export.
  \item \id{qutebrowser\_\_qutebrowser-de4a1c1a2839b5b49c3d4ce21d39de48d24e2091-v2ef375ac784985212b1805e1d0431dc8f1b3c171}\\*
        Relocation only; the test fails before the patch solely on \texttt{ImportError}.
  \item \id{qutebrowser\_\_qutebrowser-ebfe9b7aa0c4ba9d451f993e08955004aaec4345-v059c6fdc75567943479b23ebca7c07b5e9a7f34c}\\*
        The Qt message handler moves between logging modules, with an added initialization wrapper.
  \item \id{qutebrowser\_\_qutebrowser-f91ace96223cac8161c16dd061907e138fe85111-v059c6fdc75567943479b23ebca7c07b5e9a7f34c}\\*
        \texttt{hide\_qt\_warning} and \texttt{QtWarningFilter} move to the Qt logging utility module.
  \item \id{qutebrowser\_\_qutebrowser-fd6790fe8c02b144ab2464f1fc8ab3d02ce3c476-v2ef375ac784985212b1805e1d0431dc8f1b3c171}\\*
        \texttt{buffer} is renamed to \texttt{tab\_select}, and the corresponding completion model to \texttt{tabs}.
\end{itemize}

\paragraph{Function signature changes ($2$ tasks).}
The discriminating difference is the shape of the call, not its result, so the
test flips on a \texttt{TypeError} rather than on a value.
\begin{itemize}[leftmargin=1.2em,itemsep=2pt,topsep=2pt]
  \item \id{ansible\_\_ansible-39bd8b99ec8c6624207bf3556ac7f9626dad9173-v1055803c3a812189a1133297f7f5468579283f86}\\*
        A module-runner entry point drops its \texttt{job\_path} parameter in favor of a monkeypatchable global; the single test flips on a \texttt{TypeError}.
  \item \id{qutebrowser\_\_qutebrowser-e5340c449f23608803c286da0563b62f58ba25b0-v059c6fdc75567943479b23ebca7c07b5e9a7f34c}\\*
        The logic genuinely changed, from a boolean return to accept/reject side effects, but the fail-to-pass test does not observe the difference.
\end{itemize}

\paragraph{Internal data structure changes ($2$ tasks).}
Here the pre- and post-patch code compute the same values, and the test keys on
a representation or call-route difference instead.
\begin{itemize}[leftmargin=1.2em,itemsep=2pt,topsep=2pt]
  \item \id{ansible\_\_ansible-e0c91af45fa9af575d10fd3e724ebc59d2b2d6ac-v30a923fb5c164d6cd18280c02422f75e611e8fb2}\\*
        Pre- and post-patch code are extensionally identical, measured inside the instance image; the test discriminates only on which bound method is monkeypatched. Two backends had previously admitted this task and were withdrawn on adjudication, because their pre-fix models asserted a value footprint the real code never produces.
  \item \id{internetarchive\_\_openlibrary-3f580a5f244c299d936d73d9e327ba873b6401d9-v0f5aece3601a5b4419f7ccec1dbda2071be28ee4}\\*
        A \texttt{list} becomes a \texttt{tuple} and a \texttt{dict} a read-only mapping proxy, with contents identical.
\end{itemize}

\paragraph{External side effect changes ($2$ tasks).}
The change is observable only outside the value domain, in process, filesystem,
or ordering state that the verifier does not model.
\begin{itemize}[leftmargin=1.2em,itemsep=2pt,topsep=2pt]
  \item \id{ansible\_\_ansible-8127abbc298cabf04aaa89a478fc5e5e3432a6fc-v30a923fb5c164d6cd18280c02422f75e611e8fb2}\\*
        The graded tests pass merely because the task executor and the worker process stopped accepting a standard-input handle. The backends disagreed on the cause here, and it was adjudicated as an arity-only change.
  \item \id{qutebrowser\_\_qutebrowser-e57b6e0eeeb656eb2c84d6547d5a0a7333ecee85-v2ef375ac784985212b1805e1d0431dc8f1b3c171}\\*
        The same file objects and completion count are delivered in the same order before and after the patch.
\end{itemize}

\paragraph{What the list shows.}
None of the $22$ is a verifier limitation. In each case the task's own
fail-to-pass test is insensitive to the values the code computes, so the gold
patch and its predecessor are behaviorally indistinguishable in the domain a
functional specification talks about. Our methodology therefore applies where a
known gold patch exists and the requested change is observable as a change in
computed values. The remaining $2$ non-green bundles are per-backend
refusals on instances that other backends admit, so those instances stay usable
under the backends that do admit them.

\subsection{Instances with a relaxed disclosure disposition}
\label{app:leakage}

The specification view shown to a solver must reveal nothing about the fix
beyond what the public problem statement already discloses. For a minority of
instances this requirement collides with faithfulness. The behavior the hidden
tests actually check appears only in developer discussion or in the gold change,
never in the solver-visible problem statement. A faithful specification must
then name a discriminator absent from the public text, while a strictly
non-leaking specification, judged against that text alone, cannot pin the
tested behavior. The two properties are simultaneously unachievable because the
public task text and the gold fix are about different facets of the same
symptom.

For these instances we keep the bundle, permit the discriminator in the
specification view, and document the relaxation per instance. An instance
qualifies only when an adversarial analysis establishes that the leak is
unavoidable, that is, that no reformulation derivable from the public problem
statement could pin the tested behavior. Across the \sbv corpus the disposition
applies to $17$ \nagini bundles, $37$ \velvet bundles, $11$ \lean bundles, and no
\ears bundle; across \swepro it applies to $1$, $1$, $2$, and $7$ respectively,
eleven bundles in total. The
counts differ by backend: the identity of the unavoidable discriminator depends
on how the behavior is encoded, and a phrase that is unavoidable in one language
may be avoidable in another.

The verification core, namely verification, discrimination, soundness,
completeness, and axiom soundness, and the harness resolution of the gold patch
all hold unchanged. Only the secondary non-disclosure property is relaxed, for a
documented minority. The bundle records the relaxation, so a downstream user can
exclude those instances.

\mbox{\id{scikit-learn-26194}} is representative. The issue concerns the leading
threshold returned by a receiver-operating-characteristic routine. The gold fix
prepends positive infinity and the hidden test asserts exactly that, whereas the
public issue proposes the \emph{opposite} remedy of clipping thresholds to at
most one. The discriminator ``infinity'' appears nowhere in the public text, so
any faithful specification must introduce it, and any specification that avoids
it is unfaithful to the graded behavior. \Cref{app:faithfulness} shows the
specification view for this instance alongside the public text.

\section{Anatomy of a released instance}
\label{app:structure}

Every bundle is self-contained. It carries the specification, both
implementations that the specification is meant to separate, the repository diff
that resolves the task, the two documents that bind the formal model to the real
code, the executable evidence that substantiates each admission check, and the
record of what was checked and how. A reader who downloads a bundle can re-run
the verifier, re-run the probes, and re-derive every stamp.

\cref{tab:artifacts} lists the artifact classes and their typical size under
each backend. The \sbv corpus ships $3.54$ million lines of specification,
proof, probe, and audit text in total, of which the specifications themselves
are the smallest part; most of the volume is refutation machinery and recorded
evidence.

\begin{table}[!t]
\centering
\caption{Artifacts released with each bundle, and their median size in lines
under each backend. The first three rows are the objects the guarantee is
stated over; the middle two bind the formal model to the repository; the last
three are the evidence and the record. Where an artifact is not carried by all
$500$ bundles, a parenthesized count gives how many carry it; a dash means the
backend does not use it. \ears ships a requirements document and an audit
record only, since it has no verifier to drive.}
\label{tab:artifacts}
\small
\setlength{\tabcolsep}{5pt}
\begin{tabular}{@{}lp{5.1cm}cccc@{}}
\toprule
\textbf{Artifact} & \textbf{Role} & \textbf{\nagini} & \textbf{\velvet}
& \textbf{\lean} & \textbf{\ears} \\
\midrule
Specification module & The verified object: signatures, contracts, axioms, pure
helpers, and a reference implementation with its proof annotations
& 151 & 100 & 107 & 106 \\
Pre-fix twin & The same specification paired with the reported buggy behavior;
required to fail verification
& 153\,(497) & 103\,(481) & 97\,(479) & \na \\
Equivalent patch & The repository diff that applies at the base commit and
passes the hidden tests
& \multicolumn{4}{c}{shared across backends} \\
\midrule
Correspondence map & The operation-by-operation refinement from the reference
implementation to the patch & 102 & 87 & 59 & \na \\
Provenance map & The binding from each modeled symbol, type, and axiom to its
repository counterpart & 95 & 89 & 82 & \na \\
\midrule
Property test suite & Candidate implementations with the verifier outcome each
must produce & 529 & 716 & 742 & \na \\
Axiom probes & Executable checks that run each axiom against the real callee
inside the instance container & 175\,(477) & 143\,(262) & 24\,(496) & \na \\
Differential harness & The reference implementation run against the patched
repository code on generated inputs & 300 & 157\,(499) & 144 & \na \\
\midrule
Admission record & Every check, its outcome, the verifier configuration, and
the adjudication that closed it & 733 & 603 & 543 & 366 \\
\bottomrule
\end{tabular}
\end{table}

\subsection{The central pair and the executed patch}
\label{app:central-pair}

The guarantee is a statement about two objects. The \emph{specification module}
is verified; the \emph{equivalent patch} is executed. Neither alone is
sufficient, so the bundle also carries the documents that connect them.

\paragraph{The specification module.}
The module declares the signatures of the functions the fix touches, states
their preconditions and postconditions, axiomatizes the callees the fix leaves
alone, defines whatever pure helpers the postconditions need, and closes with a
reference implementation carrying the annotations the verifier requires. It is
one file, and it is small: a median of $151$ lines under \nagini, $100$ under
\velvet, $107$ under \lean, and $106$ under \ears.

The proof burden inside that file is not small. A \nagini module states a mean
of $6.8$ preconditions and $12.9$ postconditions; $264$ of the $500$ modules
need at least one loop invariant, at a mean of $10.7$ where they appear, and
$443$ state an explicit termination measure. Roughly half the non-blank content
is specification rather than implementation: a mean of $28.8$ specification
lines against $21.1$ lines of executable body. The \lean modules declare a mean
of $1.87$ theorems and $4.5$ supporting definitions each, spend a median of $13$
lines inside the proof, and no proof invokes \texttt{sorry} anywhere in
the corpus; across all $500$ modules there are exactly $7$ local axiom
declarations. \velvet modules declare a mean of $1.23$ methods with $2.32$
postconditions and hand each correctness obligation to the solver, so their
proof regions are the shortest of the three (median $4$ lines) while their pure
definitions are the longest (median $32$).

\paragraph{What the solver sees.}
The module as released contains the reference implementation, which would give
the answer away. Every evaluation setting that shows a specification to a solver
shows a \emph{view} of the module instead: the signatures, the contracts, the
axioms, and the pure helpers, with the reference body removed. The view is
generated from the verified module. \Cref{app:harness} describes what else is
withheld.

\paragraph{The pre-fix twin.}
Each bundle also ships the same specification paired with the implementation the
issue complains about. This artifact exists to \emph{fail}. It is the witness
that the specification is strong enough to see the reported defect, and its
failure is the discrimination check of \cref{app:mechanical}. Where a fix
repairs several independent behaviors, the bundle ships one twin per behavior:
$419$ \nagini bundles carry a single twin, $65$ carry two, $13$ carry three, and
$3$ carry four, for $600$ twins in total; \velvet ships $569$ and \lean $540$.
A specification whose twin verifies is not admitted.

\paragraph{The equivalent patch.}
The patch is the object that must actually resolve the task, and it is shared by
all four backends of an instance because it is a property of the repository, not
of the specification language. It applies at the frozen base commit and passes
the hidden tests in the task's official container. On \sbv the patches are
small: a median of one file and one hunk. The correspondence below is therefore
tractable to audit by hand.

\paragraph{The correspondence and provenance maps.}
The transfer from the verified object to the executed one rests on these two
documents.

The \emph{correspondence map} walks the reference implementation against the
patch one operation at a time and labels each step with the behavior-preserving
relationship that justifies it: an identical operation, a type refinement from
the modeled carrier to the concrete repository type, or a call to an
axiomatized callee. A step that fits none of the three is a defect, and the
equivalence check of \cref{app:audit} is an adversarial reading of exactly this
document. The maps run to a median of $102$ lines under \nagini, longer than the
specification modules they annotate.

The \emph{provenance map} records where the model came from. For each modeled
symbol it names the repository entity it abstracts: which attribute an integer
field stands for, which class a modeled record corresponds to, which method a
modeled operation implements. For each axiom it names the exact callee and
states the evidence for the contract asserted of it. For each representation
decision it states the decision and its justification: that strings are carried
as sequences of code points because the tested behavior inspects characters, or
that they are treated opaquely because it does not. On \sbv, $227$ \nagini
bundles model string content explicitly and $186$ declare that the instance's
behavior does not depend on it; $11$ carry an interpreted encoding for real
arithmetic. \Cref{app:where-it-binds} works through the cases where the choice
changes what a specification can express.

\subsection{Axioms for code the fix does not change}
\label{app:axioms}

A fix to one function usually sits in the middle of code that the fix does not
touch. Verifying it end to end would mean formalizing that surrounding code, and
for a real repository that is neither feasible nor useful: the issue is not about
the database driver, the template engine, or the array library. Instead, each
unchanged callee that the specification depends on is given an \emph{axiom},
which states the weakest property of that callee the proof actually needs, and
nothing more.

\begin{axiombox}[label=box:axiom]{An axiom for an unchanged callee, from \id{django-11179} (abridged)}
\begin{minted}{python}
@ContractOnly
def remove_rows(db: Database, identifier: int) -> int:
    """The unchanged row-removal primitive. It issues exactly one
    row-removal statement for `identifier`, reports the number of
    rows that statement removed, and does not touch the in-memory
    object. The store may instead refuse the statement, and which
    answer a caller gets is the store's to decide."""
    Requires(MustTerminate(1))
    Requires(-(2 ** 63) <= identifier and identifier < 2 ** 63)
    Requires(RemovalBudget(LEDGER))
    Ensures(CarriedOutRemoval(db, identifier, Result()))
    Ensures(Result() >= 0)
    Exsures(RemovalRefused, RefusedRemoval(db, identifier))
\end{minted}
\end{axiombox}

Box~\ref{box:axiom} is representative. The fix in that instance clears an
in-memory identifier after a single-object fast delete; the row-removal
primitive it calls is untouched, and the axiom asserts only what the proof
needs from it. The axiom deliberately leaves several things open. It does not
say when the store refuses a removal, so no implementation can arrange the
refusing outcome or predict it; it does not say what count comes back beyond
non-negativity; and it says nothing at all about the store's own tallies. An
axiom that fixed the refusal condition would let an implementation branch on it
and still verify, and an axiom that exposed the tallies would license
implementations whose real form pays a database round trip per read. The
justification for each assertion, and the identity of the primitive the axiom
stands for, are recorded in the provenance map; the executable probe of this
axiom calls Django's real deletion primitive.

\paragraph{Axioms are a last resort.}
An axiom is an assumption, so a bundle that can avoid one does. The preferred
alternative is to model the callee outright, and the corpus does this far more
often than it axiomatizes: $362$ \nagini bundles model at least one callee
deeply, $500$ such models in total, against $264$ axioms over $137$ instances.
The balance shifts sharply with the expressiveness of the language.
\velvet needs $80$ axioms across $44$ instances, and pure \lean over Mathlib
needs $3$ in the entire $500$-instance corpus. Taking the union over the three
prover backends, $150$ \sbv instances ($30.0\%$) require at least one axiom
somewhere, $350$ require none anywhere, and exactly one requires an axiom under
all three. On \swepro the pattern repeats at lower intensity: $57$ axioms under
\nagini, $9$ under \velvet, $3$ under \lean.

\paragraph{Axioms are tested against the running repository.}
Each bundle ships an executable probe that constructs inputs, calls the
\emph{real} callee inside the task's official container, and checks the observed
behavior against what the axiom asserts. The probes are substantial programs, at
a median of $175$ lines under \nagini. The same probes also check deeply
modeled callees against the real code, so most bundles carry one even without
an axiom. Across both corpora they contribute $25.7$ million recorded
observations of real callee behavior. On \sbv, $477$ \nagini bundles,
$262$ \velvet bundles, and $496$ \lean bundles carry a probe file; under \lean,
$286$ of these only record that there is nothing to check. No probe refutes its
axiom on any admitted bundle. \Cref{app:mechanical} reports the check itself.

\paragraph{What an axiom cannot be.}
An axiom must never contain the fix. An axiom that asserted the corrected
behavior of a callee would make the specification verify for the wrong reason
and the pre-fix twin fail for the wrong reason. The audit of \cref{app:audit}
attacks axioms specifically, and an axiom over a callee that the patch
\emph{does} touch is rejected outright. In the released corpora every
axiomatized callee lies outside the patch.

\subsection{Attached evidence and the admission record}
\label{app:evidence}

A specification must admit legitimate implementations, accept correct ones, and
reject wrong ones. Those are claims about infinitely many programs, and the
bundle discharges them with finite, re-runnable evidence: a property test suite of candidate
implementations with the verdict each must receive, a differential harness that
compares the reference implementation against the real repository code, and the
probes of \cref{app:axioms}.

\paragraph{The property test suite.}
Each bundle ships a set of candidate implementations of its modeled functions,
each labeled with the verdict the verifier must return. The labels fall into
three lanes. \emph{Admissibility} cases are implementations shaped like the ones
the task's own tests exercise; if the verifier rejects them, the specification's
preconditions are too strong to be about the real task. \emph{Soundness} cases
are correct implementations written differently from the reference (statements
reordered, a helper inlined, a different algorithm for the same function); if the
verifier rejects them, the specification has over-fitted to one way of writing
the answer. \emph{Discrimination} cases are wrong implementations, including the
original bug and deliberate near misses; if the verifier accepts them, the
specification is too weak to be worth having.

\begin{table}[htbp]
\centering
\caption{The released property test suites. Each case is a candidate
implementation carried in the bundle together with the verdict the verifier must
return. ``Must verify'' pools the admissibility and soundness lanes, except
five \nagini admissibility cases that must be rejected; ``must reject'' is
otherwise the discrimination lane. Every case in both corpora produces its
required verdict.}
\label{tab:suite}
\small
\setlength{\tabcolsep}{5pt}
\begin{tabular}{@{}lcccccc@{}}
\toprule
& & \multicolumn{3}{c}{\textbf{Lane}} & \multicolumn{2}{c}{\textbf{Outcome}} \\
\cmidrule(lr){3-5}\cmidrule(lr){6-7}
\textbf{Backend} & \textbf{Cases} & Admis. & Sound. & Discrim.
& Must verify & Must reject \\
\midrule
\multicolumn{7}{@{}l}{\sbv, \emph{500 instances}} \\
\nagini & 5{,}030 & 533 & 966 & 3{,}531 & 1{,}494 / 1{,}494 & 3{,}536 / 3{,}536 \\
\velvet & 5{,}208 & 526 & 1{,}268 & 3{,}414 & 1{,}794 / 1{,}794 & 3{,}414 / 3{,}414 \\
\lean   & 5{,}880 & 513 & 1{,}249 & 4{,}118 & 1{,}762 / 1{,}762 & 4{,}118 / 4{,}118 \\
\midrule
\multicolumn{7}{@{}l}{\swepro, \emph{266 instances}} \\
\nagini & 2{,}231 & 266 & 520 & 1{,}445 & 786 / 786 & 1{,}445 / 1{,}445 \\
\velvet & 2{,}509 & 263 & 750 & 1{,}496 & 1{,}013 / 1{,}013 & 1{,}496 / 1{,}496 \\
\lean   & 2{,}624 & 267 & 733 & 1{,}624 & 1{,}000 / 1{,}000 & 1{,}624 / 1{,}624 \\
\midrule
\textbf{Total} & \textbf{23{,}482} & \textbf{2{,}368} & \textbf{5{,}486}
& \textbf{15{,}628} & \multicolumn{2}{c}{\textbf{23{,}482 / 23{,}482}} \\
\bottomrule
\end{tabular}
\end{table}

\cref{tab:suite} reports the property test suites. The \sbv corpus carries $16{,}118$
cases and \swepro a further $7{,}364$, a median of $9$ to $11$ per bundle, and
every case produces the verdict it is labeled with. The single largest property test suite
holds $62$ cases, for an instance whose specification had to be defended against
an unusually large family of near misses.

Two features of the property test suite make the discrimination lane meaningful. First, a
rejection is credited only when the verifier's complaint is about the intended
property. Under the two \lean-based backends a discrimination case is
accompanied by a witness: a concrete input on which the candidate and the
specification disagree. The case counts as refuted only when that witness is
confirmed against the candidate and cleared against the frozen reference. The
property test suites are therefore expensive: driving the $16{,}118$ \sbv cases takes
$32{,}662$ verifier invocations and roughly $72$ hours of prover time. Second,
the property test suites are shipped as source, so anyone can add a case.

\paragraph{The differential harness.}
The property test suite checks the specification against candidate implementations. The
differential harness tests the \emph{reference implementation} against the real
thing. It generates inputs, runs the patched repository code on them inside the
task's container, runs the modeled reference on the corresponding modeled
inputs, and compares. It catches a reference implementation that satisfies the
specification but does not match the repository code. Across the \sbv corpus the
harness compares $129.8$ million inputs, and across \swepro a further $57.9$
million. There are no disagreements in either corpus.

\paragraph{The admission record.}
Each bundle closes with a record of what was checked. It lists every admission
check of \cref{app:gates} with its outcome, the reason for any check that does
not apply, the verifier configuration used, the representation decisions of
\cref{app:central-pair}, and the adjudications that closed the checks requiring
judgment. It also carries, verbatim, the attacks that an audit attempted and the
reason each failed. The records are long, at a median of $733$ lines under
\nagini. They let a reader re-audit the corpus without re-running
construction.

A check that does not apply is recorded as inapplicable rather than as a pass,
and the distinction matters when reading the per-backend tables of
\cref{app:mechanical}. An axiom probe is inapplicable to a bundle with no
axioms; a mutation check is inapplicable to a reference implementation with no
operators to mutate, such as a body that only performs an assignment. In those
cases the burden falls on the checks that do apply: for an assignment-only body,
on the pre-fix twin, which must still fail. No bundle is admitted on
inapplicability alone.

\section{Related work}
\label{sec:related}

\paragraph{Issue-resolution benchmarks and their oracles.}
\swebench \citep{jimenez2024swebench} established the real-world issue-resolution task with a set of hidden-test oracles; \sbv \citep{openai2024swebenchverified} refined it into a human-validated 500-instance subset. Extensions address visual domains \citep{yang2025swebenchmm}, additional languages \citep{zan2025multiswebench}, agent training \citep{pan2025swegym}, and contamination-resistant task streams \citep{zhang2025swebenchlive, badertdinov2025swerebench}.
A parallel line scrutinizes the oracle: \citet{aleithan2024swebenchplus} document solution leakage and weak tests, while \citet{liang2025swebenchillusion} attribute part of reported performance to memorization. The community response has been to strengthen or refresh the tests \citep{ahmed2024tddbench,ahmed2025otter,jain2025livecodebench,zhuo2025bigcodebench}. \benchmark replaces tests with a formal specification, so that a verified fix is correct on the entire specified input domain.

\paragraph{Verified code generation and verification benchmarks.}
\textsc{Clover} \citep{sun2024clover} accepts generations only when a verifier confirms mutual consistency among code, documentation, and a Dafny specification. \textsc{Verina} \citep{ye2025verina} jointly benchmarks code, specification, and proof generation in Lean, separating soundness from completeness as we do. \textsc{DafnyBench} \citep{loughridge2025dafnybench} and multi-language vericoding suites \citep{bursuc2025vericoding} measure whether models can produce verifying artifacts. A complementary thread develops proof automation: whole-proof generation and repair \citep{first2023baldur}, sound LLM-in-the-loop frameworks \citep{wu2024lemur}, retrieval-augmented proving \citep{yang2023leandojo}, self-improving verified translation \citep{aggarwal2025alphaverus}, and agentic proof generation \citep{yang2025autoverus,chen2025safe,mugnier2025laurel,misu2024towards}. AutoRocq \citep{tu2026autorocq} iteratively refines proofs via the Rocq theorem prover; Inductive Deductive Synthesis \citep{agarwal2026ids} jointly synthesizes implementation and proof for distributed systems $200\times$ faster than expert effort.
Project-level efforts target real verified codebases such as an OS microkernel \citep{zhang2024selene}, while \citet{yang2025verusage} study 849 real proof tasks from open-source Verus/Rust systems and \citet{sosso2026agenticproving} show that agentic provers nearly saturate existing Lean~4 benchmarks and motivate harder suites like ours.
Our verifier-feedback loop instantiates iterative repair from execution feedback \citep{chen2024selfdebug,madaan2023selfrefine}, with a sound verifier supplying the signal. \citet{shefer2025mainstream} evaluate LLMs producing verified code in Dafny, Verus, and Nagini on HumanEval-derived tasks; \citet{feng2026certified} validate generated specifications before synthesizing certified code against a Lean-embedded verifier. 
All of these works synthesize programs from curated descriptions.
Closest to our setting, \textsc{Vero}~\citep{ye2026vero} evaluates implementation and proof synthesis at the repository level in \textsc{Lean}~4, over $43$ instances sourced from real Dafny, Verus, Coq, and Python projects. Each instance is a hand-written \textsc{Lean} re-formalization of the upstream code rather than the repository itself, and the agent fills a fixed scaffold instead of resolving a reported issue. Its specifications are supplied and frozen, so specification synthesis is out of scope, and its audit certifies that they are satisfiable rather than that they capture the intended behavior.
\benchmark differs from these works as it is built on fixes to real GitHub issues in code repositories with rich library dependencies.

\paragraph{Specification synthesis from informal intent.}
Whether a specification faithfully captures intent is a recognized hard problem. \citet{endres2024nl2postcond} measure postconditions' \emph{discriminative power} to reject buggy behavior; \citet{lahiri2024evaluating} proposes symbolically testing specifications for quality; \citet{richter2025beyond} evaluate contracts by their ability to separate buggy from correct implementations; and \citet{ma2025specgen} generate verifiable specifications with conversational refinement. SpecSyn \citep{ma2026specsyn} strengthens specifications via mutation-based semantic discrimination, while \citet{chen2026codespecbench} find that the best frontier LLM attains only a pass rate of 20.2\% on repository-scale specification generation, corroborating our finding that specification synthesis is the binding bottleneck.
The challenge echoes classical results: dynamically-inferred invariants are unsound \citep{ernst2007daikon}, learned loop invariants are often non-inductive \citep{chakraborty2023ranking,kamath2023finding}, and specifications can be satisfied vacuously \citep{kupferman2003vacuity}. The problem parallels autoformalization, where faithfulness is enforced through round-trip checks \citep{wu2022autoformalization, jiang2023draft,li2024autoformalize,xin2024deepseekprover}. These works study faithfulness at the level of a single function or postcondition; the real-world setting adds a distinctly harder demand, namely that a specification cover the full behavioral surface of a fix spanning many functions and files. \benchmark makes this coverage requirement an explicit correctness criterion and measures the resulting faithfulness gap on real-world software issues, finding it to be the dominant obstacle to end-to-end verified issue resolution.

\section{Specification languages and verifiers}
\label{app:backends}

Every instance is built four times, in four specification languages with four
different notions of what it means to be checked. Three are machine-checkable
and one is not. The fourth separates two effects that are easy to conflate: the
benefit of giving a solver a precise statement of the intended behavior, and the
benefit of that statement being mechanically enforceable. Without a prose
backend, every result on the formal backends is confounded: a specification is
also a very good problem description.

\cref{tab:backends} states what each backend checks and what it costs.

\begin{table}[htbp]
\centering
\caption{The four specification backends. ``Trust base'' is what must be correct
for a passing check to mean anything. Verification time is the wall clock of a
single successful check of a released bundle on the \sbv corpus. Axioms are the
total number of assumed contracts for unchanged callees across all $500$
instances.}
\label{tab:backends}
\footnotesize
\setlength{\tabcolsep}{6pt}
\begin{tabular}{@{}L{3.0cm}L{2.2cm}L{2.2cm}L{2.2cm}L{2.2cm}@{}}
\toprule
& \textbf{\nagini} & \textbf{\velvet} & \textbf{\lean} & \textbf{\ears} \\
\midrule
Notation
& Python with contract annotations
& Imperative language embedded in Lean~4
& Lean~4 with Mathlib
& Structured English \\
Statement form
& Pre\slash postconditions, permissions, invariants
& Method contracts with mutable parameters
& A specification predicate and a theorem
& Requirements with acceptance criteria \\
Discharged by
& Translation to an intermediate verification language, then SMT
& An SMT-backed tactic inside Lean
& The Lean kernel
& \na \\
Trust base
& Translation, verification back end, SMT solver
& Embedding semantics, tactic, SMT solver
& Lean kernel and its three classical axioms
& Human reading \\
\midrule
Median length (lines) & 151 & 100 & 107 & 106 \\
Verify, mean / median (s) & 25.8 / 18.4 & 6.0 / 5.2 & 10.5 / 9.1 & \na \\
Verify, p95 / max (s) & 58.2 / 288.6 & 11.9 / 33.7 & 17.7 / 84.9 & \na \\
Axioms, $500$ instances & 264 & 80 & 3 & \na \\
Termination & Explicit measures ($443$ bundles) & Total-correctness semantics & Structural, by construction & \na \\
\bottomrule
\end{tabular}
\end{table}

\subsection{Contract-annotated Python}
\label{app:backend-nagini}

The first backend states specifications in Python itself, annotated with
contracts, and verifies them by translation to an intermediate verification
language and from there to an SMT solver. A specification declares the functions
the fix touches with their real signatures, states preconditions and
postconditions over their arguments and results, and supplies a reference body
that the verifier checks against those contracts.

Two features of this backend shape how the corpus looks. The first is that it
reasons about mutable heap state with explicit permissions: a specification must
say which fields it may read and write, and a postcondition about a field is
meaningless without the accompanying permission. The backend can therefore
describe the kind of behavior \sbv issues are actually about: in-place mutation
of an object, aliasing between two references, a method that changes one field
and must leave another alone. The cost is considerable annotation. The second is
that termination is not free. A specification states an explicit measure that
must decrease, and $443$ of the $500$ released bundles do so, at a mean of $2.8$
measures per bundle. Without it the verifier would prove postconditions only
under partial correctness.

The consequence is that this is the most expensive backend to write in and the
most expressive about state. Its modules are the longest ($151$ lines at the
median), carry the most contracts (a mean of $6.8$ preconditions and $12.9$
postconditions), need the most loop invariants ($264$ bundles, a mean of $10.7$
where present), and take the longest to check ($25.8$\,s on average, up to
$288.6$\,s). It is also the backend that needs the most axioms, $264$ over $137$
instances, because anything it cannot express in its verification fragment must
be summarized rather than modeled.

\paragraph{Exceptional behavior.}
Many repository fixes are about what happens when something fails, so the
specification language must be able to state it. This backend distinguishes the
normal postcondition from the postcondition that holds when a particular
exception propagates, and the corpus uses that distinction: a specification can
require that a failing operation leaves the object exactly as it found it, which
is a guarantee about the absence of a partial effect. Box~\ref{box:axiom} shows
the pattern in an axiom, where the store's refusal to remove a row is an
outcome the specification allows but no implementation may arrange.

\subsection{An imperative language inside a proof assistant}
\label{app:backend-velvet}

The second backend writes the fix in a small imperative language that is
embedded in Lean~4. A specification declares a method with typed parameters,
marks the ones it mutates, states preconditions, postconditions, and loop
invariants, and closes with a directive that asks the tooling to prove the method
correct against its contract. The proof obligations are generated from the
embedding's semantics and discharged by an SMT-backed tactic, so a specification
in this language is a Lean object whose correctness claim is discharged
automatically rather than by hand.

The result is the lightest of the three formal backends to write in and the
cheapest to check: a median of $100$ lines, a mean of $1.23$ methods with $2.32$
postconditions, and a mean verification time of $6.0$\,s. Its proof regions are
the shortest of the three (a median of $4$ lines) because the tactic does the
work. Its pure definitions are the longest (a median of $32$ lines), since the
interesting content moves into the functional model that the contracts are
stated over.

The characteristic modeling move in this backend is to \emph{parameterize} rather
than to axiomatize. Where the contract-annotated Python backend states an assumed
contract for an unchanged callee, this backend often takes the callee's result as
an input to the method and states the specification relative to it. In
\mbox{\id{django-11179}}, for example, the number of rows removed by the unchanged
deletion primitive is simply a parameter, and the specification says that the
identifier is cleared and the reported count equals that parameter. That is a
weaker statement: it does not pin what the primitive itself does. This backend
therefore needs $80$ axioms where the first needs $264$.
\Cref{app:where-it-binds} discusses what this trade costs.

\subsection{Dependent type theory over a mathematical library}
\label{app:backend-lean}

The third backend states the specification in Lean~4 with Mathlib, as a
predicate over inputs and outputs, gives a functional implementation, and proves
a theorem that the implementation satisfies the predicate for all inputs. There
is no verification condition generator and no solver: the proof is a term the
Lean kernel checks.

This is the strictest trust base of the three, and the corpus is measured against
it directly. Across the released bundles no proof anywhere contains a
\texttt{sorry}, the entire $500$-instance corpus declares $7$ local axioms, and
an audit of what the theorems actually depend on finds only Lean's three standard
classical axioms (propositional extensionality, cited by $659$ theorems;
soundness of quotients, $501$; and choice, $127$) plus eleven instance-specific
declarations that are themselves recorded as axioms in the bundles that use
them.

The characteristic modeling move here is \emph{universal quantification over the
unchanged}. Where the other backends assume a contract for a callee they do not
model, this backend can leave the callee as a variable and prove the theorem for
every possible instantiation of it. In \mbox{\id{django-11179}} the decision of whether
an object qualifies for the fast deletion path is an unchanged predicate. The
specification quantifies over all predicates and proves the outcome for each.
The backend therefore needs $3$ axioms in $500$ instances: a callee that can be
abstracted does not need to be assumed.

The cost of that strictness appears elsewhere. The backend has no native account
of mutable heap state, so a fix about in-place mutation must be recast as a
function from a value to a value, and the recasting is the hard part of writing
these bundles. It is also the backend whose modeled site most often differs from
the patch's own file set, because the recasting frequently pulls in the caller
that owns the data being transformed.

\subsection{Structured natural language}
\label{app:backend-ears}

The fourth backend states the intended behavior in structured English, using a
constrained requirements syntax in which each acceptance criterion follows one of
a small number of sentence patterns: a trigger and a required response
(\textsc{when}\,\ldots\,\textsc{then}\,\ldots\,\textsc{shall}), an unconditional
obligation (\textsc{shall}), an undesired condition and its handling
(\textsc{if}\,\ldots\,\textsc{then}), or a continuous state
(\textsc{while}). Each requirement is introduced by the developer-facing need it
serves and then decomposed into numbered criteria.

The released documents are as long as the formal ones and considerably more
readable: a median of $106$ lines, a mean of $3.7$ requirements and $14.7$
acceptance criteria, with $17.8$ obligations stated per instance. Of the $7{,}362$
criteria in the \sbv corpus, $57.3\%$ are trigger-and-response, $28.4\%$ are
unconditional, $11.8\%$ describe an undesired condition, $1.9\%$ describe a
continuous state, and $0.6\%$ use no pattern keyword at all.

These documents pass the same adversarial audit for faithfulness and
non-disclosure as the formal bundles, but they cannot be verified or checked
mechanically against a pre-fix twin. A solver given one of these documents
receives a precise account of the intended behavior and no way to check its own
work, which is the condition against which the formal backends' verification
feedback is measured. The
requirements are also noticeably \emph{broader} than their formal counterparts.
Because prose costs nothing to quantify over, these documents routinely state
regression obligations that a formal specification would have to model in order
to mention: that signals still fire, and that unrelated deletion behaviors are
preserved.

\subsection{What each language can and cannot say}
\label{app:where-it-binds}

The four backends yield four different specifications, and on the same task
they pin different things.

\paragraph{The same fix, four times.}
\mbox{\id{django-11179}} asks that a single dependency-free object being fast-deleted
have its in-memory primary key cleared, as the general deletion path already
does. It is a one-line patch. The four specifications of it differ as follows.

The contract-annotated Python bundle models the removal as an operation against a
database handle with an explicit entitlement to issue exactly one row-removal
statement. Its postconditions pin five separate facts: that one statement was
issued, that it targeted the identifier the object carried on entry, that the
returned count is the one \emph{that statement} reported rather than a number
obtained by asking the store again, that the in-memory identifier is cleared, and
that on a refusal the object is left exactly as it was found. Nothing about the
store's own tallies is readable by the implementation, which rules out
implementations that pay a database round trip per read.

The \velvet bundle states the same fix in three lines of contract: the identifier
is cleared, and the reported count equals the count it was given. The unchanged
removal is a parameter.

The \lean bundle models the collector's data as groups of records, defines the
outcome as ``every record's key cleared'' plus a branch-selected tally, and
proves that its implementation meets that specification for every possible
fast-deletability predicate.

The \ears bundle states three requirements and thirteen acceptance criteria,
including two that no formal bundle states: that the reset must use the model's
actual primary-key attribute name so that renamed primary keys work, and that
clearing the key must not issue an additional query.

All four are admitted, and none of them is implied by another.

\paragraph{Where the backends agree.}
Disagreement about how to state a fix does not imply disagreement about what the
fix is about. On the $289$ \sbv instances where all three formal backends record a
resolvable modeled location, $244$ ($84.4\%$) model exactly the same set of
repository files, the median Jaccard overlap of the three file sets is $1.0$, and
pairwise agreement runs from $87.0\%$ to $88.4\%$. On $364$ of the $500$
instances all three agree on how many functions need modeling. The independent
construction of three formal twins therefore converges on the same site and
differs on how to describe it.

\paragraph{Carriers.}
The remaining differences are largely about representation. Every specification
must choose a carrier for values the language cannot represent natively, and the
choices differ systematically: the contract-annotated Python bundles carry
strings as sequences of integer code points ($144$ bundles), as an interned
identity token ($91$), or as an explicit integer sequence ($262$ use a sequence
carrier of some kind), and declare in $186$ cases that the tested behavior does
not depend on string content at all. The \lean-based bundles, having a real
character type available, carry strings as lists of characters ($140$ bundles)
and reach for rationals ($43$) where the Python backend needs an interpreted real
encoding ($11$ bundles). None of these choices is free: a specification that
treats strings opaquely cannot state a property about their content, and the
provenance map records the decision so that a reader can see which properties
were placed out of scope.

\paragraph{What none of them can say.}
All four languages describe the values a program computes. None describes the
names it computes them under, the module a symbol is imported from, the class a
value is an instance of, or an effect visible only in process or filesystem
state. When a task's graded test discriminates on one of those, no specification
in any of the four languages can separate the fix from the bug, and the instance
is refused. \Cref{app:pro-nongreen} discusses this boundary; all three formal
backends run into it on exactly the same $22$ \swepro tasks.

\section{Construction and admission}
\label{app:gates}

A bundle is drafted by agents that are allowed to see everything, then attacked by agents that are allowed to see almost
nothing, then admitted only when thirteen independent checks all report a
positive or explicitly inapplicable outcome. This section describes that
pipeline: what is built and by whom (\cref{app:construct}), what the mechanical
checks actually establish and at what cost (\cref{app:mechanical}), how the
adversarial audit is run and adjudicated (\cref{app:audit}), and how a recorded
outcome is kept current as the checks themselves change
(\cref{app:reaudit}).

Two design commitments run through all of it. The first is that
\emph{generation is adversarial and adjudication is mechanical}. Every attack on
a specification is produced by an agent whose instructions are to break the
bundle, and then resolved by running the verifier, the repository, or the task's
own test suite. An attack takes one of three forms: a wrong implementation that
might slip through, an axiom that might be false of the real callee, a scenario
the specification might mis-describe. An attack counts only if its artifact,
when run, breaks the bundle.

The second is that \emph{inapplicable is not the same as passing}. A check that
does not apply to a bundle records that fact and its reason. This makes the
tables in this section read differently from the usual pass-rate summary: the
denominator of a check is the set of bundles it applies to.

\subsection{How a bundle is constructed}
\label{app:construct}

Construction runs in four stages, with an information asymmetry between the first
and the third: the agents that write a specification see the gold patch, and the
agents that attack it do not.

\paragraph{Stage one: modeling.}
A construction agent receives the task's problem statement, the repository at the
frozen base commit, the gold patch, and the task's own test suite. It decides
what to model (which functions the fix touches, what their arguments and results
are, what carrier each value needs in the target language) and writes the
specification module, the reference implementation, and the pre-fix twin.
The verifier is in the loop throughout, so the stage ends with an object that
verifies and a twin that does not. It also ends with the two documents that carry
the transfer to the repository: the operation-by-operation correspondence and the
provenance record of every modeling decision (\cref{app:central-pair}).

\paragraph{Stage two: proof.}
The kernel-checked backend needs a proof term, not a solver call, so a separate
stage searches for one. The output is a closed proof or nothing: a proof that
merely elaborates while leaving its obligation unfinished is not accepted, and
the check in \cref{app:mechanical} that looks for this is the one that separates
a file that compiles from a theorem that is proved. Across both released corpora
no proof anywhere is left open.

\paragraph{Stage three: attack.}
The bundle is then handed to four independent adversarial auditors, described in
\cref{app:audit}. They do not receive the gold patch. Each has a single
assignment: break the transfer, break the specification's strength, break its
agreement with the issue, or recover the fix from the view a solver would see.
Each produces concrete artifacts rather than judgments: a candidate
implementation, a concrete input, a scenario. The artifacts are adjudicated by
running them.

The output of this stage is the property test suite, the axiom probes, and the
differential harness of \cref{app:evidence}, together with a written record of
every attack that was attempted and the reason it failed. That record makes the
corpus re-auditable rather than merely re-runnable (\cref{app:reaudit}).

\paragraph{Stage four: admission.}
Finally the thirteen checks are run and their outcomes recorded. A negative
outcome does not produce a patch to the bundle; it sends the instance back for revision at
whichever stage owns the defect. A specification that admits a wrong
implementation goes back to stage one, not to stage three. The alternative,
adding a precondition until the attacker's counterexample stops verifying, would
fit the specification to the attack instead of the task.

\paragraph{Refusal.}
Some instances cannot be built, and refusal is a recorded outcome with a stated
reason rather than a silent omission. Of the $1{,}064$ candidate bundles on
\swepro, $68$ are refused, and their reasons fall into a small number of
categories: the change has no value-level footprint at all (a rename, a
relocation, an import cleanup), the change is behavior-preserving so no
differential witness exists that could separate it from the code it replaces, or
the change's only observable effect is an irreversible external side effect that
the specification languages cannot describe. \Cref{app:pro-nongreen} works
through the $22$ \swepro tasks that no prover backend can model. The remaining refusals
are per-backend: an instance can be refused under one language and admitted under
another, and $2$ further \swepro tasks are in that position, so $24$ of the $266$
carry at least one refusal.

\paragraph{What construction cost.}
\cref{tab:effort} totals the machine work behind the two released corpora.
Verifying the specifications takes under nine hours of prover time in total, so
the figures are dominated by trying to refute them. Driving the property
test suites takes an order of magnitude longer than verifying every released
bundle, and the differential and axiom evidence is measured in hundreds of
millions of executions of real repository code.

\begin{table}[htbp]
\centering
\caption{Machine effort behind the released corpora, pooled over both and over
all backends. ``Volume'' counts the units of work admission performed, and
excludes the verifier calls made while a specification was still being drafted.
Prover time is totalled for the two activities that are pure verification; a dash
means the activity was not timed separately. Wall-clock totals are serial prover
time; admission itself runs in parallel.}
\label{tab:effort}
\small
\setlength{\tabcolsep}{6pt}
\begin{tabular}{@{}lrr@{}}
\toprule
\textbf{Activity} & \textbf{Volume} & \textbf{Prover time} \\
\midrule
Verifying the released specifications & $2{,}233$ verdicts & $8.4$\,h \\
Re-failing the pre-fix twins & $2{,}622$ twins & \na \\
Driving the property test suites & $23{,}482$ cases & $87.0$\,h \\
Killing mutants of the reference bodies & $25{,}457$ mutants & \na \\
\midrule
Probing axioms against real callees & $25.7$\,M observations & \na \\
Differential inputs against patched repositories & $187.6$\,M inputs & \na \\
Resolving the equivalent patches & $3{,}026$ harness runs & \na \\
\midrule
Adversarial audit ballots & $12{,}000$ role reports & \na \\
Recorded admission outcomes & $48{,}227$ records & \na \\
\bottomrule
\end{tabular}
\end{table}

\subsection{The mechanical checks}
\label{app:mechanical}

Thirteen checks decide admission. \cref{tab:gates} lists them with what each one
establishes and how often it applies. Eight are decided by running something (a
verifier, a repository, a test suite), three by reading the bundle statically,
and two by the adversarial audit of \cref{app:audit}. None of them involves a
judgment call; the two that rest on
adversarial work read a verdict that was already adjudicated mechanically, and
they are the subject of \cref{app:audit}.

\begin{table}[!t]
\centering
\caption{The thirteen admission checks, pooled over both corpora and all four
backends ($3{,}064$ candidate bundles). ``Applies'' is the number of bundles the
check can speak about; ``inapplicable'' is the number where it explicitly cannot,
almost always because the check is meaningless for that backend (the prose
backend has no verifier) or for that bundle (no axioms to probe, no operators to
mutate). ``Decided by'' distinguishes a static reading of the bundle, an
execution inside the task's official container, and a verifier verdict.}
\label{tab:gates}
\small
\setlength{\tabcolsep}{6pt}
\begin{tabular}{@{}lL{5.9cm}lrr@{}}
\toprule
\textbf{Check} & \textbf{What a positive outcome establishes}
& \textbf{Decided by} & \textbf{Applies} & \textbf{Inapp.} \\
\midrule
Verification & The released specification module verifies, and every correctness
obligation in it is discharged & verifier & $2{,}233$ & $784$ \\
Discrimination & Every pre-fix twin fails verification under the byte-identical
specification & verifier & $2{,}233$ & $775$ \\
Gold agreement & The implementation that mirrors the gold patch verifies, and the
twins still fail against it & verifier & $2{,}233$ & $768$ \\
Mutation & Systematic mutants of the reference body are rejected at or above the
backend's floor & verifier & $2{,}133$ & $882$ \\
Property test suite & Every case in the property test suite receives the verdict it is labeled
with & verifier & $2{,}231$ & $784$ \\
\midrule
Resolution & The equivalent patch applies at the base commit and passes the
task's hidden tests & container & $3{,}001$ & $5$ \\
Axiom soundness & Every axiom holds of the real callee on generated and boundary
inputs & container & $1{,}896$ & $1{,}127$ \\
Conformance & The reference implementation and the patched repository function
agree on a large generated input set & container & $2{,}093$ & $934$ \\
\midrule
Task identity & The frozen task row is unmodified and the bundle adds only
permitted files & static & $3{,}016$ & $0$ \\
Hygiene & Region markers are intact, the solver-facing view is stable, and no
prohibited construct appears & static & $3{,}000$ & $3$ \\
Disclosure & The solver-facing view does not reveal the fix
& static & $2{,}930$ & $10$ \\
\midrule
Refinement fidelity & An adversarial reading of the correspondence found no
unjustified step & audit & $3{,}000$ & $2$ \\
Attacker panel & Three blind attackers found no weakness, no unfaithfulness, and
no disclosure & audit & $3{,}002$ & $0$ \\
\bottomrule
\end{tabular}
\end{table}

Of the $3{,}064$ candidate bundles, $3{,}002$ carry all thirteen outcomes; the
$62$ that do not are refused \swepro bundles that never reached admission, $27$
of which carry no outcome at all. On the $2{,}996$ admitted bundles every
applicable check is positive.

\paragraph{Verification, and what it costs.}
\cref{tab:verifytime} gives the wall-clock cost of a single successful check of a
released bundle. On \sbv the three formal backends differ by a factor of about
four in the mean and five at the 95th percentile, and the ordering is the one the
languages predict: reasoning about mutable heap state through an intermediate
verification language and an SMT solver is the most expensive, a solver-backed
tactic over a small imperative embedding is the cheapest, and kernel checking of
an already-found proof sits between them.

\begin{table}[htbp]
\centering
\caption{Verification wall clock for one successful check, in seconds. Every
bundle whose verification passes contributes one measurement, which on \swepro is
a few more bundles than are released, since a bundle can verify and still be
refused on another check. The totals are serial prover time.}
\label{tab:verifytime}
\small
\setlength{\tabcolsep}{5pt}
\begin{tabular}{@{}llrrrrrr@{}}
\toprule
\textbf{Corpus} & \textbf{Backend} & \textbf{$n$} & \textbf{Mean}
& \textbf{Median} & \textbf{p90} & \textbf{p95} & \textbf{Max} \\
\midrule
\sbv & \nagini & 500 & 25.8 & 18.4 & 48.0 & 58.2 & 288.6 \\
     & \velvet & 500 & 6.0 & 5.2 & 9.5 & 11.9 & 33.7 \\
     & \lean & 500 & 10.5 & 9.1 & 12.9 & 17.7 & 84.9 \\
\midrule
\swepro & \nagini & 244 & 29.4 & 27.4 & 40.6 & 46.0 & 170.0 \\
        & \velvet & 245 & 3.6 & 2.5 & 4.4 & 5.8 & 144.5 \\
        & \lean & 244 & 4.6 & 4.3 & 5.8 & 6.7 & 9.7 \\
\midrule
Both & all three & $2{,}233$ & 13.6 & 8.4 & 30.2 & 39.0 & 288.6 \\
\bottomrule
\end{tabular}
\end{table}

\begin{figure}[htbp]
\centering
\includegraphics{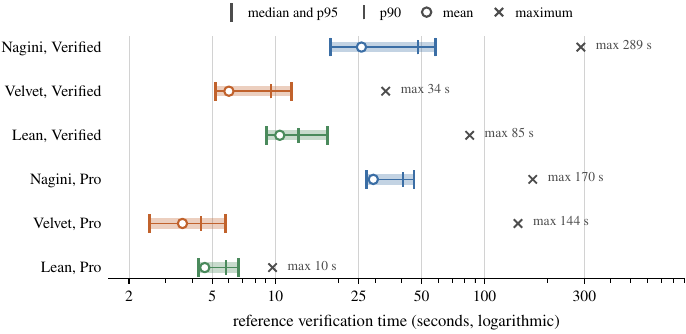}
\caption{The same measurements as \cref{tab:verifytime}, drawn on one
logarithmic scale. Each row is a backend on a corpus. The band spans the median
to the 95th percentile, the tick marks give the median, the 90th, and the 95th
percentile, the open circle is the mean, and the cross is the slowest single
bundle in that cell. The three formal backends are ordered the same way on
both corpora, and the two Lean-based backends move \emph{left} on \swepro even
though \swepro fixes are ten times larger.}
\label{fig:verify-time}
\end{figure}

\swepro tasks are ten times larger than \sbv tasks by patch size, and yet under
the two Lean-based backends they verify \emph{faster}
(\cref{fig:verify-time}), because the specifications written for them are more
abstract. A larger fix forces more of the surrounding behavior to be
parameterized rather than modeled, and a parameterized specification generates
smaller proof obligations. Under the contract-annotated Python backend, where the
surrounding state has to be modeled to be talked about at all, the expected
direction reappears: $29.4$\,s on \swepro against $25.8$\,s on \sbv, with a
median that rises by half.

The tail is dominated by a small number of bundles. Under the contract-annotated
Python backend the slowest released bundle takes $288.6$\,s where the median
takes $18.4$\,s, a factor of sixteen, and the p90-to-p95 step is steep, so a
tight fixed timeout would act as a filter on which specifications can be
admitted. A solver working under a per-call verifier budget works against this
distribution; \cref{app:verif-effort} reports what it does with it.

\paragraph{Discrimination and gold agreement.}
These two checks keep a merely consistent specification out of the release. The
first requires every pre-fix twin to fail under the exact specification the
reference implementation passes under; $1{,}709$ twins across the \sbv corpus
must fail, and all of them do. The second requires that the implementation
mirroring the gold patch also verifies and that the twins still fail against it,
so the specification is satisfied by the repository's own answer and not only by
ours. Where a fix repairs several independent behaviors the bundle ships one twin
per behavior, so these checks are not one bit of information per bundle: $81$
\nagini bundles carry two or more.

\begin{figure}[htbp]
\centering
\includegraphics{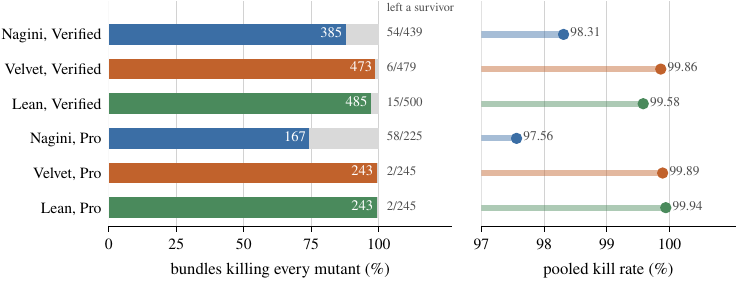}
\caption{The mutation check by backend and corpus, splitting the totals of
\cref{tab:mutation}. \textbf{Left:} the share of bundles that reject every
mutant generated for them; the number inside each bar is how many bundles those
are, and the number beside it is how many of the cell's bundles left at least
one survivor. \textbf{Right:} the pooled rejection rate over all
mutants of the cell, on an axis that begins at $97\%$; the per-bundle admission
floor of $80\%$ is far below the bottom of this range. Survivors are
concentrated in one backend and one corpus.}
\label{fig:mutation}
\end{figure}

\paragraph{Mutation.}
\cref{tab:mutation} reports the mutation check. It generates systematic
single-point mutants of the reference implementation's body, holding the
specification fixed, and requires the verifier to reject them. Pooled over both
corpora, $25{,}163$ of the $25{,}374$ judged mutants are rejected, a rate of
$99.17\%$ ($83$ \nagini mutants that do not parse are excluded), and $1{,}996$ of
$2{,}133$ eligible bundles leave no surviving mutant.
\cref{fig:mutation} splits the same counts by corpus.

\begin{table}[htbp]
\centering
\caption{The mutation check. ``Mutants'' includes $83$ \nagini mutants that do
not parse and are excluded from the rate. ``At $100\%$'' counts bundles with no
surviving mutant. ``Survivors'' counts mutants that verified against
the frozen specification. The admission floor is $80\%$ per bundle for all
backends; the pooled rate is far above it.}
\label{tab:mutation}
\small
\setlength{\tabcolsep}{5pt}
\begin{tabular}{@{}lrrrrrr@{}}
\toprule
\textbf{Backend} & \textbf{Bundles} & \textbf{Mutants} & \textbf{Rejected}
& \textbf{Pooled rate} & \textbf{At $100\%$} & \textbf{Survivors} \\
\midrule
\nagini & 664 & $8{,}973$ & $8{,}713$ & $98.01\%$ & 552 & 177 \\
\velvet & 724 & $7{,}847$ & $7{,}837$ & $99.87\%$ & 716 & 10 \\
\lean & 745 & $8{,}637$ & $8{,}613$ & $99.72\%$ & 728 & 24 \\
\midrule
\textbf{Total} & $\mathbf{2{,}133}$ & $\mathbf{25{,}457}$ & $\mathbf{25{,}163}$
& $\mathbf{99.17\%}$ & $\mathbf{1{,}996}$ & $\mathbf{211}$ \\
\bottomrule
\end{tabular}
\end{table}

The floor is $80\%$ rather than $100\%$, and it bounds specification tightness
rather than measuring correctness. A surviving mutant means the specification does
not distinguish the mutated body from the reference one. Sometimes that is a real
weakness. Sometimes it is not: a mutant that changes a value the specification
deliberately does not constrain (a subexpression whose result is discarded, an
operation on a field the postconditions leave free) is
indistinguishable, and a specification that killed it would be over-fitted to one
way of writing the answer. Demanding $100\%$ would therefore reject sound
bundles, and it would reject them preferentially on the backends where the proof
is found by search rather than supplied, since there a survivor can also mean the
search did not find the refutation. The tighter check on specification strength is
the discrimination lane of the property test suite and the attacker panel, both of which use
targeted wrong implementations rather than random perturbations.

The per-backend spread reflects the languages. The contract-annotated Python
backend has about five times the survivor count of the other two combined, on a
comparable number of mutants. Its
specifications talk about mutable state through explicit permissions, which
leaves more of the reference body outside what the postconditions constrain; the
two Lean-based backends state their obligations over pure values, where almost
every operation in the body feeds the result. The released bundles clear the
floor, and their survivors were reviewed individually.

$882$ bundles are inapplicable, of which $766$ are prose bundles with no verifier
to run and $116$ have reference bodies with nothing to mutate. A body that
performs a single assignment offers no operator to perturb. Those bundles carry
the burden on the twin and the property test suite instead.

\paragraph{Resolution.}
The equivalent patch must apply at the frozen base commit and pass the task's
hidden tests in the task's own official container. All $2{,}996$ admitted bundles
resolve. Twenty-three needed more than one attempt and are flagged as flaky, all
of them on \sbv, distributed as $10$ under the contract-annotated Python backend,
$9$ under the embedded imperative language and $4$ under the kernel-checked
backend. The flakiness is in the repository's own test suite, not in the
specification; the same patch resolves on retry. The flag is recorded so that a
downstream user comparing evaluation results against these instances knows which
ones have a noisy oracle. No \swepro bundle needed a retry.

\paragraph{Axiom soundness.}
An axiom is the one place in a bundle where something is asserted rather than
proved, so each is tested against the running repository. $1{,}893$ probe runs
constructed inputs, called the real callee inside the task's official container,
and compared the observed behavior against what the axiom claims. $428$ runs
concentrated on boundary inputs ($428{,}609$ cases) and $526$ ran generated
inputs at volume ($20.6$\,M cases), for $25.7$\,M recorded observations of real
callee behavior in total. $164$ runs were performed against the patched tree
rather than the base commit, which is the correct choice when the probe needs a
data structure the fix introduces. No probe refutes its axiom on any admitted
bundle.

$1{,}127$ bundles are inapplicable, and $815$ of the runs report that the bundle
they were built for declares no axioms at all. That is the expected outcome under
the kernel-checked backend, where a callee that can be abstracted does not need
to be assumed (\cref{app:axioms}).

\paragraph{Conformance.}
The conformance check compares the reference implementation against the real
patched repository function on a large generated input set, inside the task's
container, with a pinned seed. It is a refutation instrument: it cannot prove
that the model is faithful, but a single disagreement proves it is not. Across
$2{,}093$ recorded verdicts there are no disagreements anywhere in either
corpus, over $187.9$\,M compared inputs, and $1{,}461$ verdicts individually
confirm at least $10^5$ inputs.

What the check establishes differs by backend. Under the contract-annotated
Python backend the reference implementation is Python, so it is executed directly
against the repository function: $690$ verdicts are of this kind. Under the two
Lean-based backends the reference implementation is not executable in the
repository's language, so the comparison runs against a transliteration of the
model into Python: $1{,}402$ verdicts are of this kind. A
transliterated comparison tests the model as transliterated, which is a weaker
claim, and it is recorded as such rather than pooled with the direct
comparisons. $934$ bundles are inapplicable: $766$ prose bundles, $146$ with no
executable reference region to compare, $10$ blocked bundles, $6$ whose model
uses a construct the comparison cannot represent, and $6$ others.

\paragraph{The property test suite.}
Every case in every property test suite receives the verdict it is labeled with:
$23{,}482$ of $23{,}482$, across $2{,}231$ bundles, with no failures in either
corpus. \cref{tab:suite} gives the per-backend breakdown by lane.

Driving the property test suites is the single most expensive part of admission. The
$23{,}482$ cases consume $87.0$ hours of serial prover time at a mean of
$13.3$\,s and a median of $8.1$\,s per case. The slowest case, at $2{,}765$\,s,
is a wrong implementation that the contract-annotated Python backend's solver had
to work hard to reject. That single case costs more prover time than verifying
every released \velvet bundle in the \swepro corpus.

\paragraph{Task identity, hygiene, and disclosure.}
Three static checks guard properties that would be easy to violate by accident.

Task identity requires that the frozen task record is byte-identical to the
upstream row and that the bundle adds only permitted files. It rules out
admission against an edited task: a relaxed test list, a different base commit,
an amended problem statement.

Hygiene is a structural reading of the bundle: that the region markers which
define the solver-facing view are intact and in the right order, that projecting
the view twice gives the same result as projecting it once, that no module
outside a fixed permitted set is imported, and that no construct appears which
would let a specification be satisfied vacuously. The last of these is
per-backend, because each language admits vacuity differently: a
contract-declaration form used outside the axiom region, a missing termination
directive in the embedded imperative language, a divergence-admitting construct
or an unfinished proof in the kernel-checked one. View stability does the most
work in practice, since the view is what a solver sees.

Disclosure screens that view for content that would reveal the fix. $2{,}930$
bundles pass it outright. $76$ carry a documented relaxation, meaning the view
discloses more about the intended fix than the problem statement alone does. The
relaxation is written down and the affected instances are listed in
\cref{app:leakage}. A solver's patch is still graded by the repository's own
tests. Ten bundles are recorded as inapplicable and one as failing the screen;
none of them is in the release.

\subsection{The adversarial audit}
\label{app:audit}

The mechanical checks establish that a specification is internally consistent,
tight against perturbation, and true of the repository on the inputs anyone
tried. They cannot establish that it is about the right thing. Four properties
resist mechanization because they are claims about the relationship between a
formal object and an informal one: that the transfer from the verified model to
the executed patch is sound, that the specification is strong enough to be worth
verifying, that it agrees with what the issue asked for, and that it does not
give the answer away. Each is assigned to a blind adversarial auditor.

\paragraph{The four roles.}
The \emph{fidelity} role attacks the transfer. It reads the correspondence map
and the provenance record and looks for a step that is neither an identical
operation, nor a type refinement from a modeled carrier to a concrete repository
type, nor a call to an axiomatized callee. A step that fits none of the three is
a hole in the argument that verifying the model says anything about running the
patch, and the role's job is to find one and exhibit the behavior it permits.

The \emph{strength} role attacks the specification. Its assignment is to write an
implementation that a reader of the issue would call wrong and that nonetheless
verifies against the frozen specification. The role delivers a program, and the
program is run.

The \emph{faithfulness} role attacks the agreement between the specification and
the issue. Its assignment is to describe a concrete scenario in which the
specification and the issue's stated intent come apart: an input on which the
specification demands behavior the issue does not ask for, or permits behavior
the issue forbids. This is the hardest of the four, and it is the property that
dominates failure when solvers write specifications themselves
(\cref{app:refutation}).

The \emph{disclosure} role attacks the view. It receives only what a solver in
the specification-provided setting would receive and tries to reconstruct the
fix from it. A successful reconstruction is a finding whether or not the
mechanical screen of \cref{app:mechanical} flagged anything, because the leak it
finds is semantic rather than lexical.

\paragraph{Blindness, and what it means.}
The attacking agents do not receive the gold patch, and they run separately from
construction. An agent that has seen the intended fix cannot judge whether a
specification discloses it, and an agent that wrote a specification is a poor
judge of whether it is faithful to an issue it has already interpreted. Each role
runs against the bundle as committed, with the reference implementation withheld
from the roles for which seeing it would be disqualifying.

\paragraph{Adjudication.}
A role's report is recorded but never taken as a verdict on its own. When the
strength role produces a candidate implementation, that implementation is run
against the frozen specification, and the outcome, not the report, decides the
check. When the fidelity role claims a step is unjustified, the claim is checked
against the repository. When the disclosure role reconstructs a fix, the
reconstruction is compared against the gold patch by a separate process that does
have it.

Attacks that succeed do not become bundle patches. They become property test suite cases, and
the instance is sent back for revision. The discrimination lane of the released property test suites is
three times the size of the other two lanes combined, most of it the accumulated
residue of attacks that once worked.

\paragraph{What the audit found.}
\cref{tab:audit} reports the outcomes. All four roles ran on $3{,}002$ bundles.
The strength role reports the specification sound on $2{,}997$; two reports
claim a break, two are inconclusive, and one was skipped. Re-running the attacks
did not reproduce one claimed break and one inconclusive report, and those two
bundles are released; the other three are refused. The
faithfulness role clears all $3{,}002$. The disclosure role returns $18$
findings and $2{,}984$ clean reports; every finding is dispositioned, either
repaired by narrowing the view or recorded as one of the documented relaxations of
\cref{app:leakage}. The fidelity role clears $3{,}000$ and rejects two.

\begin{table}[htbp]
\centering
\caption{Adversarial audit outcomes, pooled over both corpora and all four
backends. Each role produces one report per bundle. ``Inconclusive'' means the
role could neither construct an attack nor certify the bundle within its budget,
which is recorded as a negative outcome for admission purposes.}
\label{tab:audit}
\small
\setlength{\tabcolsep}{5pt}
\begin{tabular}{@{}lp{5.4cm}rrr@{}}
\toprule
\textbf{Role} & \textbf{What it tries to exhibit} & \textbf{Clear}
& \textbf{Finding} & \textbf{Inconcl.} \\
\midrule
Fidelity & A step in the correspondence that no behavior-preserving relation
justifies & $3{,}000$ & $2$ & $0$ \\
Strength & A wrong implementation that verifies against the frozen specification
& $2{,}997$ & $2$ & $2$ \\
Faithfulness & A scenario where the specification and the issue disagree
& $3{,}002$ & $0$ & $0$ \\
Disclosure & A reconstruction of the fix from the solver-facing view
& $2{,}984$ & $18$ & $0$ \\
\bottomrule
\end{tabular}
\end{table}

The table records that a stated attack, run by a specific attacker under a
specific budget, did not break these specifications. Attacks that succeeded
during construction were resolved by sending the instance back for revision rather than by
recording a failure, so the finding rate on released bundles is close to zero by
construction. The audit's calibration is measured on specifications that were
\emph{not} built this way. \Cref{app:reaudit} reports that calibration, and
\cref{app:refutation} reports what the same procedure finds when it is pointed at
solver-written specifications, where it rejects a third to a half of the
specifications written under specification synthesis and close to two thirds of
those written in row~3.

\paragraph{Judges.}
The audit roles were run with a frontier model as the attacking agent
($2{,}741$ of the $3{,}002$ bundles per role) and a stronger successor model on
the remainder ($261$). Where the two overlap, the successor model is the
calibration instrument of \cref{app:reaudit} rather than the recorded verdict.

\subsection{Keeping a recorded outcome honest}
\label{app:reaudit}

An admission record computed once may be stale if the code that produced it has
since changed. Three mechanisms keep the record current.

\paragraph{Every outcome is versioned.}
Each of the $48{,}227$ recorded outcomes carries the identity of the code that
produced it: $39{,}786$ carry the checking code's version and $18{,}840$
additionally carry a digest of the harness configuration in force. The record
spans five weeks of construction, from mid-July to mid-August, during which the
checking code changed many times. $625$ distinct versions appear across the two
corpora, together with $37$ distinct harness configurations. When a check's
definition changes, the outcomes produced under the old definition become
identifiably stale and can be re-derived; $102$ outcomes were re-derived in this
way after a check tightened.

\paragraph{Admission is recomputed, never stored.}
A bundle's admitted status is derived, on every read, from the recorded outcomes
of the thirteen checks by a single procedure, and every consumer of the corpus
uses that one procedure. The counts in this appendix therefore agree with the
counts in \cref{tab:release} and with what a downloaded bundle reports about
itself. The only way a bundle becomes admitted is by acquiring a positive
outcome, and a positive outcome requires the check to have run.

\paragraph{Re-auditing is cheap because the attacks are kept.}
Each bundle's record contains, verbatim, the attacks the audit attempted and the
reason each failed, including the candidate implementations. A later audit that
wants to know whether a bundle is still sound can replay the previous attacks
against the current specification for the cost of running the verifier, and only
has to generate new attacks where the specification has changed since.
Re-auditing the corpus is therefore a routine operation rather than a full
reconstruction.

\paragraph{Calibrating the audit.}
The audit's accuracy cannot be measured on the released corpus, where it almost
never finds anything. The measurement is made instead on the specifications that
solvers write during evaluation, where findings are common enough to compare
judges against each other. On those, a blind re-audit by a stronger model, run
with no access to the original verdicts, reaches a
Fleiss $\kappa$ of $0.861$ among its three independent ballots on the
faithfulness property and $0.871$ on the overall verdict, and its verdicts are
strongly associated with the independent outcome of whether the solver's patch
resolved the task (a Mantel--Haenszel odds ratio of $13.5$, $95\%$ CI
$[7.96, 22.93]$). \Cref{app:refutation} reports that analysis in full. The same
procedure, applied to specifications built without the pipeline above, rejects
between two fifths and two thirds of them, depending on the setting. A
near-zero rate on the released corpus is therefore not explained by an audit
that never rejects.

\section{Evaluation protocol}
\label{app:protocol}

The evaluation is built around a single discipline: between any two settings we
compare, exactly one thing changes. The model, the agent scaffold, the
container, the step budget, the way the issue text is presented, and the
sanitization of the working tree are identical. What differs is either the
formal information the agent is handed or the rule by which its submission is
graded, and never both at once. Every claim in \cref{sec:experiments} is a
difference between two settings that satisfy that condition.

This appendix records the protocol in four parts. \Cref{app:modes} states what
each of the nine settings gives the agent and what it demands in return.
\Cref{app:harness} describes the environment that is held fixed across all of
them, and reports what an episode actually looks like from the inside.
\Cref{app:scoring} describes the graders, three mechanical and three agentic,
and the order in which they run. \Cref{app:settings} records the models,
budgets, and repetitions behind every number we report.

\subsection{The nine settings}
\label{app:modes}

A setting is fixed by three choices: what the agent is told, what it must hand
back, and what has to be true of what it hands back. \Cref{tab:settings-detail}
states all three for each of the nine rows of \cref{tab:main}.

\begin{table}[t]
\centering
\caption{The nine evaluation settings. ``Local.'' is edit localization: the
functions and files the fix touches. ``Spec.'' is the specification view of
\cref{app:structure}: the specification and its axioms with the reference
implementation and all proof annotations removed. ``Verifier'' means the agent
can invoke the backend's verifier on its own artifact during the episode. Row
numbers match \cref{tab:main}. The table states the protocol for the three
prover-backed backends; the structured-requirements backend deviates as
described at the end of this subsection.}
\label{tab:settings-detail}
\small
\setlength{\tabcolsep}{3pt}
\begin{tabular}{@{}clcccL{2.85cm}L{3.3cm}@{}}
\toprule
& & \multicolumn{3}{c}{\hd{Given to the agent}} & & \\
\cmidrule(lr){3-5}
\hd{\#} & \hd{Setting} & \hd{Local.} & \hd{Spec.} & \hd{Verifier}
  & \hd{Must submit} & \hd{Counts as solved iff} \\
\midrule
0 & Unaided baseline & \na & \na & \na
  & a patch
  & the patch resolves \\
\addlinespace[2pt]
1 & Baseline $+$ Audit & \na & \na & \na
  & a patch
  & the patch resolves and the coun\-ter\-ex\-am\-ple audit finds no real
    violation \\
\addlinespace[2pt]
2 & End-to-End & \na & \na & \checkmark
  & a patch; a specification and a verifying implementation are requested
  & the patch resolves \\
\addlinespace[2pt]
3 & Verified End-to-End & \na & \na & \checkmark
  & a specification, a verifying implementation, and a patch
  & the implementation verifies, the patch resolves, and the two are judged
    equivalent \\
\addlinespace[2pt]
4 & Localization Provided & \checkmark & \na & \na
  & a patch
  & the patch resolves \\
\addlinespace[2pt]
5 & Verified from Localization & \checkmark & \na & \checkmark
  & a specification, a verifying implementation, and a patch
  & the implementation verifies, the patch resolves, and the two are judged
    equivalent \\
\addlinespace[2pt]
6 & Specification Provided & \checkmark & \checkmark & \checkmark
  & a patch
  & the patch resolves \\
\addlinespace[2pt]
7 & Verified from Specification & \checkmark & \checkmark & \checkmark
  & a verifying implementation and a patch
  & the implementation verifies, the patch resolves, and the two are judged
    equivalent \\
\midrule
8 & Specification Synthesis & \na & \na & \checkmark
  & a specification and a witness implementation, and no patch
  & the specification passes all five properties of the audit \\
\bottomrule
\end{tabular}
\end{table}

\paragraph{The three axes.}
The first axis is \emph{localization}. Rows 4--7 name the functions and files
the fix touches; rows 0--3 and row 8 do not. Localization is a hint
about the repository: it names sites, never behavior. It is
computed once, offline, from the frozen corpus, so the evaluation process never
opens a gold patch while an episode is running.

The second axis is the \emph{specification}. Rows 6 and 7 hand the agent the
specification view: the formal contract, the axioms that summarize the callees
it depends on, and the operation correspondence, with the reference
implementation and every proof annotation removed. Rows 2, 3, 5, and 8 ask the
agent to write that artifact itself. This is the axis the paper is about. The
pair (row 7, row 3) isolates it exactly: identical scaffold, identical grading
rule, and the only difference is whether the specification arrives with the task
or has to be constructed.

The third axis is \emph{what is graded}. Five settings are graded on the patch
alone, by the official held-out test suite. Three are graded on the formal
artifact as well: the submitted implementation must verify, and an adversarial
panel must confirm that it and the patch describe the same behavior. Row 8 is
graded on the specification alone and asks for no patch at all.

\paragraph{The designed contrasts.}
Six differences between rows carry the analysis, and each is a comparison in which
everything but one axis is fixed.

\begin{itemize}[leftmargin=1.4em,itemsep=2pt,topsep=3pt]
  \item \textbf{Row 4 $-$ row 0} measures localization alone. Both settings are
    graded by the held-out tests and neither mentions formal machinery, so the
    difference is the value of knowing where to edit.
  \item \textbf{Row 6 $-$ row 4} measures the specification and its verifier
    \emph{on top of} localization. Row 6 gives everything row 4 gives, plus the
    specification view and the verifier, and is graded identically. This is the
    contrast that separates the benefit of a correct specification from the
    benefit of the localization it necessarily reveals.
  \item \textbf{Row 2 $-$ row 0} measures the effect of asking an agent to
    verify while changing nothing about its inputs. Row 2 has the verifier and
    the instruction to use it; the issue text, the repository, the budget, and
    the grading rule are those of row 0.
  \item \textbf{Row 7 $-$ row 3} measures a supplied specification against a
    constructed one under a single grading rule. Both rows require a verifying
    implementation, a resolving patch, and a confirmed equivalence; only the
    origin of the specification differs.
  \item \textbf{Row 3 $-$ row 2} and \textbf{row 7 $-$ row 6} measure the price
    of the stricter grade. In each pair the agent's environment is identical and
    only the pass predicate tightens, from ``the patch resolves'' to ``the
    patch resolves and the formal artifact holds up.'' A large drop in either
    pair would mean that resolution rates in the loose settings are being
    carried by submissions whose formal half does not survive inspection.
  \item \textbf{Row 1 $-$ row 0} measures the price of an independent
    counterexample search on a submission that the held-out tests already
    accepted. It is the control for the claim that a test suite is an incomplete
    correctness signal; the two rows differ in the grader and in nothing else.
\end{itemize}

Rows 5 and 8 stand slightly apart. Row 5 completes the grid: it is row 3 with
localization added, or equivalently row 7 with the specification withheld. It
therefore separates the two halves of what row 7 supplies. Row 8 is not a
patch-writing task. The other eight settings measure specification quality only
through its downstream effect on a patch; row 8 measures it directly.

\paragraph{Why the settings come in pairs.}
Eight of the nine rows form four loose/strict pairs: (0,~1), (2,~3), (4,~5),
and (6,~7). The pairing is deliberate. An agent's behavior
depends on what it is asked to produce, and a grading rule that is announced in
the prompt is part of what it is asked to produce; a grading rule applied after
the fact is not. Row 3 differs from row 2 in both respects, while row 1 differs
from row 0 only in the grader, so the (2,~3) pair alone does not separate the
effect of the information from the effect of the rule. Reading the four pairs together does: rows 0, 2, 4, and 6 vary the
information under a fixed rule, and each strict row is anchored to the loose row
it was built from.

\paragraph{What the specification view withholds.}
The view given in rows 6 and 7 is not the corpus artifact. The reference
implementation is removed, every proof annotation and lemma is removed, and the
view is screened for text that would name the fix rather than describe the
required behavior; \cref{app:leakage} describes the screen and the cases where a
relaxation was recorded instead. What remains is a formal statement of what the
changed functions must do, the axioms that pin down the behavior of the callees
they rely on, and the correspondence between the abstract operations and the
repository's own names. An agent in row 6 therefore knows exactly what
correctness means for this issue and still has to write repository code that
achieves it.

\paragraph{The natural-language backend.}
The structured-requirements backend has no prover, so the protocol changes shape
for it. Its verify slot holds a well-formedness check on the requirements
document, not a proof: ``verified'' for one of its episodes means the document is
a well-formed set of requirements with triggered acceptance criteria, never that
anything was proved. Nothing about it can be graded on verification, on
equivalence, or on a specification audit. Every one of its settings is therefore
graded on the patch, and the strict half of each pair is graded by the
adversarial counterexample audit of \cref{app:scoring}. That audit uses the Lean
sibling of the same instance as formal ground truth: the two artifacts describe
the same task, so a formal oracle is available even where the submitted document
cannot itself be checked.

The audit decides rows 3, 5, and 7 for this backend; rows 2 and 6 are decided
by the held-out tests alone. Each pair is agent-side identical (same
prompt, same tools, same budget), so within a pair the entire difference lives in
the grader. In rows 6 and 7 the requirements document handed to the agent is
complete rather than a view with parts elided, since there is no implementation
or proof to remove. There is consequently no formal artifact for the agent to
construct in those two rows. Row 8 does not apply at all: there is no formal
object for a specification audit to check.

\subsection{The agent's environment}
\label{app:harness}

Everything in this subsection is identical across all nine settings and all four
backends, with the single exception of which tool occupies the verify slot.

\paragraph{Scaffold.}
The agent runs a single reason-and-act loop in the style of
\citet{yang2024sweagent}: it emits one tool call at a time, receives the result,
and continues. There is no planner, no retriever, no subagent, no
self-consistency, and no external memory. Three actions are available. A shell
runs commands inside the task's container. A verify action runs the backend's
verifier on the artifact the agent has written, and returns its verdict. A
submit action ends the episode. Settings 0, 1, and 4 expose the shell and submit
only; the verify action is absent.

The scaffold is deliberately plain. A stronger harness could recover a weak
specification by trying more implementations, and a weaker one could fail to
exploit a good specification; a minimal loop keeps the measured difference
attributable to the difference in information.

\paragraph{The container.}
Each episode runs in the task's own official evaluation image, with the working
tree at the base commit, the state of the repository immediately before the fix.
The held-out tests are never present in that image. Before the agent is given a
shell, the version-control history is sanitized: every reference except the base
commit is removed and unreachable objects are pruned, so the future of the branch
is not recoverable from inside the container.

The task description the agent sees is a projection of the corpus record. The
gold patch, the test patch, and the lists of tests that must newly pass and must
keep passing are all dropped. Maintainer hint text, where the upstream dataset
carries it, is excluded: it frequently contains the eventual fix in prose.

\paragraph{The verify action cannot leak.}
The verify action runs on the host, not in the container, and it is given
exactly one input: the artifact the agent wrote. It has no access to the
repository, to the held-out tests, to the gold patch, or to any part of the
corpus bundle for the instance. Its output is the verifier's own verdict on the
agent's own file.

\paragraph{Budget and termination.}
Every episode has a budget of $250$ agent steps in every setting, model, and
backend. An episode ends when the agent submits or when the budget is exhausted,
and both are outcomes that are scored.
An idle turn (a reply with no tool call and no submission)
draws a reminder of the remaining budget rather than being counted as a
submission.

\paragraph{What an episode looks like from the inside.}
\Cref{tab:records} reports the interaction profile of an episode, averaged over
sampled episodes in every completed run. Three patterns stand out.

\begin{table}[t]
\centering
\caption{Interaction profile of an episode, by setting, model, and backend.
Columns are per-episode means: model turns, shell calls, shell calls returning a
nonzero status, verify calls, and verify calls that returned a passing verdict.
``Records'' is the length of the recorded episode in events (mean / median). Rows aggregate over the repetitions of a cell; $s$ is the number of
repetitions. Sampled uniformly at $60$ episodes per repetition.}
\label{tab:records}
\small
\setlength{\tabcolsep}{3pt}
\begin{tabular}{@{}lccccccrrr@{}}
\toprule
& & & & \multicolumn{2}{c}{\hd{Shell}} & \multicolumn{2}{c}{\hd{Verify}}
  & \multicolumn{2}{c}{\hd{Records}} \\
\cmidrule(lr){5-6}\cmidrule(lr){7-8}\cmidrule(lr){9-10}
\hd{Setting} & \hd{Backend} & \hd{$s$} & \hd{Turns} & \hd{calls}
  & \hd{fail} & \hd{calls} & \hd{pass} & \hd{mean} & \hd{med.} \\
\midrule
0 Unaided baseline          & \na & 4 & 12.2 & 11.2 & 0.25 & \na & \na & 26.4 & 17.8 \\
4 Localization Provided     & \na & 4 & 10.2 & 9.2 & 0.27 & \na & \na & 22.4 & 12.8 \\
\midrule
2 End-to-End                & \nagini & 4 & 22.6 & 18.9 & 0.54 & 2.74 & 1.23 & 47.2 & 35.9 \\
3 Verified End-to-End       & \nagini & 4 & 22.9 & 18.8 & 0.58 & 3.21 & 1.28 & 47.9 & 36.0 \\
6 Specification Provided    & \nagini & 4 & 19.2 & 15.1 & 0.35 & 3.27 & 1.23 & 40.6 & 27.9 \\
7 Verified from Specification & \nagini & 4 & 19.5 & 15.0 & 0.33 & 3.61 & 1.27 & 41.2 & 27.2 \\
8 Specification Synthesis   & \nagini & 4 & 18.1 & 12.0 & 0.29 & 5.11 & 1.63 & 38.1 & 26.5 \\
\midrule
\multirow{3}{*}{5 Verified from Localization}
  & \nagini & 1 & 21.9 & 17.6 & 0.57 & 3.42 & 1.32 & 46.0 & 36.0 \\
  & \velvet & 1 & 38.5 & 29.2 & 0.80 & 8.55 & 1.52 & 80.2 & 76.0 \\
  & \lean   & 1 & 20.0 & 17.2 & 0.95 & 2.05 & 1.22 & 43.0 & 33.0 \\
\midrule
2 End-to-End                & \ears & 1 & 16.7 & 14.8 & 0.62 & 1.05 & 1.02 & 35.5 & 28.0 \\
3 Verified End-to-End       & \ears & 1 & 17.9 & 16.1 & 0.50 & 1.03 & 1.03 & 38.1 & 27.0 \\
\bottomrule
\end{tabular}
\end{table}

Verify calls scale with how much formal work a setting demands, and they
peak where no patch is required at all. An agent handed a specification
(rows 6 and 7) calls the verifier about $3.3$ times; an agent constructing one
(rows 2 and 3) calls it $2.7$--$4.2$ times; an agent doing nothing but
specification synthesis (row 8) calls it $5.1$--$6.1$ times. The verifier is the
only feedback channel in row 8, so the agent leans on it hardest exactly where
the artifact is all there is.

Third, the ratio between calls and passing verdicts is a direct measure of how
hard a backend is to satisfy. Under \nagini{} an agent needs about $2.6$ calls
per passing verdict. Under \velvet{} it needs $8.55$ calls for $1.52$ passes,
better than five to one. That is the largest such ratio anywhere in the grid, and
it matches the construction-side picture of \cref{tab:verifytime}. \lean{} sits
at the other end, $2.05$ calls to $1.22$ passes, because its failures are
reported as concrete unsolved goals rather than as a search that did not
converge. Backend difficulty is not only a property of the corpus; it is
something an agent pays for turn by turn.

\paragraph{The anti-cheat screen.}
Every recorded episode is screened for the four ways an agent could obtain the
answer rather than derive it: recovering the fix from version-control history,
reading a held-out test, reaching the network, and touching the corpus bundle
for its own instance. The screen is deliberately over-broad. It matches on
surface patterns and is tuned to over-report, so a match is a pointer to be read
rather than a verdict. Only a high-severity match, which means the pattern
plausibly reached protected material, marks an episode unclean; browsing
sanitized history is recorded and left at the lowest severity, because the
history no longer contains the fix.

Across all $53{,}500$ episodes the screen raised $2{,}234$ matches on $711$
episodes: $1{,}292$ informational, $124$ low, and $818$ high. The high matches
fall on $332$ episodes, $0.6\%$ of the total, and carry three labels. Most,
$708$, are searches over the path where bundles live, which is not mounted in the
agent's container and so returned nothing. Ninety are network patterns, and
twenty match an agent writing its own regression test into an existing test file,
which is a legitimate action whose result the grader ignores. No episode in any
reported cell was excluded for cheating, and the counts above are the screen's
raw output.

\subsection{Grading a submission}
\label{app:scoring}

Six graders decide the numbers in \cref{tab:main}: three mechanical and three
agentic. \Cref{tab:graders} states, for each, what a positive outcome
establishes and what it does not.

\begin{table}[t]
\centering
\caption{The graders. ``Mechanical'' means the verdict is a program's output on
fixed inputs and is reproducible bit for bit; ``agentic'' means a model panel
decided it. Every agentic grader runs only on a submission that has already
passed the mechanical graders that apply to its setting.}
\label{tab:graders}
\small
\setlength{\tabcolsep}{3pt}
\begin{tabular}{@{}L{2.15cm}L{1.8cm}L{3.75cm}L{3.85cm}L{1.4cm}@{}}
\toprule
\hd{Grader} & \hd{Decided by} & \hd{A positive outcome establishes}
  & \hd{It does not establish} & \hd{Used in} \\
\midrule
Resolution & mechanical
  & the patch makes the held-out failing tests pass and keeps the passing ones
    passing, under the official harness
  & that the patch is correct; the suite is incomplete by construction
  & 0--7 \\
\addlinespace[2pt]
Verification & mechanical
  & the submitted implementation satisfies the submitted specification, checked
    by the backend's own verifier
  & that the specification says anything useful
  & 3, 5, 7 \\
\addlinespace[2pt]
Anti-fakery & mechanical
  & the submission does not verify by weakening the contract, stubbing the
    proof, or admitting vacuity
  & faithfulness of the specification to the issue
  & 3, 5, 7, 8 \\
\addlinespace[2pt]
Equivalence & agentic, 3 votes
  & the verified implementation and the submitted patch describe the same
    behavior
  & that either one is what the issue asked for, when the specification was
    supplied
  & 3, 5, 7 \\
\addlinespace[2pt]
Specification audit & agentic, 3 votes, tool-using
  & the synthesized specification is sound, admissible, complete, faithful, and
    rests on sound axioms
  & that a patch exists; no patch is requested
  & 8 \\
\addlinespace[2pt]
Counterexample audit & agentic, mechanically re-checked
  & a behavior on which an accepted patch violates the formal ground truth, or
    the absence of one after a directed search
  & correctness, when it finds nothing; absence of evidence
  & 1; \ears{} 3, 5, 7 \\
\bottomrule
\end{tabular}
\end{table}

\paragraph{Resolution.}
The submitted patch is applied to a clean checkout of the repository at the base
commit, inside a fresh instance of the official evaluation image, and the
official harness runs the held-out suite. Nothing the agent did to its own
container survives into this step: the patch is the only thing carried across.
An instance resolves when every test that must newly pass does, and every test
that must keep passing does. A small number of instances have tests that are
nondeterministic; a failure that flips to a pass on a bounded retry is
recorded as a retry, and \cref{app:mechanical} reports the rate.

\paragraph{Verification and anti-fakery.}
In the three strict settings, the agent's implementation is checked against the
agent's specification by the same verifier and under the same hygiene screen
that the corpus admission uses (\cref{app:mechanical}). A submission is therefore
held to exactly the standard the corpus itself was held to. The screen rejects
the obvious degeneracies: a contract weakened until it says nothing, a proof
discharged by an escape hatch, a precondition that no input satisfies. It is
mechanical, so it applies uniformly.

\paragraph{Equivalence.}
Verification and resolution are checks on two different objects: a formal
implementation and a repository patch. Passing both separately is not enough,
because an agent could verify a trivial implementation and, independently, write
a patch that resolves. The equivalence grader closes that gap. Three
independent judges each receive the submitted specification, implementation,
and patch, the issue, the gold patch, and, when one was supplied, the reference
specification, and decide whether the
implementation and the patch describe the same behavior; a majority carries.
When the specification was supplied, the judges additionally check that it is
unchanged from the one given, so the strict grade in row 7 cannot be reached by
relaxing the target. When the specification was constructed, they check that it
is a faithful reading of the issue. That is the same question the row 8 audit
asks, and it is reported the same way.

\Cref{app:refutation} gives the agreement between the three votes, how often the
majority overrides a dissent, how much machine evidence each vote rests on, and
what the panel's verdicts predict about an outcome it never sees.

\paragraph{The specification audit.}
Row 8 asks for a specification and a witness implementation and grades the
specification against five properties: that its axioms hold of the real callees,
that its preconditions admit the inputs the issue is about, that it is true of
the intended behavior, that it constrains enough of that behavior, and that it
is a faithful reading of the issue. Each property is voted on by three judges,
and the specification passes only if a majority accepts all five. The judges are
tool-using: a judge can put a candidate implementation through the verifier and
can probe a claimed axiom against the real callee in the instance's own image, so
a refutation usually comes with a concrete witness. Per-property verdicts are
recorded. The attribution in \cref{tab:synth} rests on them: a specification does
not merely fail, it fails on faithfulness.

\paragraph{The counterexample audit.}
The remaining grader inverts the question. Instead of asking whether a
submission is right, it hunts for a behavior on which the submission is wrong
despite having passed. An adversary is given the formal ground truth for the
instance from the frozen corpus, the issue, the held-out tests, and a checkout
with the agent's patch applied, and is asked to exhibit a concrete input on
which the patch violates that ground truth. Its claim is then re-run
mechanically, and must clear two bars: the exhibited behavior must fail under
the agent's patch, and it must hold under the gold patch. A claim that fails either bar is discarded. We also inspect the generated counterexamples to ensure that their inputs satisfy the formal specifications' preconditions. The audit therefore identifies concrete violations of the reference specifications within their intended input domains. The two bars keep the adversary from demoting a correct patch by inventing a requirement the task never had.

Outcomes fall into five categories: a real violation, a claim that turned out to
test something the task does not require, a claim that did not reproduce, an
inconclusive run, and a clean search. Only a real violation moves a grade, and
it only ever moves it downward. Inconclusive counts as a pass. The audit never
runs on a submission that did not already resolve, so its denominator is the
resolved set.

In row 3 on the structured-requirements backend with Opus~4.8, $135$ of $421$
resolved submissions ($32.1\%$) carry a mechanically confirmed violation of the
formal ground truth, which turns an $84.2\%$ resolution rate into a $57.2\%$ pass
rate. This measures how much a held-out suite lets through. Row 1 applies the
same audit to the unaided baseline and gives a similar rate: $132$ of $423$
resolved submissions ($31.2\%$), which turns $84.6\%$ into $58.2\%$.
This result comes from an earlier evaluation on a previous version of \benchmark{}. The updated evaluation in \cref{tab:main} overturns 134 of 425 test-passing patches (31.5\%), consistent with the 31.2\% reported here.

\paragraph{Order of adjudication.}
The graders run mechanical-first, and no agentic grader can create a pass. In a
strict setting a submission must resolve and verify and clear the hygiene screen
before a judge sees it; a judge can then only decline to confirm equivalence. In
row 1 and the audited requirements rows the adversary runs only on a submission
the tests already accepted, and can only demote it. Row 8 is the one setting
whose grade rests on an agentic verdict alone, since there is no patch to check
mechanically. Its judges are tool-using, its audit is reported per property, and
\cref{app:refutation} calibrates it against independent re-adjudication.

\subsection{Models, repetitions, and run accounting}
\label{app:settings}

\paragraph{Models.}
We evaluate Claude Opus~4.8 \citep{anthropic2025opus}. Decoding is left at the
provider default in every setting; we vary information, not sampling temperature.

\paragraph{Repetitions.}
The main text reports single-repetition numbers. Underneath, each
\nagini{} cell was run four independent times, apart from rows 1 and 5, which have
one repetition each; \cref{app:seeds} reports all four together with the spread.
The other backends were run once per cell.
Repetitions are independent replays of the same cell: the same instances, the
same prompts, the same budget, a different sampling stream. They bound
run-to-run noise, not variation across instances.

\paragraph{Budget.}
Every episode in every cell has the same $250$-step budget, and no cell was
given a wall-clock extension. Episodes ran concurrently, so elapsed time is not
a per-episode quantity; \cref{app:cost} reports the token and step cost of each
setting.

\paragraph{Coverage.}
\Cref{tab:coverage} records the repetitions behind every cell. The design calls
for $26$ cells and all $26$ were run, each repetition over all $500$ instances
of \sbv.

\begin{table}[t]
\centering
\caption{Cell coverage: the number of repetitions of each cell, every one of them
over all $500$ instances. ``N/A'' marks a cell the design does not call for.}
\label{tab:coverage}
\small
\setlength{\tabcolsep}{5pt}
\begin{tabular}{@{}lcccc@{}}
\toprule
\hd{Setting} & \nagini & \velvet & \lean & \ears \\
\midrule
0 Unaided baseline           & 4 & \multicolumn{3}{c}{\emph{backend-independent: one cell}} \\
1 Baseline $+$ Audit         & 1 & \multicolumn{3}{c}{\emph{backend-independent: one cell}} \\
2 End-to-End                 & 4 & 1 & 1 & 1 \\
3 Verified End-to-End        & 4 & 1 & 1 & 1 \\
4 Localization Provided      & 4 & \multicolumn{3}{c}{\emph{backend-independent: one cell}} \\
5 Verified from Localization & 1 & 1 & 1 & 1 \\
6 Specification Provided     & 4 & 1 & 1 & 1 \\
7 Verified from Specification& 4 & 1 & 1 & 1 \\
8 Specification Synthesis    & 4 & 1 & 1 & N/A \\
\bottomrule
\end{tabular}
\end{table}

\paragraph{Accounting rules.}
Every rate in this paper is recomputed from the per-episode records rather than
read from a run summary, and every repetition behind it covers all $500$
instances. Where a cell has four repetitions, a table that pools them says so and
gives the pooled denominator.

\section{Additional results}
\label{app:results}

\Cref{sec:experiments} reports one number per cell: a single repetition, one
criterion, no decomposition. This appendix reports everything behind those
numbers that does not fit in the main table.

The six parts answer six separate questions. \Cref{app:pro-results} reports the
full evaluation grid on the \swepro corpus, the transfer test for every finding
of the main text. \Cref{app:verif-effort} asks what
the verifier actually costs an agent during an episode, and uses the answer to
rank the four backends by how hard they are to satisfy from the inside.
\Cref{app:refutation} asks what happens when a grader is asked to adjudicate a
specification the solver wrote: how the three votes agree, how much machine
evidence each vote rests on, and what the verdicts predict about an outcome the
graders never see. \Cref{app:seeds} repeats every \nagini{} cell four times and
reports the spread. \Cref{app:breakdown} splits the corpus by structural
properties of the reference fix and asks where a supplied specification helps and
where it does not. \Cref{app:cost} reports the token, step, and wall-clock cost
of the whole campaign.

Two conventions hold throughout. Settings are named and numbered as in
\cref{tab:main}; a row number always refers to that table. And a \emph{cell} is a
model, backend, and setting, measured over all $500$ instances of the corpus.
\Cref{tab:coverage} records the repetitions behind each one.

\begin{figure}[htbp]
\centering
\includegraphics[width=0.62\textwidth]{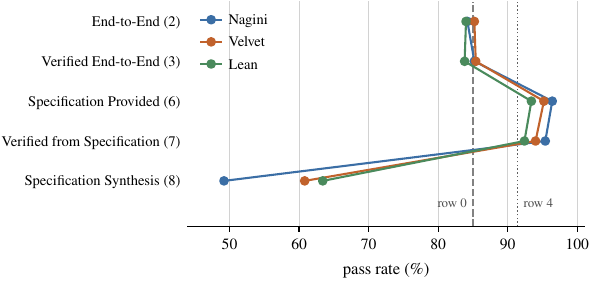}
\caption{Every backend-dependent cell of \cref{tab:main} for Opus~4.8.
Each series is one prover backend across the five settings whose outcome depends
on the backend. The two vertical rules are the settings that use no
specification and are therefore shared by all backends: the unaided baseline
(row 0, dashed) and localization provided (row 4, dotted). The three prover
backends track each other closely everywhere except specification synthesis,
where they separate by $14$ points. The prose backend is omitted here and
reported on its own in \cref{tab:ears}.}
\label{fig:main-results}
\end{figure}

\cref{fig:main-results} draws the main table before any of that decomposition
begins, because two of its features organize everything that follows: a supplied
specification moves every backend above both specification-free settings, and
requiring the agent to produce one moves every backend far below them.

\subsection{Evaluation on the \swepro corpus}
\label{app:pro-results}

\begin{table}[htbp]
\centering
\caption{The full evaluation grid of \cref{tab:main} repeated on the \swepro corpus for Opus~4.8, over the $242$ instances whose bundles are green under all four backends, plus row~8 (specification synthesis), which \cref{tab:main} omits. Columns, settings, and mode numbering are exactly those of \cref{tab:main}, and $\checkmark^*$ and $\checkmark^\dagger$ mean what they mean there. In row~8 the agent writes a specification and a verifying witness but no patch, so nothing is scored on tests and $\checkmark^\ddagger$ marks the adversarial audit of the specification itself. As in \cref{tab:main}, the EARS column has no verifier, so its \emph{Verify} entries are the adversarial counterexample audit, and row~8 does not apply to EARS\@.}
\label{tab:main-pro}
\setlength{\tabcolsep}{2.5pt}
\begin{adjustbox}{max width=\textwidth}
\begin{tabular}{@{}cp{3.8cm}ccccccccc@{}}
\toprule
& & \multicolumn{3}{c}{\textbf{Provided}} & \multicolumn{2}{c}{\textbf{Evaluated}} & \multicolumn{4}{c}{\textbf{Pass rate (\%)}} \\
\cmidrule(lr){3-5}\cmidrule(lr){6-7}\cmidrule(lr){8-11}
\textbf{\#} & \textbf{Setting} & \textbf{Verifier} & \textbf{Local.} & \textbf{Spec.} & \textbf{Tests} & \textbf{Verify} & \textbf{\textsc{Nagini}} & \textbf{\textsc{Velvet}} & \textbf{\textsc{Lean}} & \textbf{EARS} \\
\midrule
0 & Unaided baseline & --- & --- & --- & \checkmark & --- & \multicolumn{4}{c}{\modefill\enspace 61.2 \enspace\modefill} \\
1 & Baseline + Audit & --- & --- & --- & \checkmark & $\checkmark^*$ & \multicolumn{4}{c}{\modefill\enspace 17.4 \enspace\modefill} \\ \midrule
2 & End-to-End & \checkmark & --- & --- & \checkmark & --- & 50.8 & 49.2 & 50.0 & 50.8 \\
3 & Verified End-to-End & \checkmark & --- & --- & \checkmark & \checkmark & 49.6 & 50.0 & 47.9 & 12.4 \\ \midrule
4 & Localization Provided & --- & \checkmark & --- & \checkmark & --- & \multicolumn{4}{c}{\modefill\enspace 61.2 \enspace\modefill} \\
5 & Verified from Localization & \checkmark & \checkmark & --- & \checkmark & \checkmark & 52.9 & 52.1 & 53.7 & 19.4 \\ \midrule
6 & Specification Provided & \checkmark & $\checkmark^\dagger$ & \checkmark & \checkmark & --- & 55.8 & 59.9 & 58.7 & 73.1 \\
7 & Verified from Specification & \checkmark & $\checkmark^\dagger$ & \checkmark & \checkmark & \checkmark & 56.2 & 55.0 & 58.7 & 31.4 \\ \midrule
8 & Specification Synthesis & \checkmark & --- & --- & --- & $\checkmark^\ddagger$ & 31.0 & 35.5 & 49.2 & N/A \\
\bottomrule
\end{tabular}
\end{adjustbox}
\end{table}

\cref{tab:main-pro} repeats the full evaluation grid of \cref{tab:main} on the
\swepro corpus for Opus~4.8, over the $242$ instances whose bundles are green
under all four backends. The absolute level is far from saturation everywhere:
the unaided baseline resolves $61.2\%$. The value of formal information also
redistributes on these larger tasks. A supplied specification no longer adds to
localization: the model scores \emph{below} its localization-provided setting
when handed the specification ($55.8$--$59.9\%$ against $61.2\%$), and prompting
an agent to construct its own specification end-to-end costs roughly ten points
against the baseline. On tasks that change a mean of nine functions across three
files, a specification of the behavioral core evidently concentrates the agent's
attention on a slice of the required edit in a way that \sbv's single-site
patches never exposed. Two of the main text's findings do transfer intact.
First, the jointly scored settings track their test-scored twins closely under
every prover backend (within $2$ points on most cells, $5$ at worst), so
carrying a specification through to a verified, equivalent patch remains
essentially free once the patch resolves. Second, specification synthesis
remains the bottleneck: the five-property audit accepts $31.0$--$49.2\%$ of
Opus~4.8's specifications. Finally, the audited settings sharpen the
incompleteness finding of the main text: of the $61.2\%$ of instances the
baseline agent resolves, the adversarial counterexample audit confirms only
$17.4\%$ against the formal ground truth, a far larger demotion than on \sbv,
consistent with the hidden tests of these larger tasks exercising a smaller
fraction of the changed behavior.

\subsection{Verification effort and backend difficulty}
\label{app:verif-effort}

Every setting that hands an agent a verifier is a setting in which the agent can
choose how much to use it. This subsection reports what agents chose, and what
that reveals about the four backends.

\paragraph{Engagement is universal.}
In every cell where a verify tool exists, essentially every episode calls it:
the engagement rate is $1.000$ in every cell whose artifact is graded. This is a property of the protocol more
than of the agents: rows 3, 5, 7, and 8 refuse a submission that carries no
verifying artifact, so an agent that never calls the verifier cannot finish. But
rows 2 and 6 impose no such requirement, and under every prover backend their
engagement rates are $1.000$ as well. An agent offered a prover uses it whether or
not it is forced to.

\begin{table}[t]
\centering
\caption{Verifier use per episode for Opus~4.8, pooled over the complete repetitions of each cell. ``Calls'' is the number of verify invocations in an episode. Acceptance is reported two ways: \emph{pooled} is accepting verdicts divided by all calls in the cell, and \emph{per ep.}\ is the mean over episodes of that episode's own ratio. ``Verify share'' is the fraction of episode wall-clock spent inside the verifier, again pooled and per-episode. The lower block is row 5, the setting used for the backend comparison in the text.}
\label{tab:vengage}
\small
\setlength{\tabcolsep}{3pt}
\begin{tabular}{@{}lcrrrrrrrr@{}}
\toprule
& & & \multicolumn{3}{c}{\hd{Calls per episode}} & \multicolumn{2}{c}{\hd{Acceptance}} & \multicolumn{2}{c}{\hd{Verify share}} \\
\cmidrule(lr){4-6}\cmidrule(lr){7-8}\cmidrule(lr){9-10}
\hd{Setting} & \hd{Backend} & \hd{$n$} & \hd{mean} & \hd{med.} & \hd{max} & \hd{pooled} & \hd{per ep.} & \hd{pooled} & \hd{per ep.} \\
\midrule
2 End-to-End & \nagini & 2000 & 2.76 & 2 & 33 & 0.468 & 0.727 & 0.201 & 0.150 \\
3 Verified End-to-End & \nagini & 2000 & 2.96 & 2 & 75 & 0.425 & 0.698 & 0.240 & 0.158 \\
6 Specification Provided & \nagini & 2000 & 3.48 & 1 & 88 & 0.408 & 0.733 & 0.252 & 0.178 \\
7 Verified from Spec. & \nagini & 2000 & 3.52 & 1 & 70 & 0.385 & 0.730 & 0.243 & 0.170 \\
8 Specification Synthesis & \nagini & 2000 & 5.16 & 2 & 117 & 0.304 & 0.579 & 0.272 & 0.134 \\
\midrule
2 End-to-End & \ears & 500 & 1.04 & 1 & 3 & 0.990 & 0.995 & 0.000 & 0.000 \\
3 Verified End-to-End & \ears & 500 & 1.05 & 1 & 3 & 0.994 & 0.997 & 0.000 & 0.000 \\
\midrule
\multirow{3}{*}{5 Verified from Local.} & \nagini & 500 & 3.99 & 2 & 57 & 0.352 & 0.600 & 0.407 & 0.244 \\
& \velvet & 500 & 8.92 & 8 & 46 & 0.166 & 0.179 & 0.185 & 0.205 \\
& \lean & 500 & 2.06 & 2 & 9 & 0.597 & 0.753 & 0.225 & 0.248 \\
\bottomrule
\end{tabular}
\end{table}

\begin{figure}[htbp]
\centering
\includegraphics[width=\textwidth]{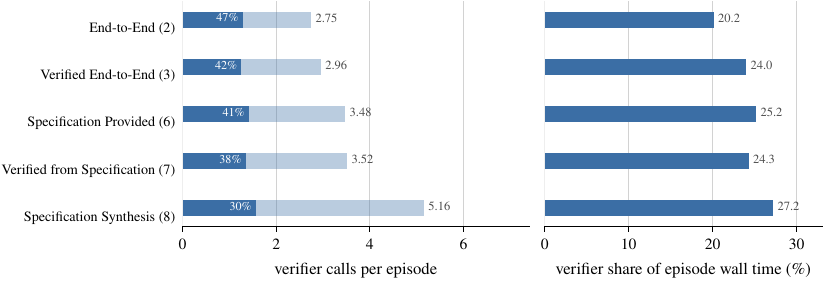}
\caption{How hard an agent leans on the verifier, contract-annotated Python
backend, pooled over the four repetitions of each cell. \textbf{Left:} verifier
calls per episode. The full bar is every call; the dark inner bar is the calls
that returned an accepting verdict, with that share printed inside it.
\textbf{Right:} the share of episode wall time spent inside the verifier. The two
settings that supply no specification make no verifier calls and are omitted.
Specification synthesis is the setting in which the model calls the verifier
most and is accepted least.}
\label{fig:vengage}
\end{figure}

\paragraph{Calls scale with the formal work the setting demands.}
\Cref{tab:vengage} orders the settings the way one would expect if the verifier
were being used as a feedback channel rather than as a formality. An agent handed
a specification (rows 6 and 7) calls the verifier about $3.5$ times. An agent that
must write the specification and the patch (rows 2 and 3) calls it $2.8$ to $3.0$
times. An agent that writes nothing but a specification and a witness (row 8)
calls it $5.2$ times. Row 8 is the setting in which the verifier is the
\emph{only} signal available: there is no patch, so there are no tests to run
against one. It is also the setting in which agents lean on the verifier hardest.

The medians tell a second story that the means hide. Under Opus~4.8 the median
episode in rows 6 and 7 makes exactly one verify call: it writes an
implementation, submits it to the verifier, and is accepted. The mean of $3.5$ is
produced by a minority of episodes that iterate hard, with a maximum of $88$ calls
in a single episode of row 6 and $117$ in row 8.

\paragraph{Acceptance rate is a measure of backend difficulty.}
The fraction of verify calls that return an accepting verdict is the cleanest
available measure of how hard a backend is to satisfy, because it is measured
from inside the agent's own loop rather than from the construction side. Ordered
by that fraction, the four backends separate sharply. \ears{} accepts $0.99$ of calls: its verify slot is a structural check on a requirements
document, not a proof, and a competent agent writes a well-formed document on the
first attempt. \lean{} accepts $0.60$. \nagini{} accepts $0.30$ to
$0.47$, depending on the setting. \velvet{} accepts $0.17$: about six
calls for every accepting verdict, against $1.7$ for \lean{}.

That ordering matches the construction-side effort of \cref{tab:verifytime} and
has the same explanation. A failing \lean{} attempt reports concrete unsolved
goals, so the next attempt is informed. A failing \velvet{} attempt reports that
a search did not converge, which is far less actionable, and the agent's next
move is closer to a guess. Backend difficulty is not only a property of how long
the reference bundles took to build; it is something an agent pays for turn by
turn.

\paragraph{The verifier is a large but not dominant share of an episode.}
Pooled over the campaign, agents spent $896$ hours inside verifiers against
$5{,}937$ hours of total episode wall-clock, or $15\%$. Per cell the share runs
from $12\%$ to $41\%$ under \nagini{} and is $0\%$ under \ears{} by construction
(\cref{fig:vengage}, right).
The pooled share exceeds the per-episode mean share in nearly every cell, by as
much as sixteen points under Opus~4.8 in row 5, which says that verifier time is
concentrated in the same minority of episodes that make many calls.

\begin{table}[t]
\centering
\caption{Distribution of verifier wall-clock per episode for Opus~4.8, in seconds, pooled over the complete repetitions of each \nagini{} cell. The \ears{} cells are omitted: their verify slot returns in well under a tenth of a second and every percentile is $0.0$. The last two columns give the corresponding step-count tail, where $250$ is the budget.}
\label{tab:vtail}
\small
\setlength{\tabcolsep}{4.4pt}
\begin{tabular}{@{}lrrrrrrrrr@{}}
\toprule
& \multicolumn{7}{c}{\hd{Verifier seconds per episode}} & \multicolumn{2}{c}{\hd{Steps}} \\
\cmidrule(lr){2-8}\cmidrule(lr){9-10}
\hd{Setting} & \hd{med.} & \hd{p75} & \hd{p90} & \hd{p95} & \hd{p99} & \hd{max} & \hd{mean} & \hd{p99} & \hd{max} \\
\midrule
2 End-to-End & 34.2 & 62.3 & 139.0 & 333.5 & 932.1 & 4887.2 & 84.9 & 99 & 149 \\
3 Verified End-to-End & 31.2 & 62.2 & 167.2 & 418.6 & 1237.7 & 6849.7 & 100.5 & 92 & 188 \\
6 Specification Provided & 33.1 & 67.4 & 205.7 & 375.3 & 833.2 & 6166.2 & 92.5 & 94 & 250 \\
7 Verified from Spec. & 21.1 & 53.8 & 177.7 & 311.4 & 695.9 & 5650.1 & 76.4 & 91 & 235 \\
8 Specification Synthesis & 37.7 & 106.6 & 426.7 & 949.3 & 4722.9 & 14696.2 & 250.6 & 103 & 250 \\
\bottomrule
\end{tabular}
\end{table}

\paragraph{The tail is long and it lives in specification synthesis.}
\Cref{tab:vtail} makes the concentration explicit. The median episode spends
about half a minute in the verifier in every \nagini{} cell, which is the cost of
one or two accepting calls on a reference-sized problem. The $99$th percentile is
$20$ to $50$ times that. The extreme is row 8 under Opus~4.8: a median of $38$
seconds, a $99$th percentile of $4{,}723$ seconds, and a single episode that spent
$14{,}696$ seconds, just over four hours, inside the prover. Those episodes
are agents that have written a specification they cannot discharge and are
attempting successive witnesses against it, each of which times out.

The step-count tail behaves differently. The $99$th percentile of agent steps
never exceeds $103$ against a budget of $250$, and the budget was reached in
$25$ of the campaign's $53{,}500$ episodes. Agents do not run
out of turns; when they fail they fail while still holding budget, which means the
reported rates measure capability rather than an interaction limit.

\paragraph{Whether the formal artifact comes first.}
In the settings that require an agent to produce both a verifying artifact and a
patch, the order matters: an agent that gets its artifact verifying before it
edits the repository is using the formal object to drive the fix, while an agent
that patches first and formalizes afterwards is using it as a certificate for a
decision already made. We record which happened by comparing, in each episode,
the time of the first accepting verify call against the time of the first shell
command that edits repository source.

\begin{table}[t]
\centering
\caption{Share of Opus~4.8 episodes in which the first accepting verify call precedes the first repository edit. Rows 3 and 7 also require a patch, so the ordering is a genuine strategy choice; row 8 requires no patch at all, and the residual below $100\%$ is agents that write exploratory edits they later discard.}
\label{tab:artifact-first}
\small
\setlength{\tabcolsep}{6pt}
\begin{tabular}{@{}lcrr@{}}
\toprule
\hd{Setting} & \hd{Backend} & \hd{$n$} & \hd{Artifact first} \\
\midrule
3 Verified End-to-End & \nagini & 2000 & 24.1\% \\
7 Verified from Spec. & \nagini & 2000 & 23.9\% \\
8 Specification Synthesis & \nagini & 2000 & 78.3\% \\
\midrule
2 End-to-End & \ears & 500 & 20.4\% \\
3 Verified End-to-End & \ears & 500 & 19.8\% \\
\midrule
5 Verified from Local. & \nagini & 500 & 26.6\% \\
5 Verified from Local. & \velvet & 500 & 1.6\% \\
5 Verified from Local. & \lean & 500 & 4.0\% \\
\bottomrule
\end{tabular}
\end{table}

\Cref{tab:artifact-first} shows that Opus~4.8 patches first in roughly three quarters of episodes.

The backend also matters, and in the direction the acceptance rates predict. Under
\velvet{} only $1.6\%$ of row-5 episodes get a verifying artifact before touching
the repository, and under \lean{} $4.0\%$, against $26.6\%$ under \nagini{} for the
same setting. When the artifact is expensive to discharge, agents
postpone it. Row 8 is the control: with no patch to write, three quarters of episodes verify before making any repository edit at all, and the
residual is exploratory editing that is later abandoned.

\paragraph{Three provers on the same instances.}
Row 5 is the setting we use for the like-for-like prover comparison. All three runs cover the same $500$ instances.
\Cref{tab:threeprovers} reports the comparison.

\begin{table}[t]
\centering
\caption{The three formal backends under row 5 with Opus~4.8, over all $500$
instances. ``Verified'' is the share whose
submitted artifact was accepted by the prover; ``Pass'' additionally requires the
patch to resolve the hidden tests and the equivalence judge to confirm that the
patch and the verified artifact agree. The last two columns repeat the
per-episode verifier statistics of \cref{tab:vengage} for reference.}
\label{tab:threeprovers}
\small
\setlength{\tabcolsep}{6pt}
\begin{tabular}{@{}lrrrrrrr@{}}
\toprule
\hd{Backend} & \hd{$n$} & \hd{Verified} & \hd{Resolved} & \hd{Pass}
  & \hd{Equiv.\ conf.} & \hd{Calls/ep.} & \hd{Accept} \\
\midrule
\nagini & 500 & 100.0\% & 92.4\% & 92.0\% & 99.6\% & 3.99 & 0.352 \\
\velvet & 500 & 100.0\% & 92.2\% & 91.8\% & 99.8\% & 8.92 & 0.166 \\
\lean   & 500 & 100.0\% & 89.8\% & 89.6\% & 99.8\% & 2.06 & 0.597 \\
\bottomrule
\end{tabular}
\end{table}

The outcome rates are within $2.4$ points of each other, and the difference is well
inside the sampling noise of a $500$-instance comparison. What differs by a factor
of four is the effort: $8.92$ verifier calls per episode under \velvet{} against
$2.06$ under \lean{}. Given a localized target and a prover to satisfy, an agent
reaches the same place under all three backends; how much work that takes
depends on the backend.

\subsection{Adjudicating a solver's specification}
\label{app:refutation}

Three of the nine settings ask a grader to decide something about an artifact the
solver wrote: whether a verified implementation agrees with the submitted patch
(rows 3, 5, 7), and whether a synthesized specification is a correct and faithful
statement of the issue (row 8). This subsection reports how those graders behave.
It also reports what happens when the same five-property audit is turned on the
specifications solvers wrote in row 3, which were never audited during scoring.

\paragraph{The equivalence judge rarely refuses.}
The equivalence judge runs only on submissions that have already cleared
verification, resolution, and the anti-fakery check. Its question is narrow:
whether the patch implements the same behavior as the artifact that verified, and
whether a handed-down specification still says what it said. Three independent
ballots decide by majority. Across the eleven adjudicated Opus~4.8 runs the judge
confirmed between $99.2\%$ and $100.0\%$ of the submissions it was given: $24$
rejections out of $4{,}987$ adjudicated submissions, half of one percent.

The grounds of those $24$ rejections correspond to one failure. All cited a
behavioral disagreement: the verified artifact and the submitted patch do not do
the same thing. No Opus~4.8 rejection cited the specification alone.
\Cref{tab:clauses} below measures the same question mechanically over the row-7
submissions.

Dissent is slightly more common than rejection, because a lone dissenting ballot is
overridden. There were $87$ rejecting ballots against $24$ rejections, and in two
runs the panel recorded rejecting ballots but no rejection at all.

No ballot in any run abstained, so every adjudicated submission carries a
confirmation or a refutation.

\begin{table}[t]
\centering
\caption{The equivalence judge for Opus~4.8. ``Judged'' is the number of episodes that reached the judge, out of $500$; the rest failed verification, resolution, or the anti-fakery check and were recorded as non-equivalent without a ballot. ``Conf.''\ is the share of judged episodes the majority accepted. The grounds columns classify each rejection as citing a behavioral disagreement between patch and artifact, an alteration of the given specification, or both. The ballot columns are the individual votes underlying the majorities; no ballot in any run abstained, so an abstention column would be zero throughout and is omitted.}
\label{tab:equiv}
\small
\setlength{\tabcolsep}{4pt}
\begin{tabular}{@{}lrrrrrrrr@{}}
\toprule
& & & & \multicolumn{3}{c}{\hd{Rejection grounds}} & \multicolumn{2}{c}{\hd{Ballots}} \\
\cmidrule(lr){5-7}\cmidrule(lr){8-9}
\hd{Run} & \hd{Judged} & \hd{Conf.} & \hd{Rej.} & \hd{behav.} & \hd{spec.} & \hd{both} & \hd{yes} & \hd{no} \\
\midrule
\multicolumn{9}{@{}l}{\emph{Row 7, Verified from Specification} (\nagini)} \\
1 & 480 & 99.4\% & 3 & 3 & 0 & 0 & 1431 & 9 \\
2 & 483 & 99.2\% & 4 & 4 & 0 & 0 & 1436 & 13 \\
3 & 480 & 99.4\% & 3 & 3 & 0 & 0 & 1430 & 10 \\
4 & 480 & 99.2\% & 4 & 4 & 0 & 0 & 1431 & 9 \\
\midrule
\multicolumn{9}{@{}l}{\emph{Row 3, Verified End-to-End} (\nagini)} \\
1 & 429 & 99.3\% & 3 & 3 & 0 & 0 & 1278 & 9 \\
2 & 424 & 99.8\% & 1 & 1 & 0 & 0 & 1267 & 5 \\
3 & 416 & 99.8\% & 1 & 1 & 0 & 0 & 1245 & 3 \\
4 & 424 & 99.8\% & 1 & 1 & 0 & 0 & 1266 & 6 \\
\midrule
\multicolumn{9}{@{}l}{\emph{Row 5, Verified from Localization} (one repetition per backend)} \\
\nagini & 462 & 99.6\% & 2 & 2 & 0 & 0 & 1377 & 9 \\
\velvet & 460 & 99.8\% & 1 & 1 & 0 & 0 & 1371 & 9 \\
\lean & 449 & 99.8\% & 1 & 1 & 0 & 0 & 1342 & 5 \\
\midrule
\hd{Pooled} & 4987 & 99.5\% & 24 & 24 & 0 & 0 & 14874 & 87 \\
\bottomrule
\end{tabular}
\end{table}

\paragraph{What the five-property panel looks like from the inside.}
The row-8 audit asks five questions of a synthesized specification: that its
axioms hold of the real callees, that it is satisfiable by some implementation,
that it is sound with respect to the reference behavior, that it excludes the
original bug, and that it is a faithful reading of the issue. Each question is
answered by three independent ballots; the specification passes only if a
majority accepts all five.

Two properties of the panel matter for reading its verdicts. First, every ballot
is grounded in machine evidence. A judge has two tools: it can submit a candidate
implementation to the prover against the specification under review, and it can
probe a claimed axiom against the real callee. Across the $5{,}991$ decisive
ballots cast in row 8, all but four issued at least one tool call and received at
least one real machine signal, so almost no verdict rests on the prose alone. A ballot made $4.0$ tool calls on
average, and $98\%$ of ballots saw both an accepting and a rejecting
prover verdict during their deliberation. Recomputing every record's verdict using
only tool-grounded ballots changes nothing, because there are no others.

Second, the panel is mostly unanimous but not degenerately so. Of all row-8
records, $82\%$ were unanimous in one direction or the other and $18\%$
split $2$--$1$. A majority
therefore overrode a dissent on roughly one specification in six. Mean
self-reported confidence was $0.85$ on accepting ballots and $0.86$ on rejecting
ones. The panel is not more confident when it rejects.

\begin{table}[t]
\centering
\caption{Behavior of the two audit panels for Opus~4.8. ``Row 8'' is the panel that scores specification synthesis during evaluation; ``row 3 re-audit'' is the same procedure applied afterwards, by a stronger judge with no access to the original verdicts, to the specifications solvers wrote in row 3. Ballots are counted over all repetitions; a record is one specification that received at least one ballot. ``Grounded'' is the share of decisive ballots that received at least one real machine signal. ``Unanimous'' and ``split'' are shares of \emph{all} records, so they sum to less than one wherever some record carries an abstention.}
\label{tab:panel}
\small
\setlength{\tabcolsep}{4pt}
\begin{tabular}{@{}lrrrrrrr@{}}
\toprule
\hd{Panel} & \hd{Records} & \hd{Ballots} & \hd{Abstain} & \hd{Tools/ballot} & \hd{Grounded} & \hd{Unanim.} & \hd{Split} \\
\midrule
Row 8 & 1997 & 5991 & 0.00\% & 4.07 & 99.9\% & 82.2\% & 17.8\% \\
Row 3 re-audit & 2000 & 6000 & 2.27\% & 7.09 & 93.0\% & 87.3\% & 8.6\% \\
\bottomrule
\end{tabular}
\end{table}

\paragraph{Agreement between the three votes.}
\Cref{tab:kappa} reports inter-judge agreement per property. The headline is that
agreement is high on the property that decides most verdicts and lower on the
properties that almost never fail. Faithfulness reaches a Fleiss $\kappa$ of
$0.77$ in the row-8 panel and $0.83$ to $0.86$ in the re-audit; the overall
verdict reaches $0.76$ and $0.84$ to $0.87$. Completeness, in the row-8
panel, reaches only $0.54$, but its raw pairwise agreement there is
$94\%$, higher than faithfulness's $89\%$. The apparent contradiction
is the standard prevalence effect: when $95\%$ of ratings fall on one side, chance
agreement is already high and $\kappa$ punishes the remainder severely. Both
columns are therefore reported, and no claim here rests on the $\kappa$ of a
property whose prevalence exceeds $0.9$.

\begin{table}[t]
\centering
\caption{Inter-judge agreement across the three independent ballots for Opus~4.8, by property and panel. $\kappa$ is Fleiss' $\kappa$ and $\alpha$ Krippendorff's $\alpha$; ``pairwise'' is the raw share of agreeing judge pairs and ``unanim.''\ the share of items on which all three agree. ``Prev.''\ is the share of individual ratings that voted the property satisfied; read $\kappa$ against it, since agreement coefficients are depressed by extreme prevalence even at very high raw agreement. ``Overall'' is the ballot's accept/reject verdict across all five properties.}
\label{tab:kappa}
\small
\setlength{\tabcolsep}{4pt}
\begin{tabular}{@{}llrrrrrr@{}}
\toprule
\hd{Panel / author} & \hd{Property} & \hd{$\kappa$} & \hd{$\alpha$} & \hd{pairwise} & \hd{unanim.} & \hd{prev.} & \hd{items} \\
\midrule
\multirow{6}{*}{Row 8, Opus 4.8} & Axiom soundness & 0.669 & 0.669 & 97.6\% & 96.3\% & 96.2\% & 1997 \\
& Admissibility & 0.647 & 0.647 & 96.7\% & 95.0\% & 95.1\% & 1997 \\
& Soundness & 0.724 & 0.724 & 94.2\% & 91.3\% & 88.1\% & 1997 \\
& Completeness & 0.535 & 0.535 & 93.7\% & 90.5\% & 92.6\% & 1997 \\
& Faithfulness & 0.772 & 0.772 & 88.6\% & 82.9\% & 51.7\% & 1997 \\
& \emph{Overall} & 0.764 & 0.764 & 88.2\% & 82.3\% & 49.3\% & 1997 \\
\midrule
\multirow{6}{*}{Re-audit, Opus 4.8} & Axiom soundness & 0.830 & 0.831 & 97.6\% & 96.4\% & 92.3\% & 1961 \\
& Admissibility & 0.808 & 0.807 & 98.5\% & 97.7\% & 95.8\% & 1961 \\
& Soundness & 0.824 & 0.824 & 95.8\% & 93.8\% & 86.3\% & 1961 \\
& Completeness & 0.755 & 0.754 & 95.5\% & 93.4\% & 89.9\% & 1961 \\
& Faithfulness & 0.861 & 0.861 & 93.4\% & 90.2\% & 38.4\% & 1961 \\
& \emph{Overall} & 0.871 & 0.871 & 94.0\% & 91.1\% & 36.9\% & 1961 \\
\bottomrule
\end{tabular}
\end{table}

The re-audit agrees with itself substantially better than the row-8 panel does:
$\kappa$ on the overall verdict rises from $0.76$ to $0.87$, and all three
ballots agree on all five properties on $76\%$ of specifications against
$67\%$. It also works harder, making $7.1$ tool calls per ballot against $4.0$,
and it abstains more often, on $2.3\%$ of ballots against under $0.1\%$. It also
has far more ungrounded ballots than the row-8 panel: $394$ of Opus~4.8's
$5{,}864$ decisive ballots reached a verdict without a usable machine signal.
Those ballots are worse ballots. Their mean self-reported confidence is $0.65$
against $0.83$, and they lean towards rejection, four to one. Recomputing every
record's verdict from tool-grounded ballots only leaves $1{,}943$ of $2{,}000$
Opus~4.8 verdicts unchanged, flips $14$, and leaves $28$ with no majority at all.
The re-audit's rejection rate would move by under a point and a half either way.
It is quoted as recorded.

\paragraph{Failures concentrate on faithfulness, and the concentration is
structural.}
\Cref{tab:cooccur} decomposes which properties fail together. Across the pooled
row-8 records, $50\%$ of Opus~4.8's specifications fail at least one property.
Faithfulness is implicated in $95\%$ of those failures, and it is the
\emph{only} failing property in $56\%$ of them. The converse also holds and is
the more informative direction: every single specification that fails
completeness also fails faithfulness, as does $96\%$ of those failing soundness
and three quarters to four fifths of those failing axiom soundness or
admissibility. Only $47$ of Opus~4.8's $1{,}002$ failing specifications fail
something without failing faithfulness.

This is not a redundant panel. A specification can be sound, admissible, complete,
and axiomatically honest and still fail faithfulness. That happens often:
conditional on passing the other four properties, faithfulness still fails on
$36\%$ of Opus~4.8's specifications. What the co-occurrence says is that the
other four properties are close to necessary conditions for faithfulness rather
than independent hurdles: a specification that mis-states what the callees do, or
that admits an implementation the reference would reject, is almost never a
faithful reading of the issue either.

\begin{table}[t]
\centering
\caption{Co-occurrence of failing properties in row 8 for Opus~4.8, pooled over the four repetitions. The upper block gives the marginals; the lower block gives, for each non-faithfulness property, how often a specification that fails it also fails faithfulness. ``Faithfulness only'' is the share of \emph{all} audited specifications whose sole failure is faithfulness.}
\label{tab:cooccur}
\small
\setlength{\tabcolsep}{6pt}
\begin{tabular}{@{}lr@{}}
\toprule
Audited specifications & 1997 \\
Fails at least one property & 1002 (50.2\%) \\
Faithfulness only & 559 (28.0\%) \\
Faithfulness among failing specifications & 95.3\% \\
Fails something but not faithfulness & 47 \\
Faithfulness fails given the other four pass & 36.0\% \\
\midrule
\multicolumn{2}{@{}l}{\emph{Also fails faithfulness, given this property fails}} \\
\quad Axiom soundness & 53/68 (77.9\%) \\
\quad Admissibility & 66/88 (75.0\%) \\
\quad Soundness & 213/223 (95.5\%) \\
\quad Completeness & 126/126 (100.0\%) \\
\bottomrule
\end{tabular}
\end{table}

\paragraph{A blind re-audit of specifications that were never audited.}
Row 3 requires a solver to write a specification, prove an implementation against
it, and submit a patch. Its pass criterion never asks whether the specification
says anything useful. That makes row 3 a natural test of the audit itself: the
specifications were written by agents that had no reason to expect them to be
read, and the audit's verdicts cannot have influenced the outcome, because the
outcome was already recorded. We re-audited all $2{,}000$ of them with a stronger
judge given no access to any earlier verdict.

The re-audit rejects $63.6\%$ of Opus~4.8's row-3 specifications. For comparison,
the row-8 panel rejects $49.2\%$--$52.0\%$ of Opus~4.8's synthesized
specifications. Both figures stand in sharp contrast to the near-zero rate at
which the corresponding construction check finds a defect in the released corpus
(\cref{app:audit}): the same procedure, pointed at specifications built without
the construction pipeline, rejects between a third and two thirds of what it is
shown.

\paragraph{What the audit's verdicts predict about an outcome they never see.}
The re-audit judge sees the specification, the issue, and a prover. It does not
see the patch, the hidden tests, or whether the episode resolved the instance.
\Cref{tab:or} reports how much its verdicts predict about that unseen outcome.

\begin{table}[t]
\centering
\caption{Association between failing the blind re-audit and failing to resolve the instance for Opus~4.8, over the row-3 repetitions. Each repetition is a $2\times2$ table on $493$--$500$ instances; the Mantel--Haenszel estimate combines them without assuming the four are independent samples. The lower block decomposes the association by property.}
\label{tab:or}
\small
\setlength{\tabcolsep}{6pt}
\begin{tabular}{@{}lr@{}}
\toprule
Resolve rate & 85.1\% \\
Specification fails the re-audit & 63.6\% \\
\midrule
$P(\text{unresolved} \mid \text{spec fails})$ & 22.3\% \\
$P(\text{unresolved} \mid \text{spec passes})$ & 2.1\% \\
$P(\text{spec fails} \mid \text{unresolved})$ & 94.9\% \\
$P(\text{spec fails} \mid \text{resolved})$ & 58.1\% \\
\midrule
Odds ratio, Mantel--Haenszel & 13.51 \\
\quad $95\%$ confidence interval & $[7.96, 22.93]$ \\
\quad $p$ & $1.7\times10^{-33}$ \\
Per-repetition odds ratio, range & $8.07$\na$25.16$ \\
\midrule
\multicolumn{2}{@{}l}{\emph{Per-property Mantel--Haenszel odds ratio}} \\
\quad Axiom soundness & 0.93 ($p = 0.85$) \\
\quad Admissibility & 1.53 ($p = 0.17$) \\
\quad Soundness & 5.75 ($p = 2.1\times10^{-37}$) \\
\quad Completeness & 1.38 ($p = 0.13$) \\
\quad Faithfulness & 12.81 ($p = 5.0\times10^{-35}$) \\
\bottomrule
\end{tabular}
\end{table}

The association is large and consistent: an odds ratio of $13.51$, with every
individual repetition between $8$ and $25$. Reading it in the direction the judge
cannot see, $94.9\%$ of the episodes that failed to resolve had written a
specification the audit rejects, against $58.1\%$ of those that resolved. The
per-property decomposition puts almost all of the association on the same two
properties that dominate the failure counts: faithfulness at an odds ratio near
$13$ and soundness near $5.8$. Admissibility and completeness are close to $1$
and not significant. A judge asked whether a specification pins down the right
behavior is answering a question with real consequences downstream.

The association could in principle be produced by a single latent variable,
instance difficulty, rather than by any relationship between specification
quality and outcome. Two observations argue against that reading. First, the
audit is not a proxy for whether the patch worked: among the episodes that
\emph{did} resolve, the audit still rejects $58\%$ of Opus~4.8's specifications.
A grader that merely tracked success would not reject half of the successes.
Second, the failure profile differs between the two groups in kind and not only
in degree. Among resolved episodes, the single most common outcome is a
specification that passes all five properties ($711$ of $1{,}689$); among
unresolved episodes, that outcome occurs $15$ times out of $296$, and the modal
profile is a faithfulness failure with or without a soundness failure alongside
it.

\begin{figure}[htbp]
\centering
\includegraphics{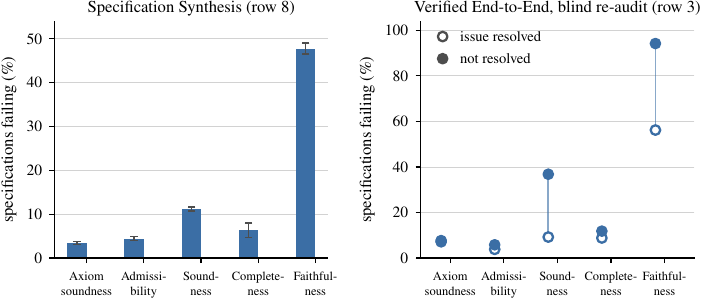}
\caption{Where audited specifications fail, by property. \textbf{Left:} the
row-8 panel, which scores a specification the agent wrote during its own episode.
Bars are the mean rejection rate over the four repetitions and whiskers the
spread across them; the marginals are those of \cref{tab:cooccur}.
\textbf{Right:} the blind re-audit of row-3 specifications
(\cref{tab:reaudit-props}), with each property's rate split by whether the
accompanying patch resolved the instance, an outcome the judge never sees. The
vertical distance inside a pair is the property's discriminative gap: soundness
and faithfulness open wide, admissibility and completeness barely move.}
\label{fig:audit-properties}
\end{figure}

\Cref{tab:reaudit-props} makes the same point property by property, and shows that
the association is not spread evenly across the panel; \cref{fig:audit-properties}
draws it beside the row-8 panel. Two properties separate the
two outcome groups sharply: soundness, which fails on $9\%$ of resolving Opus~4.8
specifications and $37\%$ of non-resolving ones, and faithfulness, which fails on
$56\%$ against $94\%$. Two barely separate them at all: admissibility moves by two
points and completeness by three. Axiom soundness moves by less than a point for
Opus~4.8 and in the wrong direction. The two properties that ask whether the
specification describes the required behavior track whether the episode produced
that behavior, and the two that ask whether the specification is internally
well-formed do not.


\begin{table}[t]
\centering
\caption{Per-property rejection rate of the blind row-3 re-audit for Opus~4.8, split by whether the episode resolved the instance, an outcome the judge never sees. Entries are the mean over the four repetitions with the spread across them; ``gap'' is unresolved minus resolved in percentage points. Recomputing every entry over tool-grounded ballots only, and dropping the instances left without one, moves no entry by more than $1.8$ points and changes no sign.}
\label{tab:reaudit-props}
\small
\setlength{\tabcolsep}{6pt}
\begin{tabular}{@{}lrrrr@{}}
\toprule
\hd{Property} & \hd{Resolved} & \hd{Unresolved} & \hd{Overall} & \hd{Gap} \\
\midrule
\multicolumn{5}{@{}l}{\emph{Specifications written by Opus~4.8 (resolve rate $85.1 \pm 0.9\%$)}} \\
Axiom soundness & $7.6 \pm 1.1$ & $7.1 \pm 1.0$ & $7.5 \pm 1.0$ & $-0.5$ \\
Admissibility & $3.8 \pm 0.8$ & $5.8 \pm 2.9$ & $4.1 \pm 0.8$ & $+2.0$ \\
Soundness & $9.2 \pm 1.0$ & $36.8 \pm 5.6$ & $13.3 \pm 1.6$ & $+27.6$ \\
Completeness & $8.8 \pm 0.8$ & $11.8 \pm 1.5$ & $9.3 \pm 0.7$ & $+3.0$ \\
Faithfulness & $56.2 \pm 1.1$ & $94.2 \pm 2.7$ & $61.9 \pm 1.5$ & $+38.0$ \\
\emph{Any property} & $58.1 \pm 0.9$ & $94.9 \pm 2.1$ & $63.6 \pm 1.3$ & $+36.8$ \\
\bottomrule
\end{tabular}
\end{table}

\paragraph{Specifications with no ballot.}
In row 8, between zero and one specification per repetition was empty: the
agent submitted a verifying witness and no specification, so no ballot could be
cast. Those instances fail the audit and are counted as failures, but flag no
individual property. In the re-audit, the per-property denominators are $493$ to
$500$ and are stated in \cref{tab:or}.

\paragraph{Specification integrity in row 7.}
Rows 6 and 7 hand the agent a specification. Row 7 additionally requires it to
prove an implementation against that specification, which creates an obvious
temptation: weaken the specification until the proof goes through. We measured
what agents actually did to the text they were given.

Verbatim preservation is close to zero. Comparing each submission against the
exact text handed to the solver, only $0$ to $1$ of $500$ submissions per
repetition are byte-identical for Opus~4.8. A byte
comparison counts reflowed docstrings, renamed local variables, reordered
imports, and added explanatory comments as changes, so it says little about what
the submission requires.

Comparing the logical content instead gives the opposite picture.
\Cref{tab:clauses} reports a clause-level comparison over the same submissions:
every precondition, postcondition, invariant, and termination obligation in the
supplied specification, matched against the submission. Between $99.9\%$ and
$100.0\%$ of the given clauses survive. No repetition dropped a single
termination obligation, the clause whose removal would most cheaply make a
proof succeed. Between $0$ and $15$ clauses per $7{,}079$ went missing, and between
$148$ and $184$ contracted helpers were added. In three of the four repetitions,
the number of submissions that changed a clause \emph{and} still verified
\emph{and} still resolved is zero; in the other it is one.

\begin{table}[t]
\centering
\caption{What Opus~4.8 row-7 submissions did to the specification they were handed. ``Verbatim'' is byte-identical to the supplied text. ``Clauses identical'' is the share of submissions whose set of preconditions, postconditions, invariants, and termination obligations matches the supplied set exactly. ``Cosmetic only'' additionally allows reflowing, comments, and renamed locals. ``Retained'' is the share of the $7{,}079$ supplied clauses that survive. ``New helpers'' counts contracted helper functions the agent added, which do not weaken anything supplied. ``Term.\ dropped'' counts supplied termination obligations that went missing, and ``Changed \& passed'' is the number of submissions that altered a supplied clause and nevertheless verified and resolved.}
\label{tab:clauses}
\small
\setlength{\tabcolsep}{5pt}
\begin{tabular}{@{}lrrrrrrr@{}}
\toprule
\hd{Rep.} & \hd{Verb-} & \hd{Clauses} & \hd{Cosmetic} & \hd{Ret-} & \hd{Term.} & \hd{New} & \hd{Changed} \\
& \hd{atim} & \hd{ident.} & \hd{only} & \hd{ained} & \hd{dropped} & \hd{helpers} & \hd{\& passed} \\
\midrule
1 & 0 & 500 & 500 & 100.0\% & 0 & 184 & 0 \\
2 & 0 & 500 & 500 & 100.0\% & 0 & 171 & 0 \\
3 & 1 & 500 & 499 & 100.0\% & 0 & 148 & 0 \\
4 & 0 & 499 & 499 & 99.9\% & 0 & 163 & 1 \\
\bottomrule
\end{tabular}
\end{table}

Two consequences follow. First, row 7 measures what it is meant to measure:
agents carry the supplied specification to a verified implementation rather than
negotiating it down. Second, a verbatim-equality test is the wrong instrument for
this question, and the equivalence judge agrees. Asked separately whether the
specification was preserved, the judge's ballots passed $480$ to $483$ of $500$
submissions per repetition, close to the clause-level result and nowhere near
the verbatim one. A scoring rule built on byte equality would have rejected almost
every correct submission in this setting.

These runs were conducted on an earlier version of \benchmark{}. After these evaluations, we added the conformance gate and re-audited the entire benchmark end-to-end with \benchproofer{}. The comparisons above use the specifications from the earlier version, which were the ones available to the agent during evaluation.

\subsection{Repetitions}
\label{app:seeds}


\Cref{tab:seeds} reports four independent repetitions of the \nagini{} evaluation on an earlier version of \benchmark{}. Following these experiments, we introduced the conformance gate and re-audited the entire benchmark end-to-end with \benchproofer{}. Tables~\ref{tab:main} and~\ref{tab:synth} report results from the updated version, while the repetitions in this section were conducted on the earlier version. Although the numerical results differ slightly, both evaluations support the same conclusions. We retain these earlier repetitions to demonstrate the consistency of our findings across independent runs.

\paragraph{The first repetition is a typical one.}
\Cref{tab:seeds} gives the four rates per cell. The largest gap between the
first repetition and the four-repetition mean anywhere in the grid is $0.9$
points, in row 4. Every other cell is within $0.85$ points. In all seven cells,
the four-repetition mean falls inside the $95\%$ Wilson interval of the first-repetition
rate, and the first-repetition rate falls inside the Student-$t$ interval on the mean.
The repetition-to-repetition standard deviation is at most $1.3$ points in every
cell.

\begin{figure}[htbp]
\centering
\includegraphics{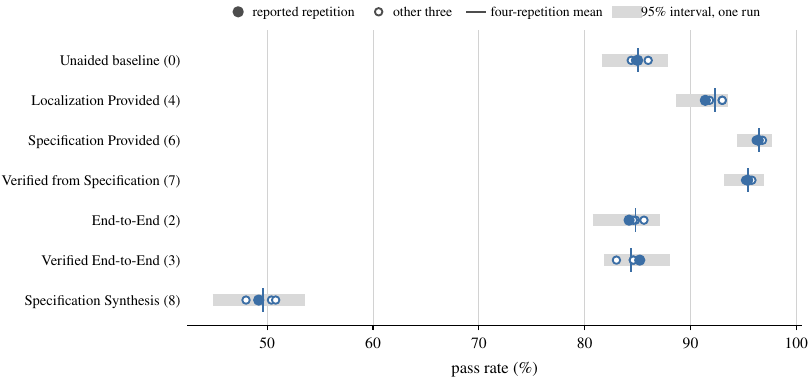}
\caption{The repetitions of \cref{tab:seeds}, one row per setting. The
filled marker is the first repetition on the earlier benchmark version, the three open markers are
the others, and the short rule is their mean. The gray band behind each row is
the $95\%$ Wilson interval of the reported repetition at $n = 500$, the
sampling uncertainty of a single run of that size. In all seven cells all four repetitions fall inside that band: the spread
between repetitions is smaller than the uncertainty any one of them already
carries.}
\label{fig:seed-variance}
\end{figure}

That variability should be read against the sampling uncertainty of a single
$500$-instance run. The $95\%$ Wilson half-width at $n = 500$ runs from $1.66$
points at a rate of $96.4\%$ to $4.37$ points at $49.2\%$. In every cell the
sampling half-width exceeds the repetition standard deviation by a factor of two
to four. Repeating a cell four times buys less precision than the finite corpus
already costs. \Cref{fig:seed-variance} shows the two quantities on one axis.

\begin{table}[t]
\centering
\caption{Four independent repetitions of each \nagini{} cell on \sbv{} for Opus~4.8, $n = 500$ per repetition. The criterion is each setting's own pass criterion, the one the main text reports. ``$r_1$'' is the repetition on the earlier benchmark version and does not correspond to the updated results in \cref{tab:main}. ``$\sigma$'' is the sample standard deviation over the four. ``Any'' is the share of instances solved by at least one repetition and ``all'' by every repetition. ``Wilson'' is the $95\%$ half-width on $r_1$. Cells for the other three backends were run once and are left blank.}
\label{tab:seeds}
\small
\setlength{\tabcolsep}{4pt}
\begin{tabular}{@{}lrrrrrrrrrr@{}}
\toprule
& \multicolumn{4}{c}{\hd{Repetition}} & & & & & & \\
\cmidrule(lr){2-5}
\hd{Setting} & \hd{$r_1$} & \hd{$r_2$} & \hd{$r_3$} & \hd{$r_4$} & \hd{mean} & \hd{$\sigma$} & \hd{range} & \hd{any} & \hd{all} & \hd{Wilson} \\
\midrule
0 Unaided baseline & 85.0 & 86.0 & 84.4 & 84.8 & 85.05 & 0.68 & 1.6 & 89.8 & 79.0 & $\pm3.13$ \\
2 End-to-End & 84.2 & 84.8 & 84.6 & 85.6 & 84.80 & 0.59 & 1.4 & 90.0 & 79.0 & $\pm3.20$ \\
3 Verified End-to-End & 85.2 & 84.6 & 83.0 & 84.6 & 84.35 & 0.94 & 2.2 & 90.6 & 76.6 & $\pm3.11$ \\
4 Localization Provided & 91.4 & 93.0 & 91.8 & 93.0 & 92.30 & 0.82 & 1.6 & 96.2 & 86.4 & $\pm2.47$ \\
6 Specification Provided & 96.4 & 96.4 & 96.8 & 96.2 & 96.45 & 0.25 & 0.6 & 98.6 & 93.0 & $\pm1.66$ \\
7 Verified from Spec. & 95.4 & 95.8 & 95.4 & 95.2 & 95.45 & 0.25 & 0.6 & 98.0 & 91.8 & $\pm1.86$ \\
8 Specification Synthesis & 49.2 & 50.4 & 48.0 & 50.8 & 49.60 & 1.26 & 2.8 & 72.2 & 23.6 & $\pm4.37$ \\
\midrule
\multicolumn{11}{@{}l}{Rows 1 and 5, and all \velvet{}, \lean{}, and \ears{} cells: one repetition, no spread available.} \\
\bottomrule
\end{tabular}
\end{table}

\paragraph{Stability is a property of the instance, not of the rate.}
The aggregate rates are stable to within a point; individual instances are not.
\Cref{tab:instability} classifies each instance by how many of the four
repetitions solved it. In the settings where a specification is supplied, $4\%$ to
$6\%$ of instances are unstable: solved by one, two, or three repetitions but
not all four. In the baseline and end-to-end settings the figure is $9\%$ to
$14\%$. In row 8 it is $38\%$ to $49\%$.

That last number reframes what row 8 measures. Opus~4.8 passes the audit on
$49.6\%$ of instances on average, but that average decomposes into $24\%$ of
instances it passes every time, $28\%$ it fails every time, and $49\%$ where the
outcome depends on the sample. Writing a faithful specification is not a
capability the model either has or lacks per instance; on half the corpus it is a
coin whose bias sits somewhere in the middle. The gap between the union over four
repetitions ($72.2\%$) and the intersection ($23.6\%$) is $48.6$ points, five
times the corresponding gap in row 6.

\begin{table}[t]
\centering
\caption{Instance-level stability over four Opus~4.8 repetitions, criterion = each setting's own pass criterion. $k$ is the number of repetitions that succeeded on the instance; ``unstable'' is $1 \le k \le 3$. $H$ is the Shannon entropy in bits of the $k$-histogram, a single summary of how far the cell is from deterministic. ``Any'' and ``all'' are counts out of $500$.}
\label{tab:instability}
\small
\setlength{\tabcolsep}{4pt}
\begin{tabular}{@{}lrrrrrrrrrr@{}}
\toprule
\hd{Setting} & \hd{$k{=}4$} & \hd{$k{=}3$} & \hd{$k{=}2$} & \hd{$k{=}1$} & \hd{$k{=}0$} & \hd{unstable} & \hd{\%} & \hd{$H$} & \hd{any} & \hd{all} \\
\midrule
0 Unaided baseline & 395 & 27 & 13 & 14 & 51 & 54 & 10.8 & 1.11 & 449 & 395 \\
2 End-to-End & 395 & 20 & 21 & 14 & 50 & 55 & 11.0 & 1.12 & 450 & 395 \\
3 Verified End-to-End & 383 & 32 & 21 & 17 & 47 & 70 & 14.0 & 1.23 & 453 & 383 \\
4 Localization Provided & 432 & 28 & 13 & 8 & 19 & 49 & 9.8 & 0.83 & 481 & 432 \\
6 Specification Provided & 465 & 16 & 9 & 3 & 7 & 28 & 5.6 & 0.49 & 493 & 465 \\
7 Verified from Spec. & 459 & 16 & 10 & 5 & 10 & 31 & 6.2 & 0.56 & 490 & 459 \\
8 Specification Synthesis & 118 & 99 & 79 & 65 & 139 & 243 & 48.6 & 2.27 & 361 & 118 \\
\bottomrule
\end{tabular}
\end{table}

\paragraph{The qualitative findings of the main text hold in every earlier repetition.}
\Cref{tab:contrasts} repeats the four contrasts \cref{sec:experiments} draws,
once per repetition, with a paired test on the four differences and an exact
McNemar test on the instance-level pairing of the first repetition. All three
positive contrasts have the same sign in all four repetitions and a
repetition-level standard deviation under one point. The null contrast, adding
a verifier and a specification obligation to a bare end-to-end setting without
supplying anything, has an inconsistent sign across repetitions and a
paired $p$ of $0.62$. It is indistinguishable from zero.

\begin{table}[t]
\centering
\caption{The contrasts on the earlier benchmark version for Opus~4.8, per repetition. Each row is a pair of settings compared on the same instances. ``$r_1$'' is the difference in the first repetition in percentage points; ``mean $\pm\sigma$'' is over the four paired repetition-level differences; ``paired $p$'' is the two-sided $p$ of a $t$-test on those four differences ($df = 3$); ``McNemar $p$'' is the exact two-sided test on the first repetition's instance pairing ($n = 500$), with $b/c$ the discordant counts.}
\label{tab:contrasts}
\small
\setlength{\tabcolsep}{5pt}
\begin{tabular}{@{}lrrrrrl@{}}
\toprule
\hd{Contrast} & \hd{$r_1$} & \hd{mean} & \hd{$\sigma$} & \hd{paired $p$} & \hd{McNemar $p$} & \hd{$b/c$} \\
\midrule
Row 6 $-$ row 0 & $+11.4$ & $+11.40$ & 0.82 & $1.0\times10^{-4}$ & $4.5\times10^{-13}$ & 63/6 \\
Row 6 $-$ row 4 & $+5.0$ & $+4.15$ & 0.98 & $3.5\times10^{-3}$ & $2.2\times10^{-5}$ & 30/5 \\
Row 6 $-$ row 2 & $+12.2$ & $+11.65$ & 0.75 & $7.5\times10^{-5}$ & $1.2\times10^{-14}$ & 66/5 \\
Row 2 $-$ row 0 & $-0.8$ & $-0.25$ & 0.91 & $0.62$ & $0.63$ & 17/21 \\
\bottomrule
\end{tabular}
\end{table}

\paragraph{An integrity check across the repetitions.}
The two criteria a setting can be scored on are nested: a pass under a verified
setting requires the patch to resolve the instance, so the set of passing
instances must be a subset of the set of resolving ones. Across all $28$ complete
\nagini{} repetitions there is no instance that passes without resolving. In rows
6, 2, 4, and 0 the two sets are identical by construction. In rows 7 and 3 the
difference is the verification and equivalence loss: one to four instances per
repetition resolve the hidden tests but do not clear verification and
equivalence, and are therefore not counted. Row 8 does not grade the patch at
all, so its resolution figures are informational only.

\subsection{Outcomes by instance property}
\label{app:breakdown}

In the earlier evaluation analyzed below, supplying a specification raises the resolution rate by 11.4 percentage points. The updated evaluation in \cref{tab:main} reports a similar improvement of 11.2 percentage points. That aggregate hides a factor of five. This
subsection asks where the gain comes from, using structural properties of the
reference fix as strata. All figures use the four repetitions of the \nagini{}
cells on \sbv.

\begin{table}[t]
\centering
\caption{Structural properties of the $500$ \sbv instances, as
measured on the reference fix and the reference bundle. ``Modeled functions'' is
the number of functions the specification constrains; ``axioms'' is the number of
assumptions the bundle states about callees it does not model.}
\label{tab:instprops}
\small
\setlength{\tabcolsep}{6pt}
\begin{tabular}{@{}lrrrrr@{}}
\toprule
\hd{Property} & \hd{mean} & \hd{p25} & \hd{median} & \hd{p75} & \hd{max} \\
\midrule
Added lines in the reference fix       & 9.94 & 2 & 4 & 10 & 202 \\
Deleted lines                          & 4.39 & 1 & 2 & 4 & 89 \\
Added $+$ deleted lines                & 14.33 & 3 & 7 & 13 & 232 \\
Files touched                          & 1.25 & 1 & 1 & 1 & 21 \\
Hunks                                  & 2.44 & 1 & 1 & 2 & 45 \\
Modeled functions                      & 1.41 & 1 & 1 & 2 & 7 \\
Axioms                                 & 0.53 & 0 & 0 & 1 & 8 \\
Bundles carrying at least one axiom    & 27\% & \na & \na & \na & \na \\
Fail-to-pass tests                     & 3.03 & 1 & 1 & 2 & 438 \\
Pass-to-pass tests                     & 120.28 & 19 & 50.5 & 111 & 2476 \\
Reference verification time (s)        & 25.84 & 13.9 & 18.4 & 29.5 & 288.6 \\
Issue text length (characters)         & 1700 & 644 & 1185 & 2036 & 24770 \\
\bottomrule
\end{tabular}
\end{table}

\paragraph{What the corpus looks like.}
\Cref{tab:instprops} records the distributions the strata are cut from. Three
features shape everything that follows. The reference fixes are small: the median
adds four lines and touches one file. The specifications are local: the median
bundle constrains a single function and states no axiom at all. And two of the
distributions have extreme tails: one instance carries $438$ fail-to-pass tests
against a median of $1$, another touches $21$ files, another adds $202$ lines. Any
linear statistic on those columns is dominated by a handful of instances, so the
rank-based figures below are the ones to read.

\begin{table}[t]
\centering
\caption{Resolution rate by stratum for Opus~4.8, four-repetition means. ``$0$'' is the unaided baseline, ``$4$'' localization provided, ``$6$'' specification provided, ``$2$'' end-to-end, ``$8$'' specification synthesis (a pass rate on the audit, not a resolution rate). ``Gain'' is row 6 minus row 0. Strata with $n < 20$ are shown but excluded from the ranking discussion; three strata below $n = 5$ are omitted, so a family's counts need not sum to $500$.}
\label{tab:strata}
\footnotesize
\setlength{\tabcolsep}{3.5pt}
\begin{tabular}{@{}lrrrrrrr@{}}
\toprule
\hd{Stratum} & \hd{$n$} & \hd{0} & \hd{4} & \hd{6} & \hd{2} & \hd{8} & \hd{Gain} \\
\midrule
All instances & 500 & 85.0 & 92.3 & 96.4 & 84.8 & 49.6 & $+11.4$ \\
\midrule
\multicolumn{8}{@{}l}{\emph{Repository}} \\
\quad django & 231 & 86.7 & 93.6 & 97.2 & 86.0 & 50.5 & $+10.5$ \\
\quad sympy & 75 & 80.3 & 89.3 & 97.0 & 82.0 & 44.3 & $+16.7$ \\
\quad sphinx & 44 & 86.4 & 93.8 & 94.9 & 84.7 & 43.8 & $+8.5$ \\
\quad matplotlib & 34 & 80.1 & 91.2 & 100.0 & 77.9 & 55.9 & $+19.9$ \\
\quad scikit-learn & 32 & 96.1 & 96.9 & 99.2 & 96.1 & 51.6 & $+3.1$ \\
\quad astropy & 22 & 73.9 & 85.2 & 93.2 & 73.9 & 35.2 & $+19.3$ \\
\quad xarray & 22 & 86.4 & 92.0 & 93.2 & 88.6 & 60.2 & $+6.8$ \\
\quad pytest & 19 & 92.1 & 93.4 & 94.7 & 94.7 & 68.4 & $+2.6$ \\
\quad pylint & 10 & 57.5 & 75.0 & 75.0 & 55.0 & 30.0 & $+17.5$ \\
\quad requests & 8 & 90.6 & 100.0 & 100.0 & 93.8 & 50.0 & $+9.4$ \\
\midrule
\multicolumn{8}{@{}l}{\emph{Added lines in the reference fix}} \\
\quad $\le 2$ & 167 & 89.4 & 94.2 & 98.8 & 90.4 & 54.0 & $+9.4$ \\
\quad $2$\na$4$ & 84 & 87.2 & 94.6 & 98.2 & 87.5 & 56.8 & $+11.0$ \\
\quad $4$\na$10$ & 132 & 88.4 & 93.6 & 98.1 & 86.7 & 50.2 & $+9.7$ \\
\quad $> 10$ & 117 & 73.5 & 86.5 & 90.0 & 72.6 & 37.4 & $+16.5$ \\
\midrule
\multicolumn{8}{@{}l}{\emph{Files touched}} \\
\quad $1$ file & 429 & 88.1 & 93.3 & 98.1 & 88.1 & 52.5 & $+10.0$ \\
\quad $> 1$ file & 71 & 66.5 & 86.3 & 86.3 & 65.1 & 32.0 & $+19.8$ \\
\midrule
\multicolumn{8}{@{}l}{\emph{Modeled functions}} \\
\quad single-site & 355 & 89.7 & 93.2 & 97.6 & 89.9 & 52.8 & $+7.9$ \\
\quad multi-site & 145 & 73.6 & 90.0 & 93.6 & 72.4 & 41.7 & $+20.0$ \\
\midrule
\multicolumn{8}{@{}l}{\emph{Axioms}} \\
\quad none & 363 & 83.7 & 91.4 & 96.1 & 82.9 & 49.7 & $+12.4$ \\
\quad $\ge 1$ & 137 & 88.7 & 94.7 & 97.4 & 90.0 & 49.5 & $+8.7$ \\
\midrule
\multicolumn{8}{@{}l}{\emph{Fail-to-pass tests}} \\
\quad $1$ & 345 & 88.9 & 94.2 & 98.0 & 89.3 & 54.0 & $+9.1$ \\
\quad $2$ & 95 & 78.9 & 90.5 & 95.3 & 78.9 & 42.1 & $+16.4$ \\
\quad $3$\na$5$ & 35 & 70.7 & 83.6 & 94.3 & 66.4 & 32.9 & $+23.6$ \\
\quad $\ge 6$ & 25 & 75.0 & 85.0 & 82.0 & 71.0 & 41.0 & $+7.0$ \\
\midrule
\multicolumn{8}{@{}l}{\emph{Reference verification time}} \\
\quad $\le 13.9$s & 131 & 86.1 & 91.4 & 95.6 & 86.5 & 54.0 & $+9.5$ \\
\quad $13.9$\na$18.4$s & 119 & 86.1 & 91.2 & 94.1 & 85.1 & 48.9 & $+8.0$ \\
\quad $18.4$\na$29.5$s & 125 & 84.0 & 93.4 & 98.2 & 81.8 & 49.0 & $+14.2$ \\
\quad $> 29.5$s & 125 & 84.0 & 93.2 & 97.8 & 85.8 & 46.2 & $+13.8$ \\
\midrule
\multicolumn{8}{@{}l}{\emph{Estimated time to fix}} \\
\quad $< 15$ min & 194 & 93.2 & 95.9 & 99.0 & 92.0 & 59.0 & $+5.8$ \\
\quad $15$ min\na$1$ h & 261 & 83.2 & 92.0 & 97.3 & 83.4 & 45.5 & $+14.1$ \\
\quad $1$\na$4$ h & 42 & 61.3 & 79.8 & 82.7 & 63.1 & 34.5 & $+21.4$ \\
\bottomrule
\end{tabular}
\end{table}

\paragraph{The specification helps most where the task is least local.}
\Cref{tab:strata} gives every stratum. Ranked by the gain from supplying a
specification, and restricted to strata with at least $20$ instances, the top of
the list is: instances with three to five fail-to-pass tests ($+23.6$ points), the
one-to-four-hour difficulty band ($+21.4$), specifications constraining more than
one function ($+20.0$), fixes touching more than one file ($+19.8$), astropy
($+19.3$), and sympy ($+16.7$). The bottom is scikit-learn ($+3.1$) and xarray
($+6.8$).

Every entry at the top of that list is a form of non-locality. A fix spread over
several files, a specification that has to constrain several functions, and a bug
whose symptom shows up in several tests are all cases where the hard part is
working out what the code has to do rather than writing it. A specification
states it, and the strata where it helps most are the strata where the
statement is hardest to reconstruct from the issue text. The low-gain repositories
are the opposite case: scikit-learn's instances in this corpus are short,
single-site, and already resolved $96\%$ of the time without help, leaving little
room.

The gain also generally grows with two graded properties: with how long the
reference bundle takes to verify ($+9.5$ to $+13.8$), and with the estimated time
to fix ($+5.8$, $+14.1$, $+21.4$). It is largest in the top quartile of added lines ($+16.5$ against $+9.4$ to
$+11.0$).
The relationship with verification time is mild and, as noted below, does not
reach significance.

One stratum runs the other way. Bundles carrying at least one axiom
(specifications that had to assume something about a callee rather than model it)
gain \emph{less} from being supplied ($+8.7$ against $+12.4$), and the difference
is not significant. Axioms mark a specification that could not be made fully
self-contained; they do not mark a harder task.

\paragraph{Which stratum differences survive a test.}
Comparing raw rates across strata compares different instance sets, so we test on
a per-instance quantity instead: for each instance, the number of the four
repetitions that resolved it with a specification minus the number that resolved
it without, divided by four. That is a within-instance difference, and it can be
compared across strata with a Kruskal--Wallis test.
\Cref{tab:strata-tests} reports the results.

\begin{table}[t]
\centering
\caption{Does the specification gain differ across the strata of a family for Opus~4.8? Kruskal--Wallis over the per-instance gain, restricted to strata with $n \ge 20$; for two-stratum families the Mann--Whitney and Welch tests are reported as well.}
\label{tab:strata-tests}
\small
\setlength{\tabcolsep}{5pt}
\begin{tabular}{@{}lrr@{}}
\toprule
\hd{Family} & \hd{Strata} & \hd{$p$ Opus} \\
\midrule
Repository & 7 & $0.214$ \\
Added lines (quartiles) & 4 & $0.188$ \\
Files touched & 2 & $0.010$ \\
Modeled functions & 2 & $0.0021$ \\
Axioms & 2 & $0.578$ \\
Fail-to-pass tests & 4 & $0.0076$ \\
Reference verify time & 4 & $0.260$ \\
Estimated time to fix & 3 & $0.0015$ \\
\bottomrule
\end{tabular}
\end{table}

Four families separate clearly: the number of functions the specification
constrains, the number of files the fix touches, the number of failing tests, and
the estimated time to fix. Four do not: repository, added lines, whether the
bundle carries an axiom, and how long the reference takes to verify. How long the
reference bundle takes to prove says nothing about how hard the instance is for
an agent to fix. The prover's difficulty and the task's difficulty are separate
axes. For the two-stratum families the effect sizes are large: the multi-site stratum
gains $12.1$ points more than the single-site stratum ($p = 0.0021$), and the
multi-file stratum $9.8$ points more than the single-file stratum ($p = 0.010$).

\begin{figure}[!t]
\centering
\includegraphics[width=\textwidth]{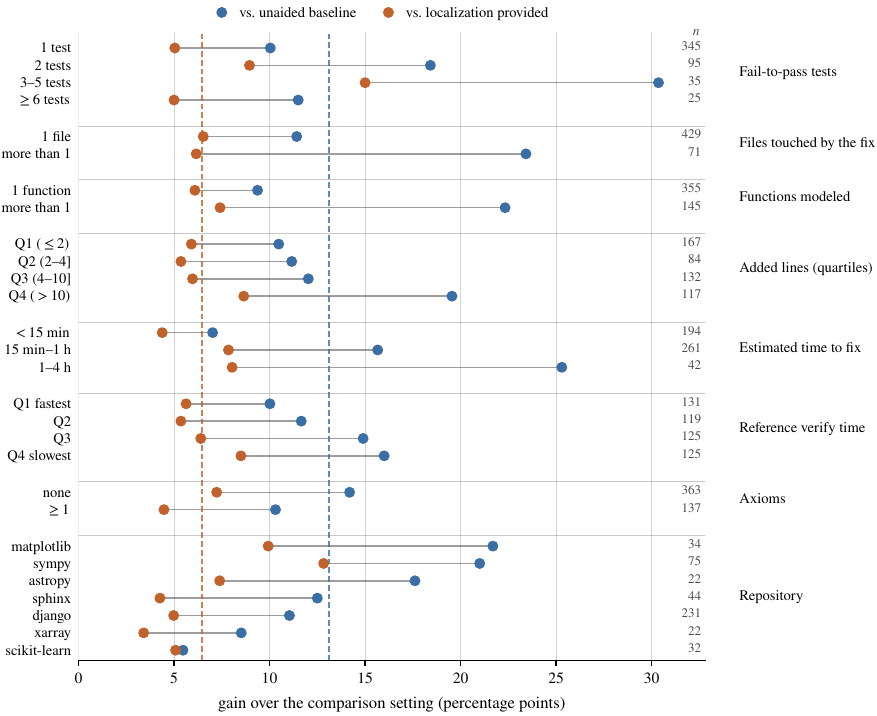}
\caption{What a supplied specification is worth, stratum by stratum, for Opus~4.8
averaged over the four repetitions. For each stratum one marker gives the gain of
Specification Provided (row 6) over the unaided baseline (row 0), and the other
gives the part of that gain which survives when the comparison is Localization
Provided (row 4) instead; the segment between them is what the localization the
specification carries was worth on its own. The dashed rules are the Opus~4.8
values, $+11.4$ and $+4.1$ points. Strata with fewer than $20$ instances are
omitted, and $n$ is the stratum size. Two of the largest total gains keep the
least: fixes touching more than one file gain $+19.8$ over the baseline and
$+0.0$ over row 4, and specifications constraining more than one function gain
$+20.0$ and $+3.6$. The largest residuals sit elsewhere: on instances with three
to five failing tests, and in sympy and matplotlib.}
\label{fig:strata}
\end{figure}

\paragraph{How much of the gain is localization?}
Row 6 supplies both a specification and a localization, so part of its advantage
over the baseline is simply knowing where to look. Row 4 supplies the localization
alone, and the difference between rows 6 and 4 isolates the part attributable to
the specification itself. For Opus~4.8, that residual is $+4.1$ points, about a
third of the $+11.4$ total. Reading it stratum by stratum changes the picture
from the one the total gain suggests. \Cref{fig:strata} plots both quantities
together.

Two of the strata with the largest total gain keep almost none of it once
localization is supplied. Fixes touching more than one file gain $+19.8$ points
over the unaided baseline; over row 4 they gain $+0.0$, which is less than the
$+4.8$ single-file stratum. Specifications constraining more than one function
gain $+20.0$ over the baseline and $+3.6$ over row 4, against $+4.4$ for the
single-function stratum. For these two families, most of what the specification
was worth was the localization it carried.

Other strata do keep their advantage. Instances with three to five failing tests
retain $+10.7$ points, the largest residual in the corpus and more than twice the
model-wide figure. sympy retains $+7.7$ and matplotlib $+8.8$, against $+1.1$ for
sphinx and $+1.2$ for xarray. The largest quartile of reference fixes retains
$+3.5$ and the longest verifying quartile $+4.6$. What a specification adds beyond
a pointer is a statement of the required behavior, and it is worth most where that
statement is hardest to reconstruct.

\paragraph{Correlates of baseline difficulty.}
\Cref{tab:corr} correlates the same properties against the unaided baseline
outcome. Seven properties are significant after a Holm correction for Opus~4.8: the
number of files touched, the estimated time to fix, added lines, added plus
deleted lines, the number of modeled functions, the number of hunks, and deleted
lines. All seven have negative sign and measure how spread out the fix is.
Among those that are not significant are whether the bundle carries an axiom, the number of pass-to-pass
tests, and the reference verification time. How long the reference bundle takes
to prove says nothing about how hard the instance is for an agent to fix. The
prover's difficulty and the task's difficulty are separate axes.

The fail-to-pass count is the instructive case for methodology. Its
point-biserial correlation with the baseline outcome is $+0.008$ for Opus~4.8,
indistinguishable from zero, while its Spearman correlation is $-0.18$ and
comfortably significant. The discrepancy is entirely due to two instances that
carry $438$ and $168$ failing tests against a median of $1$, and that Opus~4.8
resolves. A linear statistic is dominated by those two points; a rank statistic
is not. We report both columns and read the rank one.

\begin{table}[t]
\centering
\caption{Correlation of instance properties with the unaided baseline outcome for Opus~4.8, $n = 500$, first repetition. $r_{pb}$ is the point-biserial correlation with the binary outcome and $\rho$ is Spearman's. ``Holm'' is the Holm-corrected $p$-value of $r_{pb}$ across the properties of that column. ``mean$\mid$1'' and ``mean$\mid$0'' are the property's mean on resolved and unresolved instances. Read $\rho$ rather than $r_{pb}$ for the heavy-tailed count properties.}
\label{tab:corr}
\small
\setlength{\tabcolsep}{3.8pt}
\begin{tabular}{@{}lrrrrr@{}}
\toprule
\hd{Property} & \hd{$r_{pb}$} & \hd{Holm} & \hd{$\rho$} & \hd{mean$\mid$1} & \hd{mean$\mid$0} \\
\midrule
Files touched & $-0.239$ & $8.4\times10^{-7}$ & $-0.254$ & 1.14 & 1.85 \\
Estimated time to fix & $-0.235$ & $1.3\times10^{-6}$ & $-0.224$ & 1.64 & 2.07 \\
Added lines & $-0.218$ & $9.0\times10^{-6}$ & $-0.183$ & 8.31 & 19.23 \\
Added $+$ deleted lines & $-0.209$ & $2.3\times10^{-5}$ & $-0.188$ & 12.23 & 26.25 \\
Modeled functions & $-0.203$ & $4.2\times10^{-5}$ & $-0.223$ & 1.34 & 1.79 \\
Hunks & $-0.155$ & $0.0040$ & $-0.243$ & 2.20 & 3.83 \\
Deleted lines & $-0.133$ & $0.021$ & $-0.143$ & 3.92 & 7.03 \\
Issue text length & $-0.092$ & $0.233$ & $-0.076$ & 1620 & 2150 \\
Reference verify time & $-0.079$ & $0.385$ & $-0.074$ & 25.07 & 30.18 \\
Fail-to-pass tests & $+0.008$ & $1.000$ & $-0.183$ & 3.10 & 2.63 \\
Pass-to-pass tests & $+0.010$ & $1.000$ & $-0.024$ & 121.3 & 114.5 \\
Carries an axiom & $+0.057$ & $0.808$ & $+0.057$ & 0.28 & 0.21 \\
\bottomrule
\end{tabular}
\end{table}

Instances whose bundle carries at least one axiom are resolved at an odds ratio of
$1.47$ relative to those that do not, with a Fisher $p$ of $0.26$ for Opus~4.8.
Carrying an axiom is uncorrelated with whether the instance gets fixed.

\paragraph{What fails in specification synthesis.}
Row 8 has the lowest pass rate in the benchmark. A row-8 pass requires that the agent's
witness verify against its own specification, that the specification not be
vacuously satisfiable, and that it pass the five-property audit.
\Cref{tab:synth-decomp} decomposes all four repetitions.

\begin{table}[t]
\centering
\caption{What fails in specification synthesis for Opus~4.8, per repetition, $n = 500$. The three conjuncts of a pass are shown separately. ``Resolved'' is informational: row 8 does not grade a patch.}
\label{tab:synth-decomp}
\small
\setlength{\tabcolsep}{6pt}
\begin{tabular}{@{}lrrrrr@{}}
\toprule
\hd{Rep.} & \hd{Passed} & \hd{Witness verifies} & \hd{Non-vacuous} & \hd{Audit fails} & \hd{Resolved (info.)} \\
\midrule
1 & 246 & 500 & 500 & 254 & 216 \\
2 & 252 & 500 & 500 & 248 & 226 \\
3 & 240 & 499 & 500 & 259 & 218 \\
4 & 254 & 499 & 500 & 245 & 215 \\
\bottomrule
\end{tabular}
\end{table}

The prover is not the obstacle. Across all four repetitions, the agent's witness
verified against its own specification in $499$ to $500$ of $500$ episodes, and
the non-vacuity check never failed. Every remaining failure is the
audit, and within the audit, faithfulness. Agents can write specifications they
can prove; what they cannot reliably write is specifications that say the right
thing.

Row 8 is also the only setting where the structural properties lose much of their
grip. Correlating them against the row-8 outcome instead of the baseline, only the
estimated time to fix survives the Holm correction ($r_{pb} = -0.195$,
$p = 1.5\times10^{-4}$); added lines is marginal ($r_{pb} = -0.132$, $p = 0.037$)
and the axiom flag is flat (odds ratio $0.91$, $p = 0.69$). Whatever makes an
instance hard to specify faithfully is largely not what makes it hard to fix.

\paragraph{Structured requirements against a prover.}
The \ears{} backend replaces the prover with a structured-requirements document
and a well-formedness check. It is the natural control for how much of the corpus
difficulty comes from formality itself, and it produces the sharpest contrast in
the appendix.

In row 2, where neither backend's artifact is graded and the score is the hidden
tests, the two backends are indistinguishable. Over the $500$ instances, Opus~4.8
resolves $421$ under \ears{} and $428$ under \nagini{}, with $17$ instances
resolved only under \ears{} and $24$ only under \nagini{} (exact McNemar
$p = 0.35$). Writing a structured
requirements document instead of a formal specification neither helps nor
hurts an agent's ability to fix the bug.

Row 3 is where the two backends differ substantially. Under \nagini{}, the agent constructs a formal specification and proves that its implementation satisfies it, with $420$ of $500$ submissions passing. Under \ears{}, an adversarial agent searches for inputs on which the generated patch violates the frozen formal reference specification. Each counterexample is executed against both the generated patch and the ground-truth patch to confirm that it exposes an actual error. This audit accepts $286$ submissions, with $151$ instances passing under \nagini{} but failing under \ears{}, against $17$ in the other direction ($p=5\times10^{-28}$). The audit identifies genuine errors in $32\%$ of \ears{} submissions that pass the hidden tests.

\begin{table}[t]
\centering
\caption{The structured-requirements backend against \nagini{} for Opus~4.8, instance by instance. Rows 2 and 3 carry no localization, so the only difference between the two episodes is the artifact the agent must author and the tool in its verify slot. The resolution columns compare the same hidden-test criterion. For the pass columns, \nagini{} requires a formal proof against the agent-written specification, while \ears{} uses adversarial counterexample search against the frozen reference specification. Formal verification rules out violations of the specification being proved, while counterexample search can only identify violations it discovers.}
\label{tab:ears}
\small
\setlength{\tabcolsep}{4pt}
\begin{tabular}{@{}lrrrrrrr@{}}
\toprule
& & \multicolumn{2}{c}{\hd{Resolved}} & \multicolumn{2}{c}{\hd{Passed}} & \multicolumn{2}{c}{\hd{Discordant}} \\
\cmidrule(lr){3-4}\cmidrule(lr){5-6}\cmidrule(lr){7-8}
\hd{Setting} & \hd{$n$} & \hd{\ears} & \hd{\nagini} & \hd{\ears} & \hd{\nagini} & \hd{\ears\ only} & \hd{\nagini\ only} \\
\midrule
2 End-to-End & 500 & 421 & 428 & 421 & 428 & 17 & 24 \\
3 Verified End-to-End & 500 & 421 & 421 & 286 & 420 & 17 & 151 \\
\bottomrule
\end{tabular}
\end{table}

\paragraph{Formal verification provides a stronger correctness guarantee than counterexample search.} A verified implementation satisfies its formal specification for every input within its stated domain, so no counterexample violating that specification can exist. In contrast, adversarial testing can only identify violations it successfully discovers. In Row 7, the formal backends verify implementations against \benchmark{}'s supplied reference specifications, providing this guarantee directly. In Row 3, the formal backends verify against agent-written specifications, whose faithfulness is evaluated separately in Sections~\ref{sec:res-synth} and~\ref{sec:res-impact}. \ears{} requires no prover time and is inexpensive to construct, but its correctness evaluation relies on finite counterexample search. Its documents contain an average of $2.3$ requirements and $8.2$ acceptance criteria for Opus~4.8, nearly all conditionally triggered.

The cost side of the same comparison is one-directional and large. On matched
cells, an \ears{} episode uses $4$ to $5$ fewer agent steps and $113$ to
$137$ thousand fewer input tokens than the \nagini{} episode of the same setting,
and spends zero seconds in a solver against $85$ to $101$ seconds per episode
under \nagini{}.

\subsection{Interaction and cost}
\label{app:cost}

This subsection reports what the evaluation cost. Tokens and agent steps are the
comparable measures; wall-clock is reported for completeness but is a property of
the host, since twelve episodes ran concurrently throughout.

\paragraph{Accounting.}
An episode's input count is the sum of the prompt tokens of its individual model
calls. Prompt caching was not in effect in any episode of any run, so there is no
distinction between billed and presented input. The verifier wall-clock is the sum
over an episode's verify invocations of the time the verifier held the call; for
\ears{} it is $0.0$ by rounding, because the structural check returns in well under
a tenth of a second. Every figure below is pooled over the complete repetitions of
a cell only.

\begin{table}[t]
\centering
\caption{Cost per episode by setting and backend for Opus~4.8, pooled over the complete repetitions of the cell. Input and output are means in thousands of tokens; steps is mean agent steps; wall is mean episode seconds under twelve-way concurrency. ``Verify calls'' is the mean number of verify invocations. ``Input per resolved'' is the cell's total input tokens divided by the number of instances it resolved, in thousands: the cost of an outcome rather than of an episode.}
\label{tab:cost}
\small
\setlength{\tabcolsep}{3pt}
\begin{tabular}{@{}lcrrrrrrr@{}}
\toprule
\hd{Setting} & \hd{Backend} & \hd{$n$} & \hd{in (k)} & \hd{out (k)} & \hd{steps} & \hd{wall (s)} & \hd{v.\ calls} & \hd{in/resolved} \\
\midrule
0 Unaided baseline & \na & 2000 & 200.2 & 4.11 & 15.6 & 289.2 & 0.00 & 235.4 \\
4 Localization Provided & \na & 2000 & 140.1 & 3.04 & 11.5 & 231.4 & 0.00 & 151.8 \\
\midrule
2 End-to-End & \nagini & 2000 & 438.5 & 7.81 & 25.7 & 421.3 & 2.75 & 517.0 \\
3 Verified End-to-End & \nagini & 2000 & 437.8 & 8.40 & 26.2 & 418.8 & 2.96 & 516.2 \\
6 Specification Provided & \nagini & 2000 & 375.6 & 7.16 & 21.5 & 367.2 & 3.48 & 389.4 \\
7 Verified from Spec. & \nagini & 2000 & 353.8 & 7.00 & 21.0 & 314.4 & 3.52 & 368.0 \\
8 Specification Synthesis & \nagini & 2000 & 334.8 & 7.97 & 19.8 & 920.8 & 5.16 & 675.1\rlap{$^\dagger$} \\
\midrule
2 End-to-End & \ears & 500 & 325.5 & 6.56 & 21.2 & 204.2 & 1.04 & 386.6 \\
3 Verified End-to-End & \ears & 500 & 301.2 & 6.39 & 21.0 & 162.9 & 1.05 & 503.7\rlap{$^\dagger$} \\
\bottomrule
\multicolumn{9}{@{}p{\textwidth}@{}}{\footnotesize $^\dagger$ per passing episode; row 8 does not grade a patch, and row 3 under \ears{} passes fewer episodes than it resolves.}
\end{tabular}
\end{table}

\begin{figure}[htbp]
\centering
\includegraphics{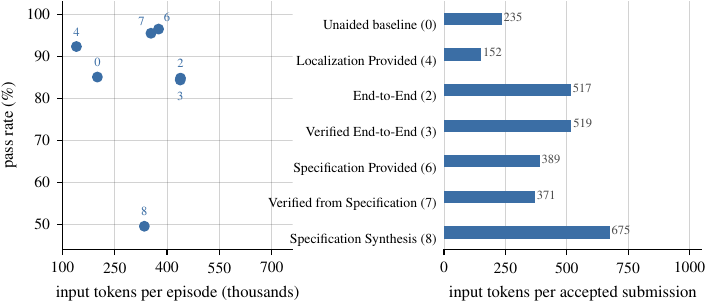}
\caption{What each setting costs and what it buys for Opus~4.8, contract-annotated
Python backend, four repetitions pooled. \textbf{Left:} pass rate against input
tokens per episode, one point per setting, labeled by row number; up and to the
left is better. \textbf{Right:} the two quantities combined as input tokens per
accepted submission, which is the scale on which settings of different accuracy
can be compared. Localization Provided is the cheapest setting per accepted
submission, and requiring a verifying artifact costs about what the same setting
costs without one.}
\label{fig:cost}
\end{figure}

\paragraph{Cost tracks the formal obligation.}
Ordering the \nagini{} cells by input tokens per episode gives the same order as
ordering them by how much the setting asks the agent to construct: localization
provided ($140$ thousand), unaided baseline ($200$), specification synthesis
($335$), verified from specification ($354$), specification provided ($376$), and
end-to-end ($438$). Supplying a specification is \emph{cheaper} than not
supplying one: row 6 costs $63$ thousand fewer tokens per episode than row 2,
while resolving $12$ more points of the corpus. The cost of an outcome separates
even more sharply: $389$ thousand input tokens per resolved instance in row 6
against $517$ in row 2 for Opus~4.8. \Cref{fig:cost} shows both scales.

Adding the verification obligation on top of a supplied specification is
essentially free: rows 6 and 7 differ by under $6\%$ in tokens and under $3\%$ in
steps. The agent was already using the verifier in row 6, where it had no
obligation to; formalizing that obligation changes the score by about a point and
the cost by almost nothing.

Row 8 is where wall-clock and tokens disagree most, and \cref{tab:vtail} explains
why. It is the second cheapest \nagini{} setting in tokens for Opus~4.8, but the
most expensive in wall-clock by more than a factor of two ($921$ seconds per
episode). The difference is prover time: row 8 spends $139$ compute-hours in the
verifier, more than any other setting.

\paragraph{The step budget rarely binds.}
Every episode had a budget of $250$ agent steps. Of the campaign's $53{,}500$
episodes, $53{,}465$ ended by submitting and $25$ reached the budget. The
submission rate is $100.0\%$ in $94$ of the $107$ runs and above $99.5\%$ in all
but one: \velvet{} in row 3, where $21$ episodes ran out of steps while still
working on a proof, which costs that cell about four points of resolution.
Elsewhere the step distributions of \cref{tab:vtail} put the $99$th percentile
between $91$ and $103$, so the reported rates are not truncated by the
interaction limit.

\paragraph{Repetition-to-repetition cost stability.}
Within a cell, the ratio of the largest to the smallest per-repetition mean is
$1.03$ to $1.22$ for input tokens, $1.02$ to $1.13$ for output tokens, and $1.01$
to $1.06$ for steps. For wall-clock it is $1.02$ to $1.79$. Token and step costs
are a stable property of a cell; wall-clock is not, because different repetitions
ran under different host load. Any comparison in this appendix that relies on
wall-clock is either within a single run or is stated as a ratio to that run's own
total.

\paragraph{Campaign totals.}
\Cref{tab:totals} reports the whole evaluation: $107$ runs and $53{,}500$
episodes, which consumed $22.6$ billion input tokens and $415$ million output
tokens over $1.36$ million agent steps, $167{,}719$ verifier invocations,
$5{,}937$ hours of episode wall-clock, and $896$ hours of verifier wall-clock.

\begin{table}[t]
\centering
\caption{Cost of the evaluation campaign, with the per-backend split. Wall-clock
hours are compute-hours summed over episodes, not elapsed time. The backend rows
partition the campaign, so they sum to the first row.}
\label{tab:totals}
\small
\setlength{\tabcolsep}{5pt}
\begin{tabular}{@{}lrrrrrrr@{}}
\toprule
\hd{Split} & \hd{Runs} & \hd{Episodes} & \hd{Input (B)} & \hd{Output (M)}
  & \hd{Steps} & \hd{Verify calls} & \hd{Episode h} \\
\midrule
Whole campaign & 107 & 53{,}500 & 22.63 & 414.6 & 1{,}360{,}874 & 167{,}719 & 5{,}937.1 \\
\midrule
\quad \nagini & 72 & 36{,}000 & 14.64 & 291.7 & 886{,}175 & 102{,}605 & 4{,}292.7 \\
\quad \velvet & 12 &  6{,}000 &  3.99 &  51.1 & 229{,}123 &  47{,}825 &   763.4 \\
\quad \lean & 12 &  6{,}000 &  2.21 &  41.4 & 135{,}077 &  14{,}240 &   592.2 \\
\quad \ears & 11 &  5{,}500 &  1.78 &  30.3 & 110{,}499 &   3{,}049 &   288.8 \\
\bottomrule
\end{tabular}
\end{table}

Two ratios follow from that table. Of the $167{,}719$ verifier invocations,
$68{,}793$ returned an accepting verdict, $41\%$ overall; the per-cell spread
behind that average is the backend-difficulty measurement of
\cref{app:verif-effort}. And the mean prover call held the prover for $19.6$
seconds, against $25.8$ seconds for a reference verification on the same corpus,
so an agent's proof obligation is no more expensive to discharge than the
reference one.

\section{Worked examples}
\label{app:examples}

The preceding sections describe the corpora in aggregate. This section does the
opposite. It follows a small number of instances all the way down, so that a
reader can see what a specification actually says, what the evidence attached to
it actually establishes, and where the construction is hard.

Two examples carry the section, and the first two subsections share one of them.
\Cref{app:four-languages} takes a single Django issue and prints the four
specifications the four backends produce for it side by side, together with the
admission evidence each one carries. \Cref{app:attack-suite} stays on that
instance and reproduces the full adversarial record for one of the four
specifications: twenty candidate implementations written to break it, the
mechanism by which each was rejected, the one that verified, and the controls that
decide whether a candidate that verifies is a specification weakness or a
legitimate alternative implementation. \Cref{app:faithfulness} then turns to a
scikit-learn issue on which two of the properties we require of a specification
cannot both hold, and states what we do about it.

The instance in the first two subsections was chosen because it is small enough
to print in full and unusual in what it demands of a specification: the reference
fix is a single added line, all four backends admit the instance, and the
specification that line has to satisfy is nonetheless the most intricate of the
four. It demonstrates that the difficulty of specifying a change is not
proportional to the size of the change.

\subsection{One task, four specifications}
\label{app:four-languages}

\paragraph{The task.}
\mbox{\id{django-11179}} reports an inconsistency in Django's deletion machinery. When
an object is deleted through the ORM, the deletion code clears the object's
primary-key attribute in memory, which is how an in-memory object records that it
no longer corresponds to a stored row. The machinery also contains a fast path
for the simplest possible job: one collected object, of one model, with no
dependent relations. That path issues the delete and returns immediately,
skipping the bookkeeping that clears primary keys. Deleting a dependency-free
object therefore left it holding a stale identifier, unlike every other deletion.
The public problem statement says exactly this and no more: it names the fast
path, states that the primary key is not cleared on it, and asks for the reset.

\begin{patchbox}[label=box:11179-patch]{The reference fix for \id{django-11179}}
\begin{minted}[escapeinside=,breakanywhere=true]{diff}
--- a/django/db/models/deletion.py
+++ b/django/db/models/deletion.py
@@ -277,6 +277,7 @@ def delete(self):
             if self.can_fast_delete(instance):
                 with transaction.mark_for_rollback_on_error():
                     count = sql.DeleteQuery(model).delete_batch([instance.pk], self.using)
+                setattr(instance, model._meta.pk.attname, None)
                 return count, {model._meta.label: count}
\end{minted}
\end{patchbox}

Box~\ref{box:11179-patch} is the whole fix: one added line, in one file, inside
one branch. The task's hidden tests check that the attribute is cleared after a
fast delete and that the fast path still reports the same counts and still
performs exactly one query.

\paragraph{What a specification for it has to pin down.}
The change is one assignment, but the behavior it establishes has several
independent facets, and a specification that misses any of them is satisfied by
code the tests reject. The identifier must end up absent. It must be the model's
real primary-key attribute that is cleared, which need not be named \texttt{id}.
The reset must happen after the delete has been issued, so that a failed delete
does not leave an object claiming to be gone. The reported counts must be
unchanged. And the reset must cost nothing: it is an in-memory assignment, and an
implementation that consulted the database to decide what to clear would pay a
query the fast path does not have.

The four backends divide these facets differently. Two of them state only the
first and the fourth; one states all of them; one states them as prose criteria
and checks them by probe. What follows is each specification as released, with
its narrative comments removed and its structure otherwise intact.

\paragraph{Contract-annotated Python.}
The \nagini specification is the only one of the four that models the cost of the
fix, and doing so forces it to model the database handle as well.
Box~\ref{box:11179-nagini} shows the solver-visible view. Three ghost resources
carry the state that the contract talks about. \texttt{RemovalBudget} is the
entitlement to issue one row-removal statement against the stored data; it is
held rather than read, so an implementation can spend it but can learn nothing
from it. \texttt{CarriedOutRemoval} and \texttt{RefusedRemoval} are the store's
two possible receipts for a statement it was asked to apply. Only the removal
primitive mints a receipt, and no implementation can alter the handle, the
identifier, or the count recorded on one. That primitive is the instance's single
axiom, shown earlier as Box~\ref{box:axiom}, and it is the code the fix leaves
alone.

\begin{specbox}[label=box:11179-nagini]{\id{django-11179} under contract-annotated Python: the solver-visible specification (narrative comments removed)}
\begin{minted}{python}
@ContractOnly
@Predicate
def RemovalBudget(ledger: RemovalLedger) -> bool: pass

@ContractOnly
@Predicate
def CarriedOutRemoval(db: Database, identifier: int, count: int) -> bool: pass

@ContractOnly
@Predicate
def RefusedRemoval(db: Database, identifier: int) -> bool: pass

def remove_single_independent(instance: PersistentObject, db: Database) -> int:
    Requires(MustTerminate(2))
    Requires(Acc(instance.pk))
    Requires(instance.pk is not None)
    Requires(-(2 ** 63) <= instance.pk and instance.pk < 2 ** 63)
    Requires(RemovalBudget(LEDGER))
    Ensures(Acc(instance.pk))
    Ensures(CarriedOutRemoval(db, Old(instance.pk), Result()))
    Ensures(Result() >= 0)
    Ensures(instance.pk is None)
    Exsures(RemovalRefused, Acc(instance.pk)
            and RefusedRemoval(db, Old(instance.pk))
            and instance.pk is Old(instance.pk))
\end{minted}
\end{specbox}

Read the postconditions as a group and each facet appears. The returning exit
must hold a carried-out receipt naming the handle it was given, the identifier
the object carried \emph{on entry}, and exactly the count being returned: the
statement was really issued, for the right row, and the number surfaced is that
statement's own report rather than a figure read off the store afterwards. The
identifier must be absent on that exit and unchanged on the other, so the object
gives up its identity exactly when the row it names has been removed. And the
entitlement is stated as a precondition of the primitive and never returned, so
the whole method may issue one statement and no more. The permission
\texttt{Acc(instance.pk)} is required on entry and re-established on both exits.
No permission on any of the store's own tallies is available anywhere in the
contract, so reading them is impossible.

\paragraph{An imperative method in an embedded DSL.}
The \velvet specification, Box~\ref{box:11179-velvet}, takes the opposite view of
what the instance is about. The removal count is an input rather than something
the method obtains, so the batch removal is not re-derived and is observed only
through that number. What remains is the state change itself, and the method
declares the identity handle mutable and pins its post-state exactly.

\begin{specbox}[label=box:11179-velvet]{\id{django-11179} under the imperative DSL}
\begin{minted}[escapeinside=]{lean}
set_option loom.semantics.termination "total"
set_option loom.semantics.choice "demonic"

open TotalCorrectness DemonicChoice

method fastDelete (mut pk : Option Int) (removed : Nat) return (count : Nat)
  ensures pk = none
  ensures count = removed
  do
    pk := none
    return removed

prove_correct fastDelete by
  loom_solve
\end{minted}
\end{specbox}

Two postconditions do all the work. The first pins the changed part of the state
to its exact post-value; the second pins the unchanged part by stating that the
returned count equals the count that came in. Without that second clause, an
implementation that clears the handle and reports a different number satisfies
the specification, and the task's own tests would reject it. The correctness
obligation is handed to the solver-backed tactic and discharged without a
hand-written proof.

\paragraph{Pure functional Lean.}
The \lean specification, Box~\ref{box:11179-lean}, is the only one that models
the whole collected job rather than one object. Records are grouped by kind; each
carries its optional primary key; the outcome is the pair of the updated groups
and the reported count. The two paths differ in how they tally, so the count is
pinned to a branch-selected value, but the key-clearing outcome is identical on
both paths and is pinned structurally.

\begin{specbox}[label=box:11179-lean]{\id{django-11179} under pure functional \lean (definitions and theorem statement; proof omitted)}
\begin{minted}[escapeinside=]{lean}
structure Instance where
  pk : Option Int
deriving DecidableEq

abbrev Label := Nat
abbrev RecGroup := Label × List Instance

def clearAllPks (data : List RecGroup) : List RecGroup :=
  data.map (fun g => (g.1, g.2.map (fun i => { i with pk := none })))

def isSingleFast (data : List RecGroup) (directlyDeletable : Instance → Bool) : Bool :=
  match data with
  | [(_, [inst])] => directlyDeletable inst
  | _ => false

def problem_spec_delete
    (data : List RecGroup) (directlyDeletable : Instance → Bool)
    (directCount generalTotal : Int)
    (result : (List RecGroup) × Int) : Prop :=
  result.1 = clearAllPks data ∧
  result.2 = (if isSingleFast data directlyDeletable then directCount else generalTotal)

theorem correct_delete :
    ∀ (data : List RecGroup) (directlyDeletable : Instance → Bool)
      (directCount generalTotal : Int),
      problem_spec_delete data directlyDeletable directCount generalTotal
        (implementation_delete data directlyDeletable directCount generalTotal)
\end{minted}
\end{specbox}

Note the universally quantified \texttt{directlyDeletable}. Whether a given
object qualifies for the direct single-row removal is an unchanged decision that
the fix does not touch. The specification quantifies over every such predicate
instead of axiomatizing it, so the theorem holds for all of them. That is the
general pattern behind the axiom counts of \cref{app:axioms}: a language that can
abstract over the unchanged decision does not need to assume anything about it.
The proof is an ordinary kernel-checked term; it splits on the shape of the
collected groups and on the predicate, and closes each case by simplification.

\paragraph{Structured natural language.}
The \ears requirements document, excerpted in Box~\ref{box:11179-ears}, also
states every facet, because prose is not constrained by
what a solver can be made to accept. It is also the only one with no verifier
behind it, so what it buys is a precise statement of intent rather than a proof.

\begin{specbox}[label=box:11179-ears]{\id{django-11179} under structured natural language (acceptance criteria, abridged)}
\footnotesize
\textbf{Requirement 1: clear the primary key on the fast-delete path.}
\begin{enumerate}[label=1.\arabic*.,leftmargin=2.6em,itemsep=1pt,topsep=2pt]
\item WHEN a single collected instance of a single model qualifies for fast
      deletion and is deleted via the collector's fast path, THEN the deletion
      machinery SHALL set that instance's primary-key attribute to
      \texttt{None}.
\item WHEN a dependency-free model instance is deleted through the model's
      \texttt{delete()} method, THEN, after the call returns, the deletion
      machinery SHALL leave the instance's primary-key attribute equal to
      \texttt{None}.
\item The system SHALL clear the primary-key attribute using the model's actual
      primary-key attribute name, so that models whose primary key is not
      literally named \texttt{id} also have the correct attribute reset.
\item WHEN the fast path clears the primary key, THEN the deletion machinery
      SHALL perform the reset after the database delete has been issued, so that
      a failure of the delete operation does not leave the instance reporting a
      cleared key for a row that still exists.
\end{enumerate}
\smallskip
\textbf{Requirement 2: preserve fast-path count and return semantics.}
\begin{enumerate}[label=2.\arabic*.,leftmargin=2.6em,itemsep=1pt,topsep=2pt]
\item WHEN a dependency-free instance is deleted via the fast path, THEN the
      deletion machinery SHALL return the total number of deleted objects
      together with a per-model-label mapping of deletion counts, exactly as
      before the primary-key reset was introduced.
\item The system SHALL leave the number of database operations performed by the
      fast path unchanged; clearing the primary key SHALL be an in-memory
      attribute assignment that issues no additional query.
\item IF the collected work does not match the single-instance, single-model,
      fast-deletable case, THEN the deletion machinery SHALL NOT take the fast
      path and SHALL instead follow the general deletion path.
\end{enumerate}
\smallskip
\textbf{Requirement 3} (six criteria, abridged) preserves the general path's
key-clearing, the queued field updates, the bulk-deletion counts, the
pre-delete and post-delete signal ordering, the configured on-delete behaviors,
and the fast-deletability decision itself.
\end{specbox}

\paragraph{What each specification pins, and what it leaves open.}
\cref{tab:11179-facets} lays the four side by side. They do not agree on how much
of the change to model. Each is required to be strong enough that the pre-fix
behavior fails it and weak enough that legitimate implementations pass, and there
are many specifications between those two bounds.

\begin{table}[htbp]
\centering
\caption{The same task under four backends. Each row is a facet of the behavior
the reference fix establishes; each cell states how that backend's specification
pins it, or that it does not. ``Not modeled'' means the facet is outside what
that specification talks about, which is admissible as long as the pre-fix
behavior still fails; the discrimination check of \cref{app:mechanical} enforces
that.}
\label{tab:11179-facets}
\small
\setlength{\tabcolsep}{3pt}
\begin{tabular}{@{}L{2.9cm}L{2.4cm}L{2.25cm}L{2.4cm}L{2.25cm}@{}}
\toprule
\textbf{Facet} & \textbf{\nagini} & \textbf{\velvet} & \textbf{\lean}
& \textbf{\ears} \\
\midrule
Modeled scope & one object and the handle it is removed through & the object's
identity handle and the reported count & every object in every collected group &
the deletion path in place \\
\addlinespace[1.5pt]
Carrier of the key & heap field of optional integer type, with explicit access
permission & mutable parameter of optional integer type & optional-integer field
of a record inside labeled groups & the model's primary-key attribute, by name \\
\addlinespace[1.5pt]
Key is cleared & absent on the returning exit & pinned to absent & every record
in every group, structurally & criteria 1.1--1.3 \\
\addlinespace[1.5pt]
Counts unchanged & equal to the count the receipt names & equal to the count
supplied & equal to the branch-selected tally & criterion 2.1 \\
\addlinespace[1.5pt]
Cost of the fix & one removal statement, held as a consumable entitlement & not
modeled & not modeled & criterion 2.2 \\
\addlinespace[1.5pt]
Reset happens after the delete & pinned on both exits, one per receipt &
not modeled & not modeled & criterion 1.4 \\
\addlinespace[1.5pt]
Store may refuse & second exit with its own postcondition & not modeled & not
modeled & not stated \\
\addlinespace[1.5pt]
Correct attribute name & modeled as the named field the attribute stands for &
not modeled & not modeled & criterion 1.3 \\
\midrule
Unchanged callees assumed & 1 & 0 & 0 & \na \\
Obligation discharged by & SMT solver, via the intermediate verifier &
solver-backed tactic & kernel-checked proof term & reading, plus executable
probes \\
\bottomrule
\end{tabular}
\end{table}

\paragraph{The pre-fix twin, four times.}
Each of the three prover backends ships the same specification paired with the
behavior the issue complains about, and each of those three must fail. They fail
for different reasons. Under contract-annotated Python the twin drops the
assignment and the verifier reports that the postcondition placing the identifier
absent might not hold. Under the imperative DSL the twin returns the handle
unchanged and the solver refuses the obligation directly. Under pure \lean the
twin returns the collected groups untouched on the fast branch, and the theorem
no longer closes because the structural equality with the cleared groups fails on
the single-record case. The \ears bundle has no verifier, so its analogue is the
probe suite: the same six requirement probes are run against the repository
before and after the fix, and they must separate the two trees.

\paragraph{The evidence each bundle carries.}
\cref{tab:11179-evidence} reports what was actually run for this instance. The
\nagini bundle carries one axiom, so it carries an axiom probe: $93$ recorded
observations of Django's real deletion primitive, covering live rows, absent
identifiers, and the extremes of the stated $64$-bit domain. The other two
prover bundles have no axioms and therefore no probe. The \nagini reference
body performs an assignment and returns a value it did not compute, so there
are no operators to mutate and the mutation check does not apply; the twin, the
property test suite (whose discrimination lane holds $36$ of its $40$ cases), and
the differential harness carry the burden. Under the imperative DSL the whole
body is tagged as reference, so the differential check does not apply, while one
mutant of the body is generated and rejected.

\begin{table}[htbp]
\centering
\caption{Admission evidence recorded for \id{django-11179}, per backend.
Property test suite cases are split into the admissibility, soundness, and discrimination
lanes of \cref{app:evidence}. ``Attacks'' counts the candidate implementations an
adversarial audit submitted against the specification and the number that broke
it. A dash marks a check that does not apply, with the reason given in the
text.}
\label{tab:11179-evidence}
\small
\setlength{\tabcolsep}{5pt}
\begin{tabular}{@{}lcccccc@{}}
\toprule
\textbf{Backend} & \textbf{Verify} & \textbf{Twin} & \textbf{Suite}
& \textbf{Attacks} & \textbf{Mutants} & \textbf{Differential} \\
\midrule
\nagini & 19.9\,s & fails & 40 \,(1/3/36) & 0 of 19 break & \na
& 100{,}000 inputs, 0 disagree \\
\velvet & 3.5\,s & fails & 6 \,(1/2/3) & 0 of 9 break & 1/1 killed & \na \\
\lean & 7.8\,s & fails & 18 \,(1/3/14) & 0 of 7 break & 8/8 killed & pass \\
\ears & \na & \na & \na & 0 of 1 breaks & \na & 6 probes hold \\
\bottomrule
\end{tabular}
\end{table}

The differential figure for \nagini comes from $100{,}000$ generated inputs. The
harness ran Django's patched fast branch on each of them inside the task's
container, ran the modeled reference on the corresponding modeled inputs, and
found no disagreement. Of those runs, $33{,}217$ produced a refused removal
rather than a count, which is the outcome the second exit of the contract
describes. About a third of the observed runs take that exit, so the
postcondition on it was exercised against the real primitive.

\paragraph{What the four-way comparison shows.}
Three things, none of which is visible from the aggregate tables.

First, verification cost tracks how much of the environment a specification
models, not how large the code change is. The same one-line fix takes $3.5$
seconds to verify in the DSL that models the object alone and $19.9$ seconds in
the contract language that models the handle, the entitlement, and both exits.

Second, a specification that models more is harder to write and harder to game,
and the adversarial effort spent on it reflects that. The contract-annotated
specification is the only one of the four in which reading the store's tallies is
even expressible as an attack, and it is the only one whose audit needed twenty
candidate implementations; the DSL specification, which says nothing about the
store, needed nine. The count of attacks reflects attack surface.

Third, one instance carries four different difficulties under a single reference
fix. A solver asked to satisfy the DSL specification must produce a state
mutation that preserves a count. A solver asked to satisfy the contract-annotated
one must additionally reason about a consumable resource and two exits, and must
do so without reading state it has no permission to touch. Both are legitimate
formalizations of the same issue, and reporting per-backend results separately, as
\cref{app:results} does, is a consequence: the numbers are not measuring the same
task in four notations.

\subsection{Twenty attacks on one specification, and how each is calibrated}
\label{app:attack-suite}

A specification that verifies its own reference implementation has established
almost nothing. The property that matters is that it \emph{rejects} the wrong
implementations, and that property cannot be checked by inspection. The attacker
panel of \cref{app:audit} therefore attacks each specification directly: an agent
with the specification view, the repository, and the verifier writes candidate
implementations intended to satisfy the contract while doing something the issue
forbids, and each candidate is run through the verifier. A candidate that
verifies is a finding, and it is then checked against the task's tests to
determine whether it is a specification weakness or a legitimate reshaping of the
reference.

This subsection reproduces the record for the contract-annotated specification of
\mbox{\id{django-11179}} in full. Twenty candidates were submitted. Nineteen reached the
verifier, and none of them broke the specification; one could not be translated
and is counted in neither tally. \cref{tab:11179-attacks} lists all twenty,
grouped by what the candidate was trying to get away with, together with the
mechanism that stopped it.

\begin{table}[htbp]
\centering
\caption{The twenty candidate implementations submitted against the
contract-annotated specification of \mbox{\id{django-11179}}, grouped by intent. Each
was written to satisfy the contract while violating something the issue requires.
The right column names the mechanism that rejected it. Candidate 16 verifies; its
materialized patch also passes the task's hidden tests, so it is a legitimate
alternative implementation rather than a weakness. Candidate 15 was rejected by
the translation layer before reaching the solver and counts in neither tally.}
\label{tab:11179-attacks}
\small
\setlength{\tabcolsep}{4pt}
\begin{tabular}{@{}rL{5.5cm}L{6.2cm}@{}}
\toprule
& \textbf{What the candidate does} & \textbf{Why it fails} \\
\midrule
\multicolumn{3}{@{}l}{\emph{Do less than the fix asks}} \\
1 & Re-encodes the reported bug: issue the removal, skip the reset & The
postcondition placing the identifier absent might not hold \\
6 & Clears the identifier only when it is nonzero & Same postcondition, on the
excluded part of the domain \\
\addlinespace[2pt]
\multicolumn{3}{@{}l}{\emph{Pay for information the fast path does not have}} \\
2 & Reads the store's count of carried-out removals and returns it & Insufficient
permission to access that field \\
3 & Reads the module-wide total of issued statements & Insufficient permission \\
4 & Discards the removal's own return value and echoes the store's report of the
last removal instead & Insufficient permission; the echo has no readable field
left to echo \\
14 & Fabricates a count without a matching receipt & No carried-out receipt for
the identifier and count returned \\
20 & Obtains a receipt for one identifier and returns it as though it were
another's & The receipt named in the postcondition is not the one held \\
\addlinespace[2pt]
\multicolumn{3}{@{}l}{\emph{Spend more than one removal statement}} \\
5 & Rebinds the handle parameter to a freshly constructed store and removes
there & No carried-out receipt against the handle the caller supplied \\
7 & At the top of the representable range, spends the statement on the
neighboring identifier & No receipt for the identifier the object carried on
entry \\
9 & Removes once, then removes again through a self-made handle & The entitlement
required at the second call site is no longer available \\
10 & Retries the removal after the store refuses & Same: the entitlement was
spent by the refused statement \\
\addlinespace[2pt]
\multicolumn{3}{@{}l}{\emph{Take an exit that is not available}} \\
11 & Manufactures a refusal without asking the store & No refusal receipt \\
12 & Has the removal carried out, then reports a refusal & Holds the carried-out
receipt, not the refusal one \\
18 & Manufactures a refusal only at the bottom of the representable range & No
refusal receipt \\
19 & Pays for the removal, holds the receipt, then raises an unrelated exception
on a condition the store's answer decides & The method may raise only the refusal
exception \\
\addlinespace[2pt]
\multicolumn{3}{@{}l}{\emph{Get the ordering wrong}} \\
13 & Clears the identifier before asking the store, and does not restore it on a
refusal & On the refusing exit, the identifier is required to be the one the
object came in with \\
16 & Clears the identifier before asking, restores it on a refusal, reads
nothing & \textbf{Verifies.} Behaviorally equivalent; its patch resolves the
task \\
\addlinespace[2pt]
\multicolumn{3}{@{}l}{\emph{Escape the obligation instead of discharging it}} \\
8 & Does the job, then diverges in a loop & The loop does not carry the
termination obligation out of the surrounding context \\
17 & Recurses on itself under a guard that appears unreachable & The self-call
does not carry the caller's termination obligation \\
15 & Attempts to forge the entitlement by folding the ghost resource directly &
Rejected by the translation layer; unexpressible \\
\bottomrule
\end{tabular}
\end{table}

\paragraph{A family that was once open.}
The second group of \cref{tab:11179-attacks} is in the suite because an earlier
version of this specification failed it. An earlier audit found two verifying
wrong implementations: one read the store's own count of carried-out removals
before issuing the delete and restored the identifier if the delete was refused,
and one discarded the removal's return value and echoed the store's report of
the last removal instead. Both verified against the specification as it then
stood, and neither is a correct program: obtaining either quantity in reality
costs a query the fast path does not have.

The repair was structural rather than a patched-in clause. Instead of adding
postconditions to forbid the two reads, the specification was rewritten so that
no access permission on any of the store's fields reaches the modeled path at
all, and the information the path legitimately has was moved into the three ghost
receipts. The original attacks were then re-encoded verbatim and re-run, and all are
now rejected. The rejection happens at the read rather than at a
postcondition, because the branch that performs the read cannot be written. A
specification patched clause-by-clause invites the same family to reappear
parameterized on the next unpinned difference; a specification from which the
whole class of unpriced reads is inexpressible does not.

\paragraph{The families are not interchangeable.}
Each group in \cref{tab:11179-attacks} is stopped by a different part of the
specification, and removing any one part would open a family. The permission
discipline stops the second group: because no access to the store's own fields is
granted anywhere in the contract, a candidate that reads them does not fail a
postcondition, it fails to compile a read. The entitlement stops the third group:
one statement is available, it is consumed by the removal primitive on either
answer, and nothing mints another, so a second removal has no precondition to
satisfy at its call site. The two receipts stop the fourth: an implementation
cannot mint either one, so it cannot reach an exit whose postcondition names a
receipt it does not hold. The termination measure stops the sixth, and it has to
be stated at the method level to do so; loop invariants alone leave the diverging
candidate verifying.

\paragraph{Attacks that target the boundary rather than the behavior.}
Candidates 7 and 18 attack the domain rather than the postconditions. Both work
at the extremes of the representable identifier range, where a specification
stated over unbounded integers and an implementation working on machine integers
can come apart. Candidate 7 spends the one available statement on the identifier
immediately below the one the object carries, which is a distinct value inside the
domain; candidate 18 arranges the refusing exit only at the very bottom of the
range. Both are rejected on the same grounds as their non-boundary counterparts:
the receipts name the identifier, and naming it leaves no room for a near miss.

\paragraph{The candidate that could not be judged.}
Candidate 15 tried to forge the entitlement by constructing the ghost resource
directly rather than receiving it. The translation layer rejected the program
before the solver saw it, so no verdict was produced. We record this outcome as
its own category and count it in neither tally, because a candidate that the
tooling cannot express is evidence about the tooling and not about the
specification. Across the corpus these cases are rare, and each is recorded with
the reason the translation failed.

\paragraph{Calibrating an attack against the task's own tests.}
An attack that verifies is not yet a finding. The specification is supposed to
admit every correct implementation, so a candidate that verifies may simply be a
correct implementation written differently from the reference. The two cases are
distinguished by materializing the candidate as a repository patch and running it
against the task's hidden tests. If it resolves the task, the specification
correctly admitted a correct program. If it does not, the specification admitted
a program the task rejects, and that is a weakness.

Three controls anchor the suite, and together they show the calibration working
in both directions.

\paragraph{The candidate that verifies, and should.}
Candidate 16 clears the identifier before asking the store and restores it if the
store refuses. It reads nothing it has no permission to read, holds the correct
receipt on each exit, and leaves the identifier absent exactly when a removal was
carried out. It verifies, in $15.8$ seconds. Materialized as a patch and run
against the task's tests, it resolves the instance: the one fail-to-pass test
passes and all $40$ pass-to-pass tests pass. The reordering is
observationally invisible: the difference between clearing before and clearing
after is visible only if the delete fails, and the specification's second exit
requires the identifier to be restored in exactly that case.

\paragraph{The candidate that would have been a finding.}
Candidate 2 reads the store's own count of carried-out removals and returns it
instead of the count the removal statement reported. Under the specification it
is rejected outright, for lack of permission on that field. To check that the
rejection was right, the candidate's behavior was written out as a faithful
repository patch and run against the tests. It does not resolve the task. It
fails a pass-to-pass test that asserts the fast path performs exactly one query,
because obtaining the store's tally in reality requires a second one. Three
independent runs agree. The specification's refusal to grant that permission is
the formal counterpart of a test the task already has.

\paragraph{The reference itself, and the check that the proof is not vacuous.}
The positive control is the reference fix verbatim. Its patch is byte-identical
to the reference diff of Box~\ref{box:11179-patch}, and it both verifies and
resolves. Alongside it the bundle carries a vacuity probe: the reference
implementation with an assertion of falsity placed on the returning path. That
probe must fail to verify, and it does. Without it, a specification whose
preconditions were unsatisfiable, or whose returning path was unreachable, would
verify every candidate and reject none, and the entire attack suite above would be
vacuous. Every prover bundle carries a positive control; this one also carries
the explicit vacuity probe.

\paragraph{Why the calibration has to be part of the protocol.}
Without the test-harness step, the attacker panel is a source of false findings
in one direction and false confidence in the other. An agent instructed to break
a specification will produce candidates that verify, because many correct
programs verify; grading those as weaknesses would make every well-written
specification look broken. Conversely, an agent whose candidate is rejected has
learned nothing about whether the rejection was justified. Running the candidate
against tests the task already ships resolves both cases against an authority
neither the specification nor the agent controls. \Cref{app:audit} reports how
often each outcome occurs across the corpus.

\paragraph{What the suite costs, and why it is retained.}
Driving nineteen candidates through the verifier for this one specification costs
minutes of prover time apiece, and writing them took an agent considerably
longer. The bundle therefore keeps them: each candidate is stored as source
alongside the verdict it must receive and the reason, so a later audit that
reconstructs the same attack reproduces the outcome without re-deriving it. The
twenty attacks this specification survived are stated, and new candidates can be
added to the suite.

\subsection{When faithfulness and non-disclosure collide}
\label{app:faithfulness}

The specification view shown to a solver has to satisfy two requirements that
usually sit comfortably together. It must be faithful: strong enough that
satisfying it entails the behavior the hidden tests check. And it must not
disclose the fix: it may reveal nothing about the change beyond what the public
problem statement already says. For a minority of instances these requirements
are jointly unsatisfiable. \Cref{app:leakage} states the disposition we adopt and
reports how many instances it applies to; this subsection works through one of
them in detail.

\paragraph{The task.}
\mbox{\id{scikit-learn-26194}} concerns the thresholds returned alongside a
receiver-operating-characteristic curve. When the classifier's scores are
probability estimates, the routine's leading threshold could exceed $1$, which is
not a value any probability can take, and the public issue reports this as a bug.
The issue then tentatively suggests clipping the thresholds so that none exceeds
$1$.
Box~\ref{box:26194-collision} shows the public text alongside the change the
maintainers actually made and the assertions the hidden tests make.

\begin{patchbox}[label=box:26194-collision]{\id{scikit-learn-26194}: what the public text asks for, what the fix does, and what the tests check}
\footnotesize
\textbf{From the public problem statement (paraphrased).} The first threshold returned can be
above $1$ when the scores are probability estimates, which is not a meaningful
probability. The suggested fix is to clip the thresholds to a maximum of $1$.
\smallskip

\textbf{The reference fix.}
\begin{minted}[escapeinside=]{diff}
-    thresholds = np.r_[thresholds[0] + 1, thresholds]
+    thresholds = np.r_[np.inf, thresholds]
\end{minted}
\smallskip

\textbf{What the hidden tests assert.} That the returned thresholds are exactly
\texttt{[inf, 1.0, 0.7, 0.0]} and \texttt{[inf, 1.0, 0.9, 0.7, 0.6, 0.0]} on the
two fixtures, and that the leading threshold satisfies \texttt{isinf}.
\end{patchbox}

The fix does not clip. It replaces the leading sentinel, previously the largest
observed threshold plus one, with positive infinity. The reasoning is that the
leading entry is not a real threshold at all but the degenerate point at which
nothing is predicted positive, and that infinity is the honest name for it. The
tests assert that value literally: two of them compare the whole threshold array
against a list beginning with infinity, and one checks the leading entry with an
infinity predicate.

\paragraph{Why the two requirements cannot both hold.}
A faithful specification has to distinguish the fixed behavior from the behavior
the tests reject. The behavior the tests reject includes the clipping the issue
itself proposes: clipping produces a leading threshold of $1.0$, which fails the
infinity predicate and fails both array comparisons. So a specification that pins
only ``the leading threshold is at most $1$'' (everything the public text
supports) is satisfied by an implementation the task fails. To be faithful, the
specification must name the infinite sentinel.

But the discriminator is absent from the public text. The tokens
\texttt{inf}, \texttt{isinf}, and \texttt{np.inf} do not occur in the problem
statement, and this instance ships no developer discussion at all, so there is no
other public source for them. What the public text does contain is
\texttt{clip}, and a suggested non-regression test asserting that every returned
threshold is at most $1$ or at least $0$. Every real number satisfies that
predicate, and the predicate mentions no sentinel at all. Naming the sentinel in
the specification view therefore tells a solver what the leading value changes
to, which is the entire content of the fix. The public text omits that value and
points away from it.

\begin{specbox}[label=box:26194-spec]{\id{scikit-learn-26194}: the discriminating clause, under contract-annotated Python}
\begin{minted}{python}
@Pure
@ContractOnly
def is_pos_inf(x: float) -> bool: pass

@Pure
@ContractOnly
def pos_inf() -> float:
    Ensures(is_pos_inf(Result()))

def anchor_thresholds(thresholds: List[float]) -> List[float]:
    Requires(Acc(list_pred(thresholds), 1 / 2))
    Requires(len(thresholds) >= 1)
    Requires(MustTerminate(2))
    Ensures(Acc(list_pred(thresholds), 1 / 2))
    Ensures(Acc(list_pred(Result())))
    Ensures(len(Result()) == len(thresholds) + 1)
    Ensures(is_pos_inf(Result()[0]))
    Ensures(ToSeq(Result()).drop(1) == ToSeq(thresholds))
\end{minted}
\end{specbox}

Box~\ref{box:26194-spec} is the specification as released. The clause
\texttt{is\_pos\_inf(Result()[0])} is the discriminator, and it is unavoidable:
an implementation satisfying the length and tail clauses while prepending any
finite value verifies and fails the task. There is no weaker clause that
separates the two, because the tests are stated as an equality against a specific
value. A specification that says ``the leading value is greater than every
threshold in the input'' would be non-disclosing and would still be satisfied by
the pre-fix code, which prepends the maximum plus one. A specification that says
``the leading value is not finite'' has already said everything the fix says.

\paragraph{The four ways out, and why we take the third.}
The collision admits exactly four responses, and the choice among them affects
how the corpus should be read.

Weaken the specification and keep non-disclosure: the instance is retained, but
the guarantee is void, since a verified submission need not resolve the task.
This is the one option we rule out unconditionally: it breaks the property the
benchmark exists to provide.

Strengthen the public text to mention the sentinel: this would restore
consistency, at the cost of editing the task. We do not modify the problem
statements we inherit, so this is unavailable.

Keep the instance, permit the discriminator in the specification view, and record
the relaxation per instance. This is what we do, and it is the only option that
preserves both the guarantee and the task.

Discard the instance. The instances that exhibit the collision are not a random
sample. They are precisely the tasks where the public issue report is a poor
description of the fix that was made: the reporter proposed one remedy and the
maintainers chose another. Removing them would systematically strip the corpus of
the cases in which the natural-language statement underdetermines the answer, and
that underdetermination is what distinguishes a real repository issue from a
competition problem.

\paragraph{What is relaxed, and what is not.}
The relaxation is local to one property. For this instance every other admission
check holds in its ordinary form. The specification verifies in $13.5$ seconds,
under an interpreted encoding for real arithmetic. This bundle is one of the $11$
\nagini bundles that need one, since the instance is about the values of
floating-point thresholds. The pre-fix twin fails, and it fails on the
discriminating clause itself, reporting that the finite anchoring value is not the
infinite sentinel; the two specifications are byte-identical outside the
implementation region, so the twin isolates that one substitution and nothing
else. The differential harness compares the modeled reference against the real
patched routine on $120{,}012$ generated inputs with no disagreement. All seven
mutants of the reference body are killed. The property test suite's eight cases each produce
their required verdict. Seven adversarial candidates were submitted against the
specification, and none broke it. The two numeric primitives the specification
assumes (the infinity sentinel and the test for it) are probed against the real
library, on boundary values, the test fixture, and five thousand generated
inputs. What is relaxed is only the non-disclosure screen, and the record states
which tokens it would otherwise have flagged.

An instance qualifies for the relaxation only if an independent adversarial
analysis, run with the public text and the repository in hand, fails to find any
reformulation derivable from that text which pins the tested behavior. For this
instance the analysis reconstructed the mismatch from both sides: it confirmed
that the discriminating vocabulary occurs in the reference fix and the hidden test
and nowhere in the solver-visible text, and it confirmed that the remedy the
public text does propose is incompatible with the behavior the tests require. The
mechanical screen, which is lexical, passes this instance: the specification view
contains no diff line, no patched path, and no patched symbol. The relaxation is
recorded against the semantic collision, which no lexical screen can see.

\paragraph{Reading the corpus in light of this.}
Two consequences follow, and both are stated in the released records. First, the
affected instances are marked, so anyone who wants a corpus in which no
specification view exceeds the public text can exclude them and lose the counts
reported in \cref{app:leakage}. Second, the settings that supply a specification
to a solver are, on those instances, supplying more help than the problem
statement contains. The specification-supplied rows of \cref{app:results} should
therefore be read as an upper bound on that setting's value. That is also why the
per-stratum analysis there separates instances by how much the public text
already determines.

\section{Representative evaluation episodes}
\label{app:episodes}

The results in \cref{app:results} report how often a submission clears each
grader. They do not show what an episode looks like from the inside, and they do
not show a submission that clears one grader and fails another with both
verdicts in view. This section prints three episodes in full, quoting the
recorded interaction.

The three were chosen to occupy three different cells of the outcome table, and
two of them deliberately share an instance. \Cref{app:episode-e2e} is an episode
in the Verified End-to-End setting that clears every grader: the agent writes a
patch, writes a specification, is rejected twice by the verifier, repairs, and is
accepted by the equivalence panel. \Cref{app:episode-unresolved} is an episode in
the same setting whose specification verifies on the first attempt and whose
patch does not resolve. That quadrant exists only because the two deliverables
are graded separately, and it shows where a formal artifact can be honest and
still miss. \Cref{app:episode-synth} is a Specification Synthesis episode on the
same instance as \cref{app:episode-unresolved}, in a different run and a
different setting. Its specification verifies and is then rejected unanimously by
the audit panel on faithfulness, for exactly the reason the previous episode's
patch failed its hidden test. The closing discussion of this appendix draws the three
together.

All three are Opus~4.8 episodes on the \nagini{} backend from the first
repetition, chosen so that the verifier output, the specification language, and
the model are constant across the section and only the setting and the outcome
vary. The two that share an instance are independent: they come from different
runs, in different settings, and neither agent could see the other's work.

\paragraph{What an episode record contains.}
Every episode is recorded in five parts.

The \emph{task view} is exactly the material the agent was given: the standing
instruction that describes the environment and the ground rules, the setting
instruction that states what is to be produced and in what order, the issue
text, and the localization hint or the specification view in the settings that
supply them. The view is recorded as the agent saw it rather than reconstructed
afterwards. \Cref{app:modes} describes what each setting supplies.

The \emph{interaction log} is an ordered list of events. Each model turn carries
its visible text and the action it requested. Each shell call carries the command
issued, its exit status, and the output the agent saw. Each verifier call carries
the artifact submitted, the verdict, the verifier's own message, and the wall
time the call took. The log ends with the submission event and its note.

The \emph{submission} is the patch, captured as a diff of the agent's edits
against the base commit, together with the formal artifact in the settings that
require one. Nothing else the agent did to its container is carried forward.

The \emph{mechanical grades} record resolution under the official harness,
verification under the backend's own verifier, and the hygiene screen, each as a
verdict with a short reason.

The \emph{adjudications} record every agentic verdict at the level of the
individual ballot: for equivalence, each judge's decision, confidence, and stated
reason; for the specification audit, each judge's five per-property booleans, its
reason, and the list of tool calls it made with the signal each returned. The
anti-cheat screen's result for the episode is recorded alongside.

Counters derived from the log are stored with the record: model turns, shell
calls, verifier calls and accepting verdicts, verifier wall time, episode wall
time, and token consumption. \Cref{tab:records} and \cref{app:cost} aggregate
them.

\paragraph{Transcript conventions.}
In this section a gray box is material given to the agent, a blue box is the
agent's own turn, a tan box is the output of a shell call, a red box is a verdict
returned by the verifier or by a grader. Everything inside a box is quoted from the record. Elisions are marked
\texttt{[...]}; nothing is reworded. The standing instruction is cut throughout: it is identical across the episodes of a setting and is described in
\cref{app:harness} rather than reprinted. Long test output is shown as the agent
saw it, which in these episodes means the last lines of the run, because the
agent piped it that way.

\begin{table}[t]
\centering
\caption{The three episodes. Wall time and token counts are the recorded totals
for the episode; ``verify'' gives the number of verifier calls, the number that
returned an accepting verdict, and the wall time spent inside the verifier. The
grader columns give the outcome of each grader that applies to the setting;
\na{} means the grader does not run in that setting, and ``skipped'' means it
was not reached because an earlier grader had already failed.}
\label{tab:episodes}
\small
\setlength{\tabcolsep}{4.5pt}
\begin{tabular}{@{}llll@{}}
\toprule
& \hd{\cref{app:episode-e2e}} & \hd{\cref{app:episode-unresolved}}
  & \hd{\cref{app:episode-synth}} \\
\midrule
Instance & \id{django-11179} & \id{django-16667} & \id{django-16667} \\
Setting & 3 Verified End-to-End & 3 Verified End-to-End
  & 8 Specification Synthesis \\
Run & first repetition & first repetition & first repetition \\
Model / backend & Opus 4.8 / \nagini & Opus 4.8 / \nagini & Opus 4.8 / \nagini \\
\midrule
Model turns & $10$ & $7$ & $7$ \\
Wall time & $300.8$\,s & $259.7$\,s & $505.0$\,s \\
Verify calls / accepted & $3$ / $1$ & $1$ / $1$ & $2$ / $1$ \\
Verifier wall time & $70.1$\,s & $27.0$\,s & $25.5$\,s \\
Input / output tokens & $53{,}211$ / $1{,}681$ & $36{,}822$ / $1{,}689$
  & $38{,}619$ / $2{,}217$ \\
\midrule
Resolution & pass & \hd{fail} & \na \\
Verification & pass & pass & pass \\
Anti-fakery & pass & pass & pass \\
Equivalence & pass ($3$/$3$) & skipped & \na \\
Specification audit & \na & \na & \hd{fail ($0$/$3$)} \\
Hygiene screen & clean & clean & clean \\
\midrule
\hd{Episode outcome} & \hd{pass} & \hd{fail} & \hd{fail} \\
\bottomrule
\end{tabular}
\end{table}

\subsection{A verified end-to-end episode}
\label{app:episode-e2e}

This episode runs on \mbox{\id{django-11179}}, the instance whose four corpus
specifications are printed side by side in \cref{app:four-languages}. The agent
here has seen none of them: in the Verified End-to-End setting both the patch and
the specification are constructed from the issue and the code.

\begin{trajprompt}[title={Setting instruction, row 3 (excerpt)}]
\begin{trajtext}
Your task has TWO graded deliverables you construct from scratch (no spec is
given): a Nagini MODEL and a real-code PATCH. You are scored on ALL of: (i) the
model VERIFIES; (ii) the patch RESOLVES the hidden tests; and (iii) an
independent audit confirming your model's specification faithfully represents
the issue's intent AND is logically equivalent to your patch. [...]
  (1) LOCATE THE REAL FIX FIRST. [...] confirmed by observing the wrong
      behavior there, not guessed from the issue title.
  (2) CONSTRUCT a Nagini specification of the CORRECTED behavior at those
      site(s) [...] It does not need to be an exhaustive or airtight contract;
      it needs to genuinely represent the issue's intent rather than a vacuous
      or trivially-true property.
  (3) Write a model implementing that spec and iterate with `nagini_verify`
      until it VERIFIES [...]
  (4) DERIVE your /testbed edit as a faithful translation of the verified model
      (case for case), at the real site(s) from step 1 [...]
  (5) VALIDATE END-TO-END. [...] Then `submit` with model_path set to your
      verified model.

--- ISSUE (problem statement) ---
delete() on instances of models without any dependencies doesn't clear PKs.
Description

Deleting any model with no dependencies not updates the PK on the model. It
should be set to None after .delete() call.
See Django.db.models.deletion:276-281. Should update the model line 280.
\end{trajtext}
\end{trajprompt}

The issue names a file and a line range, which is unusually specific for this
corpus, and the agent's first action is to read them.

\begin{trajobs}[title={Step 1: shell (output abridged)}]
\begin{trajtext}
$ sed -n '260,330p' /testbed/django/db/models/deletion.py

    def delete(self):
        [...]
        # Optimize for the case with a single obj and no dependencies
        if len(self.data) == 1 and len(instances) == 1:
            instance = list(instances)[0]
            if self.can_fast_delete(instance):
                with transaction.mark_for_rollback_on_error():
                    count = sql.DeleteQuery(model).delete_batch([instance.pk], self.using)
                return count, {model._meta.label: count}
        [...]
        for model, instances in self.data.items():
            for instance in instances:
                setattr(instance, model._meta.pk.attname, None)
        return sum(deleted_counter.values()), dict(deleted_counter)
[exit 0]
\end{trajtext}
\end{trajobs}

One read is enough. The two branches are visible in the same window: the fast
path returns before the loop that clears identifiers, and the general path runs
it. The agent states the diagnosis in a single sentence and edits.

\begin{trajagent}[title={Step 2: agent}]
\begin{trajtext}
The fast-delete path returns early without setting the PK to None. Let me fix
it.
\end{trajtext}
\end{trajagent}

The edit, applied through a scripted textual replacement, inserts one statement
before the fast path's return:

\begin{trajobs}[title={Step 2: the resulting change}]
\begin{trajtext}
             if self.can_fast_delete(instance):
                 with transaction.mark_for_rollback_on_error():
                     count = sql.DeleteQuery(model).delete_batch([instance.pk], self.using)
+                setattr(instance, model._meta.pk.attname, None)
                 return count, {model._meta.label: count}
\end{trajtext}
\end{trajobs}

This is byte-for-byte the reference fix. The remainder of the episode is about
the other deliverable.

\begin{trajagent}[title={Step 3: agent writes its specification}]
\begin{trajtext}
Now the Nagini model.

from typing import Optional
from nagini_contracts.contracts import *


class Instance:
    def __init__(self) -> None:
        self.pk = 1  # type: Optional[int]
        Ensures(Acc(self.pk) and self.pk == 1)


@ContractOnly
def delete_batch(pk: Optional[int]) -> int:
    Ensures(Result() >= 0)


def fast_delete(instance: Instance) -> int:
    Requires(Acc(instance.pk))
    # After delete on a no-dependency instance, the PK must be cleared.
    Ensures(Acc(instance.pk))
    Ensures(instance.pk is None)
    Ensures(Result() >= 0)
    count = delete_batch(instance.pk)
    instance.pk = None
    return count
\end{trajtext}
\end{trajagent}

The removal itself is an unchanged callee and is axiomatized as one, with the
weakest fact the proof needs: it returns a count that is not negative. The
obligation the issue asks for is stated directly as a postcondition on the
changed operation: after it returns, the identifier is cleared. The class exists
only to give the identifier somewhere to live.

\begin{trajverdict}[title={Step 4: verifier, $23.5$\,s}]
\begin{trajtext}
Verification failed
Errors:
cand_model.py:8:34:8:46: error: The precondition of (self.pk == 1) might not hold.
Verification took 21.67 seconds.
\end{trajtext}
\end{trajverdict}

This rejection is not about deletion. It is about permission discipline: reading
a field inside a contract requires holding access to that field, and in the
constructor's postcondition the conjunction asserts the permission and consumes
it in the same expression. The agent splits the conjunction into two clauses.

\begin{trajagent}[title={Step 5: agent (the changed lines)}]
\begin{trajtext}
        Ensures(Acc(self.pk))
        Ensures(self.pk == 1)
\end{trajtext}
\end{trajagent}

\begin{trajverdict}[title={Step 6: verifier, $24.7$\,s}]
\begin{trajtext}
Verification failed
Errors:
cand_model.py:9:17:9:29: error: The precondition of (self.pk == 1) might not hold.
Verification took 22.87 seconds.
\end{trajtext}
\end{trajverdict}

The same rejection, one line lower. Splitting the clauses does not help, because
each clause is checked on its own and the second one still reads a field whose
permission the first one asserts. The agent then gives up part of its claim.

\begin{trajagent}[title={Step 7: agent}]
\begin{trajtext}
The constructor postcondition referencing self.pk needs Acc in same Ensures.
Let me simplify—remove the constructor entirely since it's not needed.
\end{trajtext}
\end{trajagent}

What it actually writes keeps the constructor but drops the claim about the
identifier's initial value, leaving only the permission:

\begin{trajagent}[title={Step 7: the changed constructor}]
\begin{trajtext}
class Instance:
    def __init__(self) -> None:
        Ensures(Acc(self.pk))
        self.pk = 1  # type: Optional[int]
\end{trajtext}
\end{trajagent}

\begin{trajverdict}[title={Step 8: verifier, $21.9$\,s}]
\begin{trajtext}
Verification successful
Verification took 20.33 seconds.
\end{trajtext}
\end{trajverdict}

The claim that was dropped is not part of the obligation. Nothing in the issue
depends on what the identifier held before the deletion; the obligation is that
afterwards it holds nothing. The agent's simplification drops a claim the issue
does not need and keeps the one it does. That is the distinction the
anti-fakery screen enforces, and the screen accepts this artifact.

The last two actions are validation and submission.

\begin{trajobs}[title={Step 9: shell}]
\begin{trajtext}
$ python tests/runtests.py delete --parallel 1
Creating test database for alias 'default'...
..s......................................
----------------------------------------------------------------------
Ran 41 tests in 0.831s

OK (skipped=1)
[exit 0]
\end{trajtext}
\end{trajobs}

\begin{trajagent}[title={Step 10: agent submits}]
\begin{trajtext}
Tests pass. The fix sets the PK to None on the fast-delete path.

submit(note="Fast-delete path in Collector.delete now clears the instance PK
(setattr to None) before returning, matching the slow path behavior.")
\end{trajtext}
\end{trajagent}

Grading is then mechanical up to the last step. The patch is re-applied to a
clean checkout inside the official image and resolves. The specification and
implementation are re-checked by the verifier on the grading host and verify. The
hygiene screen passes. Only then does the equivalence panel run.

\begin{trajverdict}[title={Grading: equivalence panel, $3$ accept / $0$ reject}]
\begin{trajtext}
vote 1 (confidence 0.95): Spec clause 'instance.pk is None' matches the issue's
intent that PK be cleared after delete; model's setting pk=None after
delete_batch mirrors patch's setattr(instance, model._meta.pk.attname, None)
after delete_batch.

vote 2 (confidence 0.95): Spec's Ensures(instance.pk is None) faithfully
captures the issue's intent that PK be cleared after delete; model's
delete_batch->set pk=None->return count mirrors the patch's
delete_batch->setattr(None)->return count.

vote 3 (confidence 0.95): Spec's Ensures(instance.pk is None) captures the
issue's intent of clearing PK after delete; model's delete_batch+pk=None
mirrors patch's delete_batch+setattr(pk,None).
\end{trajtext}
\end{trajverdict}

\paragraph{What the formal channel contributed.}
Three verifier calls, $70.1$ seconds of the episode's $300.8$, and every one of
them was about the specification language rather than about Django. The first two
rejections concern where a permission may be stated; the third call accepts. Not
one of them said anything the agent could have used to locate or improve the fix,
because the fix was already correct at step 2. In this episode the verifier
functioned as a proof checker. The episode passes because the agent's reading of
the issue was right before it started specifying.

\paragraph{What the specification does and does not say.}
Set beside the corpus specification for the same instance in
\cref{app:four-languages}, this one is very small. It says nothing about the
count reported to the caller, nothing about the general path, and nothing about
the store the removal talks to; the corpus specification prices all three,
because it has to survive twenty adversarial candidates rather than one panel.
The agent's specification is nonetheless not vacuous: the pre-fix behavior
violates it, which is the property the corpus construction requires of a
specification and the property the anti-fakery screen checks a submission for.

\subsection{A verified submission that does not resolve}
\label{app:episode-unresolved}

The next episode is in the same setting, with the same model and backend, and its
formal artifact verifies on the first attempt. It fails, because the patch does
not resolve. This is the outcome that the two-deliverable design exists to
expose; a resolution-only benchmark cannot see it.

\begin{trajprompt}[title={Issue (excerpt)}]
\begin{trajtext}
SelectDateWidget can crash with OverflowError.
Description

Given a relatively common view like this: [...]
A user can trigger a server crash, reproducible by running locally and visiting
http://127.0.0.1:8000/repro/?my_date_day=1&my_date_month=1&my_date_year=1234567821345678,
which results in
[...]
 File "[...]/django/forms/widgets.py", line 1160, in value_from_datadict
    date_value = datetime.date(int(y), int(m), int(d))
OverflowError: signed integer is greater than maximum
[...]
Specifically, y, m and d are user controlled, and the code does this:
 date_value = datetime.date(int(y), int(m), int(d))
When large integers (larger than sys.maxsize) are supplied to date's
constructor it will throw an OverflowError [...]
\end{trajtext}
\end{trajprompt}

The agent reads the function and finds a guarded region that already handles one
failure mode.

\begin{trajobs}[title={Step 1: shell (output abridged)}]
\begin{trajtext}
$ sed -n '1140,1175p' django/forms/widgets.py

    def value_from_datadict(self, data, files, name):
        [...]
        if y is not None and m is not None and d is not None:
            input_format = get_format("DATE_INPUT_FORMATS")[0]
            input_format = formats.sanitize_strftime_format(input_format)
            try:
                date_value = datetime.date(int(y), int(m), int(d))
            except ValueError:
                # Return pseudo-ISO dates with zeros for any unselected values,
                # e.g. '2017-0-23'.
                return "%s-%s-%s" % (y or 0, m or 0, d or 0)
            return date_value.strftime(input_format)
        return data.get(name)
[exit 0]
\end{trajtext}
\end{trajobs}

\begin{trajagent}[title={Step 2: agent}]
\begin{trajtext}
The fix is to also catch `OverflowError`. Let me make the edit.
\end{trajtext}
\end{trajagent}

\begin{trajobs}[title={Step 2: the resulting change}]
\begin{trajtext}
             try:
                 date_value = datetime.date(int(y), int(m), int(d))
-            except ValueError:
+            except (ValueError, OverflowError):
                 # Return pseudo-ISO dates with zeros for any unselected values,
                 # e.g. '2017-0-23'.
                 return "%s-%s-%s" % (y or 0, m or 0, d or 0)
\end{trajtext}
\end{trajobs}

The specification follows, and it follows the patch closely.

\begin{trajagent}[title={Step 3: agent writes its specification (comments abridged)}]
\begin{trajtext}
# Abstracts datetime.date(y, m, d): [...] We model both error conditions as a
# single "raises" flag.
@Pure
@ContractOnly
def date_ok(y: int, m: int, d: int) -> bool:
    # True iff datetime.date(y, m, d) succeeds (no ValueError, no OverflowError).
    ...


@Pure
@ContractOnly
def date_strftime(y: int, m: int, d: int) -> int:
    Requires(date_ok(y, m, d))
    Ensures(Result() >= 0)
    ...


# Result encoding:
#   tag 0 -> formatted valid date (value = date_strftime(...))
#   tag 1 -> pseudo-ISO fallback string (value unused), used whenever the
#            date constructor would raise (ValueError OR OverflowError).
# The KEY behavior of the fix: an out-of-range/too-large value must NOT
# propagate an exception; it must fall into the fallback (tag 1).
def value_from_datadict(y: int, m: int, d: int) -> Tuple[int, int]:
    Ensures(date_ok(y, m, d) == (Result()[0] == 0))
    Ensures(Implies(not date_ok(y, m, d), Result()[0] == 1))
    Ensures(Implies(Result()[0] == 0, Result()[1] == date_strftime(y, m, d)))
    if date_ok(y, m, d):
        return (0, date_strftime(y, m, d))
    else:
        # ValueError or OverflowError both handled -> pseudo-ISO fallback.
        return (1, 0)
\end{trajtext}
\end{trajagent}

\begin{trajverdict}[title={Step 4: verifier, $27.0$\,s}]
\begin{trajtext}
Verification successful
Verification took 25.10 seconds.
\end{trajtext}
\end{trajverdict}

The agent then validates, and validates carefully: it runs the widget's own test
module, and it reproduces the crash from the issue.

\begin{trajobs}[title={Steps 5 and 6: shell}]
\begin{trajtext}
$ python tests/runtests.py forms_tests.widget_tests.test_selectdatewidget
Found 15 test(s).
...............
----------------------------------------------------------------------
Ran 15 tests in 0.145s

OK
[exit 0]

$ python - <<EOF
[...]
w = SelectDateWidget()
print(repr(w.value_from_datadict(
    {'d_day':'1','d_month':'1','d_year':'1234567821345678'}, {}, 'd')))
EOF
'1234567821345678-1-1'
[exit 0]
\end{trajtext}
\end{trajobs}

\begin{trajagent}[title={Step 7: agent submits}]
\begin{trajtext}
No more crash.

Let me verify the model and patch align case for case: `date_ok` true -> tag 0
-> strftime; false (ValueError or OverflowError) -> tag 1 -> fallback. The patch
catches both `ValueError` and `OverflowError`, returning the pseudo-ISO
fallback. Matches.
\end{trajtext}
\end{trajagent}

The self-check in that turn is correct on its own terms. The patch and the
specification do agree, case for case. The reproduction did stop crashing.

\begin{trajverdict}[title={Grading}]
\begin{trajtext}
resolution   : FAIL   "patch did not resolve"
verification : pass   "verified"
anti-fakery  : pass   "model is an honest Nagini contract"
equivalence  : skipped ("verify/resolve/anti-fakery not all green")
hygiene      : clean  (0 flags)
\end{trajtext}
\end{trajverdict}

\paragraph{What the reference fix does instead.}
The gold patch does not widen the existing handler. It adds a second, separate
one, which returns a different string.

\begin{patchbox}[label=box:16667-gold]{The reference fix for \id{django-16667}}
\begin{minted}[escapeinside=]{diff}
                 # Return pseudo-ISO dates with zeros for any unselected values,
                 # e.g. '2017-0-23'.
                 return "%s-%s-%s" % (y or 0, m or 0, d or 0)
+            except OverflowError:
+                return "0-0-0"
             return date_value.strftime(input_format)
\end{minted}
\end{patchbox}

Two of the held-out tests turn on that distinction. One asserts that an
oversized year produces exactly the string \verb|"0-0-0"|; the other requires
that the same string, fed back through the field, is rejected as invalid, so
that the form reports an error rather than accepting a value it cannot represent.
Widening the existing handler makes the crash stop and produces
\verb|'1234567821345678-1-1'| instead. That is a pseudo-date that echoes the
attacker-supplied year and that the field will try to parse. The agent printed
that string, read it as evidence that the crash was gone, and submitted. It was
also the evidence that the fix was wrong.

\paragraph{Why the specification could not have caught it.}
The specification is not vacuous. Its abstraction of the date constructor is
two-valued: the components are accepted, or they are not. The pre-fix behavior,
which lets the rejection escape, does not satisfy it. It would pass the
discrimination requirement that every corpus specification has to meet.
But because the abstraction has two values where the required behavior has
three, both patches satisfy it: the reference fix and the agent's widened handler
map to the same tag on an oversized year, and the returned text is not modeled
at all. The information the hidden test checks was discarded when the abstraction
was chosen, before a single line of proof was written.

The corpus specification for the same instance distinguishes the two rejections
and pins the sentinel exactly.

\begin{specbox}[label=box:16667-corpus]{The corpus specification for \id{django-16667} (postconditions, \nagini{})}
\begin{minted}[escapeinside=]{python}
def value_from_components(all_blank: bool, all_present: bool,
                          kind: int) -> List[int]:
    Requires(Implies(all_blank, all_present))
    Requires(kind == 0 or kind == 1 or kind == 2)
    Requires(MustTerminate(3))
    Ensures(Acc(list_pred(Result())))
    Ensures(Implies(all_blank, ToSeq(Result()) == absent_text()))
    Ensures(Implies(not all_present, ToSeq(Result()) == raw_value()))
    # every outcome returns text: NEITHER rejection escapes
    Ensures(Implies(all_present and not all_blank, len(ToSeq(Result())) > 0))
    # a capacity-overflow failure yields EXACTLY the "0-0-0" sentinel
    Ensures(Implies(all_present and not all_blank and kind == 2,
                    ToSeq(Result()) == zero_zero_zero()))
    Ensures(Implies(all_present and not all_blank and kind == 0,
                    ToSeq(Result()) == success_text()))
    Ensures(Implies(all_present and not all_blank and kind == 1,
                    ToSeq(Result()) == out_of_range_text()))
\end{minted}
\end{specbox}

The construction differs in three ways, all of them consequences of the same
decision. The outcome of the date constructor is three-valued, tagging the two
rejections separately. The result is text rather than a tag, modeled as a
sequence of code points, so a postcondition can name a particular string;
\verb|zero_zero_zero()| is the five code points of \verb|"0-0-0"|. And each
branch is pinned to a distinct value, so an implementation that returns the
out-of-range rendering on an overflow violates the contract. The corpus
specification's own summary calls this ``the discriminating obligation'', and it
is the obligation the held-out test checks.

\paragraph{Two procedural notes.}
The equivalence panel did not run. It is skipped whenever the mechanical graders
are not all green: a judgment that a specification and a patch describe the same
behavior is uninformative when the patch is known to be wrong. \Cref{app:scoring}
states the ordering rule. The hygiene screen is clean and the anti-fakery screen
accepts. Nothing about this submission is dishonest; it is a faithful proof of an
insufficient claim.

\paragraph{How common this is.}
In this run, $69$ of $500$ episodes verify and do not resolve, against $429$ that
do both; two resolve without a verifying artifact. Conditional on verification,
the resolution rate is $86.1\%$. Across the four repetitions of the cell the
conditional rate ranges from $83.4\%$ to $86.1\%$. \Cref{app:verif-effort}
reports the structure of this gap across settings and backends.

\subsection{A synthesized specification the audit rejects}
\label{app:episode-synth}

The last episode is on the same instance, in the Specification Synthesis setting,
in a different run. Here there is no patch to grade: the specification is the
deliverable, any repository edit is ignored, and the grade rests on the audit
panel. The setting instruction states three demands on the specification, and it
is the third that decides this episode.

\begin{trajprompt}[title={Setting instruction, row 8 (excerpt)}]
\begin{trajtext}
Your task: from the issue and the code, CONSTRUCT a formal Nagini SPECIFICATION
of the fix the issue requires, and a small implementation that VERIFIES under
it. You are graded on the quality of the SPECIFICATION — not on editing the
repository (any repo edits are ignored). [...]
  (2) WRITE THE SPECIFICATION as `Requires`/`Ensures` on the function(s) the fix
      concerns [...] Make it:
    - PERMISSIVE ENOUGH: the preconditions must admit every input the corrected
      code legitimately handles [...]
    - TRUE: every postcondition must be a correct statement about the intended
      behavior — never assert something the correct fix would violate.
    - STRONG ENOUGH: the postconditions must pin the behavior tightly enough
      that ONLY a correct implementation can satisfy them. [...] After writing
      it, ask yourself: could an incorrect implementation still verify against
      this? If yes, strengthen it.
  (3) For any UNCHANGED system or library function the behavior depends on, you
      may add a `@ContractOnly` axiom [...] Do NOT axiomatize the behavior the
      fix itself introduces [...]
\end{trajtext}
\end{trajprompt}

The agent reads the same function as before and writes a specification that goes
one step further than the previous episode's: its abstraction of the date
constructor is three-valued.

\begin{trajagent}[title={Step 2: agent writes its specification (comments abridged)}]
\begin{trajtext}
# We model the date constructor as an axiomatized callee that has three
# possible outcomes encoded as an int return:
#   0 -> success (a valid date)
#   1 -> ValueError (out of normal range)
#   2 -> OverflowError (integer too large)
# This is TRUE of real datetime.date: it never crashes other than via these.

YEAR_MIN = 1
YEAR_MAX = 9999

@Pure
@ContractOnly
def date_outcome(y: int, m: int, d: int) -> int:
    # 0 success, 1 ValueError, 2 OverflowError
    Ensures(Result() >= 0 and Result() <= 2)
    # A genuinely valid date (in-range parts and within C long) succeeds.
    Ensures(Implies(y >= YEAR_MIN and y <= YEAR_MAX and m >= 1 and m <= 12
                    and d >= 1 and d <= 28, Result() == 0))
    # Out-of-C-long-range years overflow.
    Ensures(Implies(y > YEAR_MAX, Result() != 0))


RESULT_OK = 0
RESULT_FALLBACK = 1


def value_from_datadict_branch(y: int, m: int, d: int) -> int:
    # POSTCONDITION pins the fix: it NEVER propagates an error; for any input
    # that the date constructor rejects (ValueError OR OverflowError) it must
    # return the fallback; only a genuinely valid date returns OK.
    Ensures(Result() == RESULT_OK or Result() == RESULT_FALLBACK)
    Ensures(Result() == RESULT_OK) == (date_outcome(y, m, d) == 0)
    Ensures(Result() == RESULT_FALLBACK) == (date_outcome(y, m, d) != 0)
    oc = date_outcome(y, m, d)
    if oc == 0:
        return RESULT_OK
    else:
        # both ValueError (oc==1) and OverflowError (oc==2) hit this branch
        return RESULT_FALLBACK
\end{trajtext}
\end{trajagent}

The distinction the axiom draws does not survive into the postconditions: the
result is again two-valued, and the two rejections are mapped to one outcome by
the implementation. Before the panel gets to say so, the verifier rejects the
artifact for a different reason.

\begin{trajverdict}[title={Step 3: verifier, $9.1$\,s}]
\begin{trajtext}
Translation failed
Not supported:
(Ensures((Result() == RESULT_OK)) == (date_outcome(y, m, d) == 0))
\end{trajtext}
\end{trajverdict}

This is a language-level refusal rather than a failed proof. In the two lines it
names, the equality sits outside the postcondition instead of inside it, so the
text does not say what its author meant it to say. The backend declines to
translate the construction. The agent recognizes the problem immediately.

\begin{trajagent}[title={Step 4: agent}]
\begin{trajtext}
The `Ensures(...) == (...)` syntax is the issue. Let me fix it.

    Ensures((Result() == RESULT_OK) == (date_outcome(y, m, d) == 0))
    Ensures((Result() == RESULT_FALLBACK) == (date_outcome(y, m, d) != 0))
\end{trajtext}
\end{trajagent}

\begin{trajverdict}[title={Step 5: verifier, $16.4$\,s}]
\begin{trajtext}
Verification successful
Verification took 15.24 seconds.
\end{trajtext}
\end{trajverdict}

The agent then applies the repository edit as well, unprompted, using the same
widened handler as the row-3 episode. It is the same misreading of the fix, by
the same model, in a different setting and a different run. The setting ignores
the edit, so it does not affect the grade, but the repetition is evidence that
the reading is systematic rather than a sampling accident.

\begin{trajagent}[title={Steps 6 and 7: agent submits}]
\begin{trajtext}
Verified. The spec captures the fix: the OverflowError case (`oc==2`, which the
buggy code does not catch) must return the fallback rather than propagate. Let
me also apply the actual repo fix for completeness.

$ sed -i 's/            except ValueError:/            except (ValueError, OverflowError):/' \
    django/forms/widgets.py

submit(note="Spec for SelectDateWidget.value_from_datadict: the
date-construction branch must catch both ValueError AND OverflowError,
returning the pseudo-ISO fallback rather than propagating. [...]")
\end{trajtext}
\end{trajagent}

\paragraph{The audit panel.}
Three judges vote on five properties each. All three accept four of the five and
reject the same one. Every ballot ran the verifier on a candidate implementation
and probed the axiom against the real callee inside the instance's own image.

\begin{trajverdict}[title={Grading: specification audit, $0$ accept / $3$ reject}]
\begin{trajtext}
per-property rejections (of 3 decisive ballots):
  axiom soundness 0    admissibility 0    soundness 0    completeness 0
  faithfulness    3
tool grounding: 3/3 ballots ran the verifier; 3/3 ran an axiom probe

ballot 2 (confidence 0.85): The probe confirmed date_outcome is sound against
the real datetime.date (unchanged callee, not a wrap), the correct impl verifies
(sound), and the pre-fix buggy impl that drops the OverflowError fallback fails
to verify (complete). But faithful FAILS: the candidate models the output only
as an opaque int code and collapses ValueError and OverflowError into one
RESULT_FALLBACK, never modeling the returned text — so it cannot pin the gold's
discriminating obligation that an OverflowError yields EXACTLY the '0-0-0'
sentinel (vs the pseudo-ISO fallback for ValueError) that
test_value_from_datadict asserts. An impl returning the wrong fallback string on
overflow maps to the same RESULT_FALLBACK and verifies, so verify does not imply
correct on the behavior the issue requires.

ballot 1 (confidence 0.85): [...] the branch method imposes no Requires ->
admissible; the correct impl (both reject-kinds -> fallback) verified -> sound,
and the buggy port (overflow treated as success) failed the 'RESULT_OK iff
outcome==0' postcondition -> complete. Faithful is FALSE: the candidate
collapses ValueError (kind 1) and OverflowError (kind 2) into a single abstract
RESULT_FALLBACK and models NO output string, leaving the gold's central
discriminating obligation [...] entirely unconstrained [...]

ballot 3 (confidence 0.85): [...] FAITHFUL fails: the gold pins a discriminating
obligation that OverflowError(kind 2) yields EXACTLY the '0-0-0' sentinel while
ValueError(kind 1) yields the distinct pseudo-ISO '%s-%s-%s' fallback — both
checked by test_value_from_datadict — but the candidate abstracts away all
output strings and collapses kinds 1 and 2 into one indistinguishable
RESULT_FALLBACK [...]
\end{trajtext}
\end{trajverdict}

\paragraph{What the machine settled and what it did not.}
Soundness and completeness were decided by running the verifier: each judge
ported a correct implementation into the candidate's own signatures and
confirmed it verifies, then ported the pre-fix behavior the same way and
confirmed it does not. Axiom soundness was decided by executing a probe against
the real date constructor in the instance's image. Four of the five properties
therefore rest on machine evidence, and all four hold.

The property that fails is the one the tools used here do not decide, because it is not a
question about the specification's internal consistency. It asks whether the
abstraction is the right abstraction: whether what the specification chose to
model includes what the issue requires. Here it does not, and all three judges
locate the omission in the same place: the returned text is not modeled, so the
two rejections are indistinguishable, so an implementation that returns the wrong
string on an overflow satisfies the contract.

\paragraph{On the panel's privilege.}
The judges see more than the agent did in \cref{app:episode-unresolved}. An audit
runs after the episode has ended, and the judges are shown the issue, the
reference fix, the names of the held-out tests, and the corpus's own artifacts
for the instance. The instruction accompanying that material states that these
are references and not equivalence targets: a specification that differs from
ours but carries the guarantee passes. The ballots quoted above use that
material: they name the sentinel and the test. Given the reference behavior, the
panel identifies an abstraction gap in a specification the verifier had already
accepted, and attributes the failure to the right property.
\Cref{tab:synth} reports the audit per property, and \cref{app:refutation}
calibrates the panel against independent re-adjudication.

\paragraph{How common this is.}
Faithfulness is the bottleneck of this setting. In this run the specification
passes the full audit in $49.2\%$ of episodes, and the five properties fail at
$3.0$, $5.2$, $11.2$, $4.8$, and $48.6$ percent respectively. Faithfulness fails
an order of magnitude more often than axiom soundness. Restricting to the
cleanest form of the verdict, $81$ of the $500$ episodes are rejected
unanimously, on faithfulness alone, with every other property accepted by every
judge; this episode is one of them.
\Cref{app:refutation} reports the distribution over all four repetitions.

\subsection{What the three episodes show}
\label{app:episode-lessons}

\paragraph{The verifier checks proofs, not fixes.}
Across the three episodes it was called six times, accepted three, and consumed
$122.6$ seconds of $1{,}065.5$ seconds of episode time, or $11.5\%$. What it
rejected was two permission-discipline errors and one construction it declined to
translate. Not one of the six calls said anything about Django, about the issue,
or about whether the fix was the right fix. That is the correct behavior for a
proof checker. A proof-checking channel on its own cannot carry a correctness
claim about a repository. \Cref{app:cost} shows the same pattern at scale: for
Opus~4.8 the verifier's share of episode time is $20$ to $27\%$ under
\nagini{}, and the calls concentrate where the artifact is the only deliverable.

\paragraph{Verification does not entail resolution, and the gap has a shape.}
The gap is not vacuity. All three specifications in this section are
discriminating: the pre-fix behavior violates each of them, which is the
property we require of every specification in the corpus. The gap is the choice
of abstraction. Two of the three episodes model a three-valued behavior with a
two-valued abstraction, and once that choice is made the missing distinction
cannot be recovered by proving harder. \Cref{app:episode-e2e} passes because its small
specification happens to capture the whole of a one-line obligation, and
\cref{app:episode-unresolved} fails because its equally small specification
captures two-thirds of a three-case one.

\paragraph{The property audit is the instrument that sees the gap.}
A verifier cannot report an abstraction that is too coarse. A coarse abstraction
is not an error; it is a smaller claim, correctly proved. The audit asks the
separate question, and its per-property verdicts say where the answer went
wrong: in \cref{app:episode-synth} four properties were discharged by machine and
the fifth, faithfulness, was rejected unanimously with the omission named. The
same asymmetry holds in aggregate. Faithfulness accounts for $48.6\%$ of failures
in this run against $3.0$ to $11.2\%$ for the other four. We report the synthesis
setting per property rather than as a single rate.

\paragraph{Provenance of the quoted material.}
The three episodes are \mbox{\id{django-11179}} and \mbox{\id{django-16667}} in the
Opus~4.8 \nagini{} runs of the Verified End-to-End and Specification Synthesis
settings, first repetition. Every quoted turn, verdict, ballot, and timing is
read from their records, and every count in \cref{tab:episodes} is stored with
them.

\section{Artifacts, licenses, and intended use}
\label{app:licenses}

The release has two provenances, and the terms that apply follow from which one a
file belongs to. The task data is copied unchanged from two existing benchmarks,
which in turn draw on the public issue trackers and commit histories of fifteen
open-source projects. The specifications, proofs, evidence, and admission records
are new work written for this release. This section states, for each part, what it
is, where it came from, and under what terms it may be used. It then states what a
released bundle certifies and what it does not.

\paragraph{What the release contains.}
\Cref{tab:release-parts} lists the parts. Three of them we do not redistribute:
the repository snapshots, the container images, and the verifier distributions are
named, pinned, and fetched, which keeps our correctness oracle identical to the
upstream one. For tasks originating from copyleft-licensed repositories, some artifacts are not redistributed directly. The release provides the available formal specifications and references to the upstream tasks, together with scripts for restoring the omitted materials. Details are documented in the \benchmark repository.

\begin{table}[htbp]
\centering
\caption{The parts of the release, their provenance, and their terms. ``Not
redistributed'' means the release names the object and the pipeline fetches it;
nothing is vendored. The three rows in the last block are what a reader must
obtain elsewhere to re-derive a stamp.}
\label{tab:release-parts}
\small
\setlength{\tabcolsep}{4pt}
\begin{tabular}{@{}L{3.4cm}L{5.7cm}L{3.5cm}@{}}
\toprule
\hd{Part} & \hd{Where it comes from} & \hd{Terms} \\
\midrule
Task rows: problem statement, base and environment commits, reference patch,
test patch, hidden test names
& Copied byte-identically from \sbv and \swepro, which draw them from the
projects' issue trackers and commit history
& Source benchmark; quoted repository text under its project's license \\
\addlinespace[2pt]
Specification modules and pre-fix twins
& Written for this release from the issue text, the reference patch, and the
repository behavior the patch changes
& CC BY-NC 4.0 \\
\addlinespace[2pt]
Correspondence and provenance maps
& Written for this release; they name repository symbols and quote short source
excerpts
& CC BY-NC 4.0; quoted source under its project's license \\
\addlinespace[2pt]
Property test suites, axiom probes, differential harnesses
& Written for this release
& CC BY-NC 4.0 \\
\addlinespace[2pt]
Admission records
& Emitted by the construction pipeline as each check closed
& CC BY-NC 4.0 \\
\addlinespace[2pt]
Construction pipeline, evaluation harness, analysis and figure scripts
& Written for this release
& CC BY-NC 4.0 \\
\midrule
Repository snapshots
& Not redistributed; a bundle names the project and the commit
& Project license, \cref{tab:projects} \\
\addlinespace[2pt]
Per-instance container images
& Not redistributed; pulled by name from the source benchmarks' published
registries
& Image publisher's terms \\
\addlinespace[2pt]
Verifiers, provers, SMT solvers
& Not redistributed; installed at the pinned revisions of \cref{tab:pins}
& \cref{tab:pins} \\
\bottomrule
\end{tabular}
\end{table}

\paragraph{Task data provenance.}
Both source benchmarks are publicly distributed dataset records, and we resolve
each at a fixed revision and copy the task fields verbatim. The task-identity
check of \cref{tab:gates} re-verifies byte identity for every admitted bundle at
release time, so any number reported on \benchmark is comparable to the
corresponding number upstream. Neither of the two dataset records we resolve
declares a license identifier, and we assert none on their behalf. We redistribute
the task fields because a bundle is not usable without them, and we do not
relicense them; a user who needs terms for the task data should take them from the
source distribution. All of that text is derived from public project artifacts:
issue reports, review comments, commits, and test files. The projects' own
licenses are the operative ones wherever repository text is quoted.

\paragraph{The upstream projects.}
\Cref{tab:projects} lists the fifteen projects the two corpora draw on, with their
instance counts and licenses. Both distributions are skewed: Django alone supplies
$231$ of the $500$ \sbv instances, and the three \swepro projects are split
$96/91/79$. The permissive-license families dominate the \sbv side: six projects
under BSD-3-Clause, one under BSD-2-Clause, two under Apache-2.0, one under MIT,
and Matplotlib under its own PSF-derived license, with one copyleft exception,
Pylint. \swepro is the opposite: all three of its projects are copyleft, two under
GPL-3.0-or-later and one under AGPL-3.0. Anyone redistributing modified bundles
should read those three rows carefully, because the reference patches and the
source excerpts quoted in a \swepro bundle's maps carry the copyleft terms of the
project they came from. We do not restate or reinterpret any project's license,
and we do not ship a project's source tree. A bundle names the project and the
commit, and \cref{tab:projects} maps the project to its terms.

\begin{table}[htbp]
\centering
\caption{The projects the corpora draw on, with instances per corpus and the
license each project distributes itself under. License identifiers are SPDX where
one applies; Matplotlib distributes under its own PSF-derived license, for which
there is no SPDX identifier.}
\label{tab:projects}
\small
\setlength{\tabcolsep}{8pt}
\begin{tabular}{@{}lrrl@{}}
\toprule
\hd{Project} & \hd{Verified} & \hd{Pro} & \hd{License} \\
\midrule
django/django                 & 231 & \na & BSD-3-Clause \\
sympy/sympy                   &  75 & \na & BSD-3-Clause \\
sphinx-doc/sphinx             &  44 & \na & BSD-2-Clause \\
matplotlib/matplotlib         &  34 & \na & Matplotlib license \\
scikit-learn/scikit-learn     &  32 & \na & BSD-3-Clause \\
astropy/astropy               &  22 & \na & BSD-3-Clause \\
pydata/xarray                 &  22 & \na & Apache-2.0 \\
pytest-dev/pytest             &  19 & \na & MIT \\
pylint-dev/pylint             &  10 & \na & GPL-2.0-or-later \\
psf/requests                  &   8 & \na & Apache-2.0 \\
mwaskom/seaborn               &   2 & \na & BSD-3-Clause \\
pallets/flask                 &   1 & \na & BSD-3-Clause \\
\midrule
ansible/ansible               & \na &  96 & GPL-3.0-or-later \\
internetarchive/openlibrary   & \na &  91 & AGPL-3.0 \\
qutebrowser/qutebrowser       & \na &  79 & GPL-3.0-or-later \\
\midrule
\hd{Total}                    & \hd{500} & \hd{266} & \\
\bottomrule
\end{tabular}
\end{table}

\paragraph{Container images.}
The correctness oracle runs the task's hidden tests inside the per-instance image
the source benchmark publishes, and neither corpus uses an image we built. \sbv
images are published by the source benchmark under
\texttt{swebench/\allowbreak sweb.eval.x86\_64}, one per instance at the
\id{latest} tag; \swepro images are published under
\texttt{jefzda/\allowbreak sweap-images}, with the tag fixed per instance by the
frozen task row. We publish no image. An image bakes a repository checkout and its
whole dependency environment, so it carries the project's license along with the
terms of everything installed alongside it.

\paragraph{Verifiers, provers, and solvers.}
A verdict is only meaningful relative to the prover that produced it, so the
toolchain pins in \cref{tab:pins} are part of the artifact. Both Lean-family
backends elaborate under one toolchain and import the same Mathlib revision, so a
single prover configuration accounts for every Lean-family verdict. The table
records each Lean package by revision, including the transitive ones, and the two
SMT solvers by release version.

\begin{table}[htbp]
\centering
\caption{Verifier and prover pins. Revisions are shown as the leading twelve
hexadecimal digits; the release records them in full. The two solver rows give
release versions rather than package revisions.}
\label{tab:pins}
\small
\setlength{\tabcolsep}{6pt}
\begin{tabular}{@{}llll@{}}
\toprule
\hd{Component} & \hd{Role} & \hd{Pin} & \hd{Terms} \\
\midrule
Nagini            & Contract verifier for annotated Python & 1.2.0 & MPL-2.0 \\
\midrule
Lean toolchain    & Elaborator and kernel & v4.24.0 & Apache-2.0 \\
Mathlib           & Mathematical library & \texttt{f897ebcf72cd} & Apache-2.0 \\
Velvet            & Imperative language and its logic & \texttt{2cf0acb04c51} & Apache-2.0 \\
Loom              & Solver-backed proof automation & \texttt{d10340821daf} & Apache-2.0 \\
batteries         & Standard library extensions & \texttt{8da40b72fece} & Apache-2.0 \\
aesop             & Proof search & \texttt{725ac8cd67ac} & Apache-2.0 \\
Qq                & Typed quotations & \texttt{dea6a3361fa3} & Apache-2.0 \\
proofwidgets      & Interactive display & \texttt{556caed0eadb} & Apache-2.0 \\
importGraph       & Dependency tooling & \texttt{d768126816be} & Apache-2.0 \\
plausible         & Randomized falsification & \texttt{dfd06ebfe8d0} & Apache-2.0 \\
LeanSearchClient  & Search client & \texttt{99657ad92e23} & Apache-2.0 \\
auto              & Automation front end & \texttt{36d85bf6372f} & Apache-2.0 \\
Cli               & Command-line front end & \texttt{91c18fa62838} & MIT \\
\midrule
z3                & SMT solver & 4.15.4 & MIT \\
cvc5              & SMT solver & 1.3.1 & BSD-3-Clause \\
\bottomrule
\end{tabular}
\end{table}

\paragraph{Software the pipeline depends on.}
The construction pipeline and the evaluation harness run on Python $3.10$ or
later. They use the source benchmark's own package (MIT) for the task rows and the
container conventions, the dataset and model-hub clients (both Apache-2.0) to
resolve the frozen rows, and a unified-diff library (MIT) for patch handling.
Hypothesis (MPL-2.0) generates the inputs behind the differential harnesses and
the axiom probes. Model access goes through the cloud provider's SDK (Apache-2.0), and the
figures in this appendix are drawn with Matplotlib (its own license) and NumPy
(BSD-3-Clause). Container work goes through Docker; nothing in the pipeline
requires a GPU\@.

\paragraph{Our own terms.}
Everything we wrote is released under the Creative Commons Attribution-NonCommercial 4.0 International Public License (CC BY-NC 4.0): the construction pipeline,
the evaluation harness, the analysis and figure scripts, and the artifacts we
authored inside each bundle. That grant does not extend to the copied task fields,
to repository source quoted inside a bundle, to the container images, or to the
verifier distributions; those remain under the terms listed above.

\paragraph{Model access and what is reproducible.}
The model was reached through its provider's hosted API
(\cref{app:settings}). We release no model weights, and we did not fine-tune
anything. The corpus, the harness, and the mechanical graders are reproducible; the
hosted models, versioned by their providers, are not. Every number in this appendix names
the model, the backend, the setting, and the repetition it came from, and the
repetition spread of \cref{app:seeds} is reported next to the single-run numbers it
bounds.

\paragraph{Personal and sensitive content.}
Task text is public issue-tracker and repository text. It names contributors,
quotes their words, and carries their commit identifiers, and we copy it
byte-identically. We add no personal data, we collect nothing from any person, and
no field in any bundle records a demographic attribute. The prompt an agent sees is
assembled only from the frozen task row and our own instructions, so an episode
record carries no information about any person that the task row did not already
carry. Instances keep their upstream identifiers, so a task withdrawn from a
source benchmark is identifiable here under the same name.

\paragraph{What a green bundle certifies.}
A green stamp carries five scoping statements.

\begin{itemize}[leftmargin=1.4em,itemsep=2pt,topsep=3pt]
\item \textbf{Verification is relative to the specification.} A passing verdict
says the implementation satisfies the contracts that were written down. It does not
say the repository is correct, and it does not say the contracts are the right
ones.
\item \textbf{The specification is partial by design.} It describes the behavior
the issue is about and is silent on the rest of the function and the rest of the
repository. \Cref{app:where-it-binds} states what each of the four languages can and
cannot express.
\item \textbf{Resolution is relative to the hidden tests.} It is the upstream
oracle, unchanged, and it is a test outcome rather than a proof.
\item \textbf{Verification and resolution are different claims.} The corpus
contains instances where a submission earns one and not the other in both
directions; \cref{app:episode-unresolved} works one through end to end, and
\cref{app:refutation} gives the joint distribution.
\item \textbf{The construction-time audit is a measurement, not a proof.} It is a
panel of models running a stated attack suite (\cref{app:attack-suite}). Its
calibration is measured on specifications that the construction loop did not
shape, and \cref{app:reaudit} reports it.
\end{itemize}

\paragraph{Intended use.}
The release is built for four uses. The first is measuring verified program
repair: an evaluation in which a submission has to satisfy a
specification and resolve the issue, with the two graded separately. The second is
ablating what a specification is worth, using the nine settings of
\cref{app:modes}: the same instances with localization, a specification, a
verification obligation, or none of them. The third is studying specification
synthesis and its failure modes, where the per-property audit outcomes of
\cref{app:refutation} are more informative than a single rate. The fourth is
narrower. Each bundle is a self-contained verifier input with a labeled expected
outcome, and the property test suites together comprise $23{,}482$ labeled cases
across three verifiers, usable on their own as a regression suite for the
verifiers themselves.

\paragraph{Uses we advise against.}
Four, each with its reason.

\begin{itemize}[leftmargin=1.4em,itemsep=2pt,topsep=3pt]
\item \textbf{Training on the corpus and then reporting on it.} The corpus is
small enough to memorize and public enough to be scraped. A model trained on the
bundles will score higher without being better at the task. The specification
settings are the most vulnerable, since a memorized specification is a memorized
answer. Anyone whose training data includes these bundles should say so alongside
the number.
\item \textbf{Reading a pass rate as a claim about the projects.} The instances
are the ones for which a formal twin could be built and admitted;
\cref{app:pro-nongreen} lists what that excludes and why, and the exclusions are
not a random sample of maintenance work.
\item \textbf{Reusing the audit panel as a general-purpose specification grader.}
The panel's usefulness is a measured property of the panel together with its
attack suite and its retained controls. The prompt on its own has no calibration
outside the harness that measures it.
\item \textbf{Reading a verified-but-unresolved submission as a defect in the
specification.} It can be the opposite: a correct proof of a claim that was
coarser than the issue. \Cref{app:episode-unresolved} is a worked instance, and
\cref{app:episode-synth} shows the audit naming the same omission in a synthesized
specification.
\end{itemize}

\paragraph{Cost of re-deriving the release.}
Re-deriving the mechanical evidence is cheap and needs no model access.
Re-verifying every released specification is $2{,}233$ verifier verdicts and $8.4$
compute-hours; re-running the property test suites is $23{,}482$ cases and $87.0$
compute-hours; re-running mutation adds $25{,}457$ generated mutants. The
contract-annotated Python backend dominates all three, at $25.8$ seconds per
reference verification on the \sbv corpus, against $10.5$ and $6.0$ for the two
Lean-family backends (\cref{app:verif-effort}). The container-side checks
(resolution, axiom probes, conformance) are dominated by image pull and
environment setup rather than by the checks themselves. Only the two agentic
admission checks and the evaluation campaign need model access; the campaign's
own cost is in \cref{tab:totals}.

\paragraph{Versioning and re-admission.}
The release is a snapshot. Every bundle carries its own admission record naming
each check, its outcome, and the verifier configuration it ran under, so a bundle
can be re-admitted from the bundle alone. Re-running a check changes the record,
never the artifacts; \cref{app:reaudit} describes the case where a re-audit
overturns a recorded outcome and what happens to the bundle when it does.

\end{document}